\documentclass{article}

    \PassOptionsToPackage{numbers, compress}{natbib}
\usepackage[numbers,sort]{natbib}
\usepackage{multicol}
\usepackage[preprint]{neurips_2023}
\usepackage{floatrow}
\newfloatcommand{capbtabbox}{table}[][\FBwidth]
\usepackage{wrapfig}

\usepackage{graphicx}
\usepackage[utf8]{inputenc} % allow utf-8 input
\usepackage[T1]{fontenc}    % use 8-bit T1 fonts
\usepackage{hyperref}       % hyperlinks
\usepackage{url}            % simple URL typesetting
\usepackage{booktabs}       % professional-quality tables
\usepackage{amsfonts}       % blackboard math symbols
\usepackage{nicefrac}       % compact symbols for 1/2, etc.
\usepackage{microtype}      % microtypography
\usepackage{xcolor}         % colors
\usepackage[english]{babel}
\usepackage{amsthm}
\usepackage[ruled,linesnumbered]{algorithm2e}
\theoremstyle{definition}

\theoremstyle{assumption}
\newtheorem{assumption}{Assumption}[section]

\title{Unleashing the Potential of Unsupervised Deep Outlier Detection  through Automated Training Stopping}

\author{%
  Yihong Huang \\
  East China Normal University \\
  \texttt{hyh957947142@gmail.com} \\
  \And
  Yuang Zhang \\
  East China Normal University \\
  \texttt{51255902045@stu.ecnu.edu.cn} \\
  \AND
  Liping Wang\thanks{Corresponding author} \\
  East China Normal University \\
  \texttt{lipingwang@sei.ecnu.edu.cn} \\
  \And
  Xuemin Lin \\
  Shanghai Jiao Tong University \\
  \texttt{lxue@cse.unsw.edu.au} \\
}

\begin{document}

\maketitle

\begin{abstract}
Outlier detection (OD) has received continuous research interests due to its wide applications. With the development of  deep learning, increasingly  deep OD algorithms are proposed. Despite the availability of numerous deep OD models, existing research has reported that the performance of  deep models is extremely sensitive to the configuration of hyperparameters (HPs). However, the selection of HPs for deep OD models remains a notoriously difficult task due to the lack of any labels and long list of HPs.
In our study. we  shed light on an essential factor, training time,  that can introduce significant variation in the performance of deep model.  
Even the performance is stable across other HPs, training time itself
can cause a serious HP sensitivity issue.
  Motivated by this finding, we are dedicated to formulating a strategy to terminate model training at the optimal iteration. Specifically, we propose a novel metric called \textit{loss entropy} to internally evaluate the model performance during training while an automated training stopping algorithm is devised.
To our knowledge, our approach  is the first to enable reliable identification of the optimal training iteration during training without requiring any labels. 
Our experiments on tabular, image datasets show that our approach can be applied to diverse deep models and datasets. It not only  enhances  the robustness of deep models to their HPs, but also improves the performance and reduces  plenty of  training time compared to  naive training.
\end{abstract}

\section{Introduction}
Outlier Detection (OD), identifying the instances that significantly deviate from the majority \cite{uod-definition}, has received continuous research interests \cite{od-survey,ADbench,Ts-benchmark,god-benchmark} due to its wide applications in various fields, such as finance, security \cite{financial-example,security-example}. With the rapid development of deep learning, increasingly deep OD algorithms  are  proposed \cite{deep-od-survey-2021,deep-od-survey-2019,deep-od-survey-3}. Compared to traditional algorithms, deep ones can handle kinds of complex data (e.g. image, graph) more effectively with an end-to-end optimization manner and incorporate the most recent advances in deep learning \cite{cv-survey,gnn-survey}. 

Unsupervised OD aims to identify outliers in a contaminated dataset (i.e., a dataset consisting of both normal data and outliers) without the availability of labeled data \cite{inlier-priority}. We limit our discussion to unsupervised OD, which is more challenging and widely studied.  Despite the impressive performance of deep OD, deep detectors from various families suffer from the issue of hyperparameters (HP) sensitivity and tuning  in fully unsupervised setting \cite{robod}. In specific, not only their performance  heavily relies on the selection of HPs , but also their long list of HPs (compared to traditional detectors)  are difficult to tune \cite{Internal-evaluation-paper} in the absence of labels for evaluation. To overcome this obstacle, unsupervised model selection for OD \cite{Internal-evaluation-paper, MetaOD} aims at selecting the best model (i.e. an algorithm with its HP configuration) from a pool of given models. More recently, a deep hyper-ensemble solution \cite{robod} is proposed to unleash the potential of existing algorithms by aggregating all predictions from models with different HP configurations.  However, both these approaches require training a large number of models, resulting in a significant increase in training time.

\begin{wrapfigure}[19]{r}{0.5\textwidth}
\vspace{-0.6cm}
  \begin{center}
\includegraphics[width=1.0\textwidth]{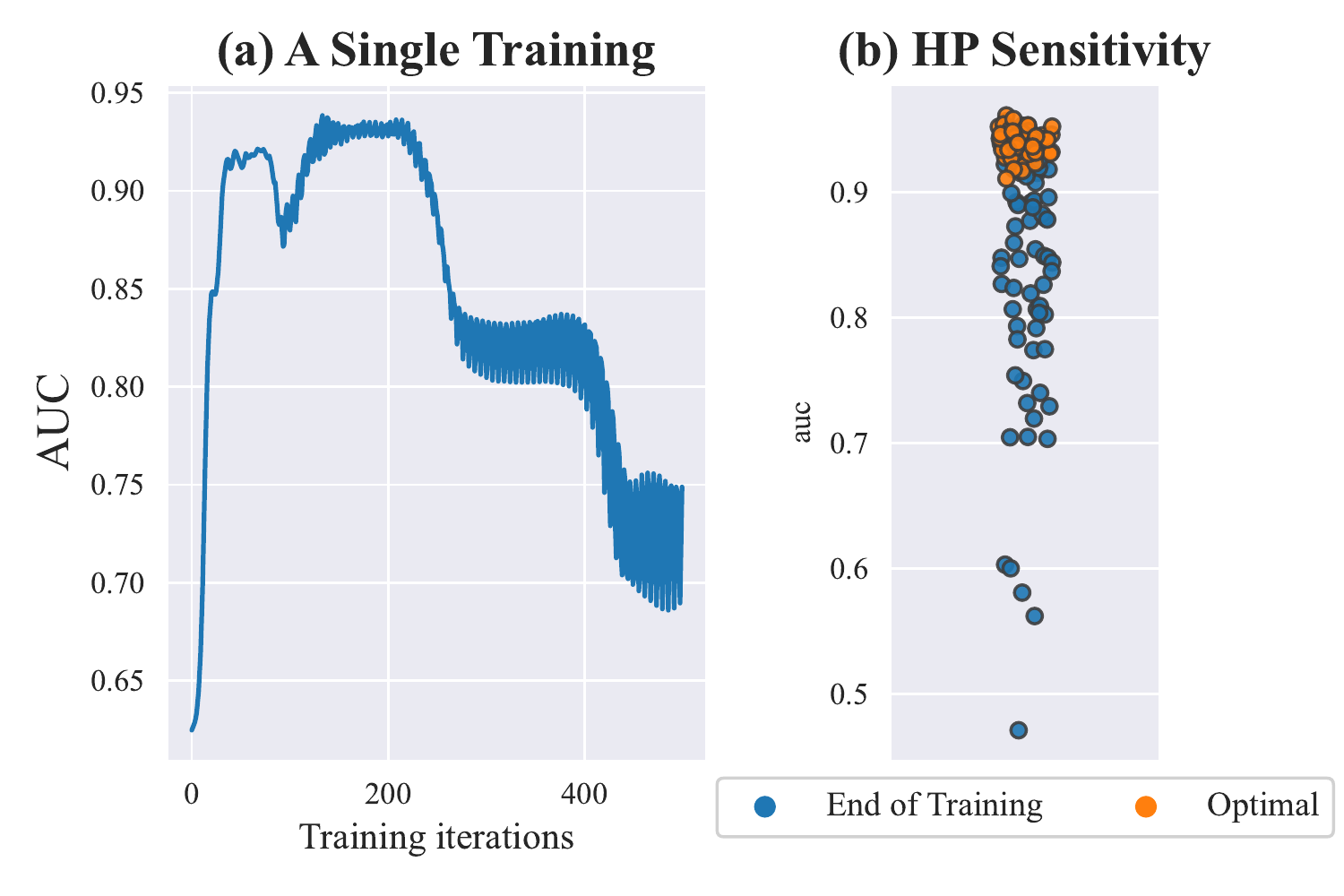}
  \end{center}
  \vspace{-0.2cm}
  \caption{AUC performance of Autoencoder (AE) on  dataset \textit{vowels}. (a) An example of unsupervised training AE. (b) The end-of-training AUC varies across different HP configurations (circles), but the AUC distribution of the optimal iteration during training  shows a notable reduction in variation. }
  \label{auc-iteration}
\end{wrapfigure}
In this paper, we revisit the HP sensitivity issue from the perspective of utmost potential of model during  training. Our key finding is that the training process can introduce significant variation in model’s detection performance, as depicted in Fig \ref{auc-iteration} (a). This variation is attributed to the presence of outliers in  training data. In this case, even if the model’s performance is stable across the other HPs, training time itself can cause serious HP sensitivity issue (see Fig \ref{auc-iteration} (b)), as the model with different HP configurations may require varying  training iterations to achieve its optimal performance. According to our knowledge, there is no systematic study on training time’s effect on unsupervised  OD. Only a few studies \cite{gan-ensemble,randnet,robod} employ model ensembles to enhance the robustness of  models to training time.

% \begin{figure}[h] \centering \begin{minipage}[b]{150.0pt} \centering \includegraphics[width=150.0pt]{fig/auc-iter-skin.pdf} \caption{Left half} \end{minipage} \hfill \begin{minipage}[b]{150.0pt} \centering \includegraphics[width=150.0pt]{fig/auc-iter-skin.pdf} \caption{Right half} \end{minipage} \end{figure}

The goal of our research  is two-fold. \textbf{Firstly}, our primary objective is to illustrate that training time plays a vital role in reducing HP sensitivity in  deep OD models. To this end, we conduct a large-scale analysis of experiments on deep OD models, which quantitatively demonstrates that HP sensitivity in some cases can be greatly alleviated by halting training at the optimal moment. Interestingly, our experimental results also reveal that deep OD models typically require much less training time  than previous thought. \textbf{Secondly}, motivated by our finding, we propose a novel internal evaluation metric called \textit{loss entropy}, which evaluates the model's performance solely based on the data loss. Through the use of loss entropy, optimal training stop moment can be predicted with remarkable accuracy for the first time. Based on this powerful metric, we devise an automated training stopping algorithm that can dynamically determine the optimal training iteration. Our main contributions are as follows.
\begin{itemize}
    \item \textbf{The first systematic study on training time in unsupervised OD:} To the best of our knowledge, we are the first to associate HP sensitivity with training time, and investigate how to evaluate a model's in-training performance without using any validation labels.
    \item \textbf{Loss entropy, a novel internal evaluation metric:}  We propose a novel metric to evaluate the model’s in-training performance solely based on the data loss, which  captures the model's state on learning useful signals from a contaminated training set efficiently. (Sec. \ref{loss-entropy-sec})
    \item \textbf{EntropyStop, an automated training stopping algorithm:} We develop an automated training stopping algorithm that adapts to the dataset and HP configuration to determine the optimal training iteration. Our algorithm eliminates the need for manual tuning the training iterations and improves the robustness of deep OD algorithms to their HP configuration, while reducing  training time by 60-95\% compared to naive training. (Sec. \ref{entropy-stop-sec})
    \item \textbf{Extensive experiments:} We conduct extensive experiments to validate the effectiveness of our approach. The results show that our approach can be applied to diverse deep OD models and datasets to alleviate the HP sensitivity issue and improve the detection performance, while taking much less time than existing solutions. (Sec. \ref{sec:exp})
\end{itemize}
Overall, our contributions provide a  comprehensive understanding of the role of training time in deep OD models and offer practical solutions for improving their performance.  To foster future research, we
open-source all code and experiments at  \url{https://github.com/goldenNormal/AutomatedTrainingOD}.

 \section{Related Work}
\textbf{Unsupervised Outlier Detection (OD).}
Unsupervised OD is a vibrant research area, which aims at detecting outliers in  contaminated datasets without any labels providing a supervisory signal during the training \cite{od-survey}.
In some literature like \cite{anogan-unsupervised-semi-example,video-unsupervised-semi-example}, “unsupervised outlier/anomaly detection” actually refers to semi-supervised OD by our definition, as their training set only contains normal data. 
Solutions for unsupervised OD  can be broadly categorized into shallow (traditional) and deep (neural network) methods.
 Compared to  traditional counterparts, deep methods are more adept at handling  large, high-dimensional and complex data \cite{ADbench}. Despite their relatively recent emergence, multiple surveys have been published to cover the growing literature \cite{deep-od-survey-2019,deep-od-survey-2021,deep-od-survey-3}. However, recent research \cite{robod} has shown that popular deep OD methods are highly sensitive to their hyperparameter(s) (HP) while there is no systematical way to tune these HPs, hindering their application in real world. 

\textbf{Model Selection in Unsupervised OD.} 
Various evaluation studies have reported existing outlier detectors  to be quite sensitive to their HP choices \cite{robod,tradional-param-sen-1,tradional-param-sen-2,tradional-param-sen-3}, both traditional and deep detectors. 
Thus, it promotes some  study on selecting a good OD detector with its HP configuration when a new OD task comes, which is called Unsupervised Outlier Model Selection (UOMS) \cite{Internal-evaluation-paper}.  For example,  MetaOD \cite{MetaOD} employs meta-learning to select an effective model (i.e. algorithm with its HPs) which has a good performance on similar historical tasks. Some internal (i.e., unsupervised)  evaluation strategies have been proposed \cite{emmv,ireos,cluster-eval} to evaluate the detection performance of OD detectors, which solely rely on the input data (without labels) and the output (i.e., outlier scores). However, all the above works only focus on the selection of traditional detectors.  For a comprehensive study, we investigate the effectiveness of existing UOMS solutions on deep OD in our experiment, compared with our solution.

\section{Methodology: Entropy-based Automated Training Stopping}
% In this section, we elaborate the technique details of our approach. We will begin by providing the necessary background on Unsupervised OD and present the problem formulation of our study. 
\textbf{Background.}
Unsupervised OD presents a collection of $n$ samples $D = \{\textbf{x}_1,...,\textbf{x}_n\}\in \mathbb{R}^{n\times d}$, which consists of  inlier set $D_{in}$ (i.e. the normal) and
 outlier set $D_{out}$. The goal is to train a  model $M$ to output outlier score $\textbf{O} := M(D) \in \mathbb{R}^{n\times1}$, where a higher  $o_i$ indicates greater outlierness for $\textbf{x}_i$.  

% \begin{definition}[Optimal Training Iteration]
% Given a deep OD model trained on 
% \end{definition}

\textbf{Problem (Unsupervised Outlier Training Halting)}  \textit{Given a deep model $M$'s unsupervised training on a contaminated dataset $D \in \mathbb{R}^{n\times d}$ where $T$ is the optimal iteration when $M$ has the best detection performance on $D$ evaluated by ground truth label $\textbf{Y} \in \mathbb{R}^{n\times 1} $. The goal is to halt the training process at iteration $T$ without relying on the label $\textbf{Y}$ and obtain the corresponding prediction  $\textbf{O} \in \mathbb{R}^{n\times 1}$.}

To clarify, an iteration in this context denotes the process of   gradients updates computed on a single batch of training data, while an epoch denotes the process of iterating over all the batches  once.

\subsection{Basic Assumption}
To simplify the problem, we assume the composition of the loss function. This assumption is applicable to a diverse set of deep OD models, as elaborated in a survey \cite{self-supervised-od}.

% The main cause of training time sensitivity is the presence of outlier in training set, which provides some harmful signals for learning the distribution of normal data. In this case, we propose a novel metric, \textit{loss entropy}, to capture the model’s state on learning useful signals from polluted dataset. Our metric is build on the following assumption:
\begin{assumption}[Loss function]  Training loss can be formulated as $\mathcal{L}(D;\Theta) = \mathcal{J}(D;\Theta) + \mathcal{R}(D;\Theta)$.
where $\mathcal{L}$ is the training loss  on dataset $D$ and $\Theta$ is the  learnable parameters of model $M$; $\mathcal{J}(\cdot)$  and $\mathcal{R}(\cdot)$ denote  self-supervised loss  and   auxiliary loss (e.g. L2-regularization), respectively.
\end{assumption}

\begin{assumption}[Alignment]  $\forall \textbf{x}_i,\textbf{x}_j \in D$, if $M(\textbf{x}_i;\Theta) < M(\textbf{x}_j;\Theta)$, then $\mathcal{J}(\textbf{x}_i;\Theta) < \mathcal{J}(\textbf{x}_j;\Theta)$.
\label{assum:align}
\end{assumption}

Assumption \ref{assum:align} ensures that the rank of learned outlier score function (i.e. $M(\cdot)$) is aligned with self-supervised loss, which is valid for many deep OD models, e.g. AE \cite{AE}, Deepsvdd \cite{deep-svdd}, NTL \cite{NTL}. We assume $\mathcal{J}_\Theta(\textbf{X};\Theta) \geq 0$ for  convenience. If not established, $\mathcal{J}_\Theta(\textbf{X};\Theta)^2$ can be used as a substitute. 
As $\mathcal{R}(\cdot)$  serves as  the role like regularization, we omit it in the following analysis for simplicity.

\subsection{Loss Entropy: An Internal Evaluation Metric}
\label{loss-entropy-sec}
\textbf{What does the model learn?} As model $M$ is trained on a polluted set $D$,  learning signals can be classified into useful ones $\{\mathcal{J}(\textbf{x}_i;\Theta)|\textbf{x}_i \in D_{in}\} $ and harmful ones $\{ \mathcal{J}(\textbf{x}_i;\Theta)|\textbf{x}_i \in D_{out}\}$.

\textbf{Overview.}  We introduce  \textit{loss entropy} to track a model's progress in learning useful and harmful signals. Firstly, we demonstrate that  the model puts more efforts on learning harmful signals only when the outlier's loss is significantly greater than that of inliers due to "inlier priority" \cite{inlier-priority}. Drawing on this insight, we analyze the changes in the  loss distribution when the model learns different signals. Finally, we propose using the entropy of the loss as the metric to capture such changes.

\subsubsection{Inlier Priority: a foundational concept in self-supervised OD}
Inlier priority \cite{inlier-priority,DRAE-inlier-priority} means that the intrinsic class imbalance of inliers/outliers in the dataset will make the deep model prioritize minimizing inliers’ loss when inliers/outliers are indiscriminately fed into the model for training. When $\mathcal{R}(\cdot)$ is ignored, the loss $\mathcal{L}$ of $M$ on $D$ can be written as:
\begin{equation}
\small
    \mathcal{L} = \mathcal{J}(D;\Theta) = \frac{1}{n} \sum \mathcal{J}(\textbf{x}_i;\Theta)
    \label{loss-objection-all}
\end{equation}
Then the corresponding loss on $D_{in}$ and  $D_{out}$  can be defined as:
\begin{equation}
\small
    \mathcal{L}_{{in}} = \mathcal{J}(D_{in};\Theta) = \frac{1}{|D_{in}|} \sum \mathcal{J}(\textbf{x}_i;\Theta), \quad \textbf{x}_i \in D_{in}
    \label{loss-objection-in}
\end{equation}
\begin{equation}
\small
    \mathcal{L}_{{out}} = \mathcal{J}(D_{out};\Theta) = \frac{1}{|D_{out}|} \sum \mathcal{J}(\textbf{x}_i;\Theta), \quad \textbf{x}_i \in D_{out}
    \label{loss-objection-out}
\end{equation}
Inlier priority refers to that $M$ will prioritize minimizing the loss on $D_{in}$, resulting in $\mathcal{L}_{in} < \mathcal{L}_{out}$ constantly during the training process. The priority can be justified in the following two aspects:

\textbf{Priority due to the larger quantity of inliers.} Recalling above Eq \ref{loss-objection-all}-\ref{loss-objection-out},   $M$ aims to minimize the overall loss $\mathcal{L}$, which can  be  represented as
$\mathcal{L} = \frac{|D_{out}|}{n}\mathcal{L}_{out} + \frac{|D_{in}|}{n} \mathcal{L}_{in}$.
Since $|D_{in}| \gg |D_{out}|$ usually establishes, the weight of $\mathcal{L}_{in}$ is much larger. Thus, $M$ puts more efforts to minimize  $\mathcal{L}_{in}$.

 \textbf{Priority by gradient updating direction.}
By  minimizing the overall loss $\mathcal{L}$,  learnable weights $\Theta$ are updated by gradient descent
    $\overline{g} = \frac{1}{n} \sum g_i = \frac{1}{n} \sum \frac{\mathrm{d}\mathcal{J}(\textbf{x}_i;\Theta)}{\mathrm{d}\Theta}$,
where $g_i$ is the gradient contributed by $\textbf{x}_i$. 
The effect of gradient update  for minimizing  $\mathcal{J}(\textbf{x}_i;\Theta)$ is
    $g_i^{effect} = \frac{<g_i, \overline{g}>}{|g_i|} = |\overline{g}| cos\theta(g_i,\overline{g})$,
where $\theta(g_i,\overline{g})$ is the angle between  two vectors. 
In most cases, outliers are randomly distributed in the feature space, causing opposing gradient directions, while inliers are densely distributed, leading to more consistent gradient directions. Thus, $\theta(g_i,\overline{g})$ is smaller for inliers than outliers, resulting in a larger $g_i^{effect} $ for $\textbf{x}_i\in D_{in}$.This implies that the overall gradient focuses more on reducing $\mathcal{L}_{in}$.

% \begin{wrapfigure}{r}{0.48\textwidth}
%   % \begin{center}
% \includegraphics[width=0.20\textwidth,height=0.15\textwidth]{fig/inlier_priority_letter.pdf}
% \includegraphics[width=0.20\textwidth,height=0.15\textwidth]{fig/inlier_priority_mnist.pdf}
%   % \end{center}
%   \caption{ An example of unsupervised training of Autoencoder on contaminated dataset \textit{skin}.}
%   \label{inlier-priority}
% \end{wrapfigure}
Overall, our analysis confirms that the model prioritizes minimizing $\mathcal{L}_{in}$ when $\mathcal{L}_{in}$ $\approx$ $\mathcal{L}_{out}$ .  \textit{\textbf{The model shifts its focus to outliers only when  $\mathcal{L}_{out} \gg \mathcal{L}_{in}$}}.

\subsubsection{Loss Entropy: The novel internal evaluation metric}
\textbf{How to evaluate model's learning state?}  The model's detection performance improves with useful signal learning and deteriorates with harmful signal learning. To analyze the model's learning state at iteration $i$, we quantify the ratio of drop speed between two losses as
    $\frac{V_{in}^{i}}{V_{out}^{i}} = \frac{\mathrm{d}\mathcal{L}_{in}^{i}}{\mathrm{d}\mathcal{L}_{out}^{i}} $, where $V_{in}^{i}$ and $V_{out}^{i}$ are drop speed of $\mathcal{L}_{in}$ and $\mathcal{L}_{out}$ at iteration $i$, respectively.  Since learning  signals is essentially the loss reduction, $ \frac{V_{in}^{t_i}}{V_{out}^{t_i}}$ tells which signals  $M$ learns more at iteration $i$.
However, $\mathcal{L}_{in}^{{i}}$ and $\mathcal{L}_{out}^{{i}}$ cannot be calculated without the ground truth labels.

\begin{figure*}
  \centering
  \includegraphics[width=400.00pt]{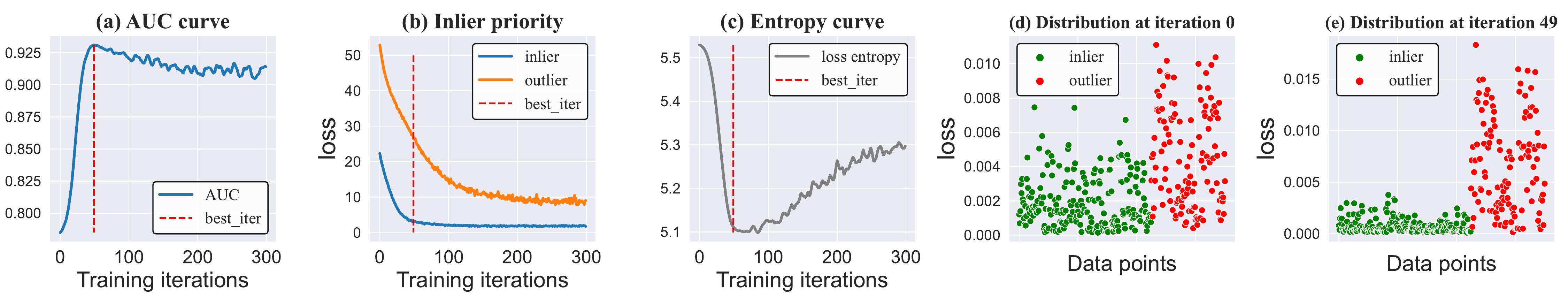}
  \caption{An example of  AE model's training on the dataset \textit{Ionosphere} with 300 iterations.  The optimal iteration evaluated by \textit{AUC} is  the $49^{th}$ iteration.  The y-axis of two  scatter plots (i.e. (d) and (e)) is normalized data loss value $p_i^{loss} \in \mathcal{P}^{loss}$.}
  \label{Fig:loss-distribution}
\end{figure*}
\textbf{Changes in the loss distribution.}  Though direct calculation  is infeasible,  the loss distribution  over the dataset at iteration $i$ can reflect   $\frac{V_{in}^{i}}{V_{out}^{i}}$ due to the intrinsic class imbalance. When $ V_{in}^{i} \gg V_{out}^{i}$, then  the majority of loss (i.e. $\{ \mathcal{J}(\textbf{x}_i;\Theta)|\textbf{x}_i \in D_{in}\} $) has a dramatic decline while the minority of loss (i.e. $\{ \mathcal{J}(\textbf{x}_i;\Theta)|\textbf{x}_i \in D_{out}\} $) remains relatively  large, leading to  a  steeper distribution (See the example in Figure \ref{Fig:loss-distribution}).  Conversely, when $ V_{in}^{i} \ll V_{out}^{i}$,  the   distribution will change towards flatter. Thus, the changes in the shape of the distribution can give some valuable insights into the training process. 

\textbf{Our metric.} We employ the entropy  of loss distribution $\mathcal{P}^{loss}$ to  capture such changes, where entropy \cite{entropy} is a natural metric to measure the flatness of an arbitrary probability distribution $\mathcal{P}$: 
\begin{equation}
\small
    entropy(\mathcal{P}) = \sum_{p_i \sim \mathcal{P}} (p_i \log{p_i}), \quad s.t. \sum_{p_i\sim \mathcal{P}}p_i = 1, p_i\geq0
        \label{entropy}
\end{equation}
Basically, the more flat or uniform  $\mathcal{P}$ is, the larger  $entropy(\mathcal{P})$ is.  To convert  $\{\mathcal{J}(\textbf{x}_i;\Theta) | \textbf{x}_i \in D\}$  into  $\mathcal{P}^{loss}$, we adopt  $\sum_{\textbf{x}_i \in D} \mathcal{J}(\textbf{x}_i;\Theta)$
    to normalize each loss value as
    $
        p_i^{loss} = \frac{\mathcal{J}(\textbf{x}_i;\Theta)}{\sum_{\textbf{x}_i \in D} \mathcal{J}(\textbf{x}_i;\Theta)}
     $.

\begin{figure*}
  \centering
  \includegraphics[width=95.00pt]{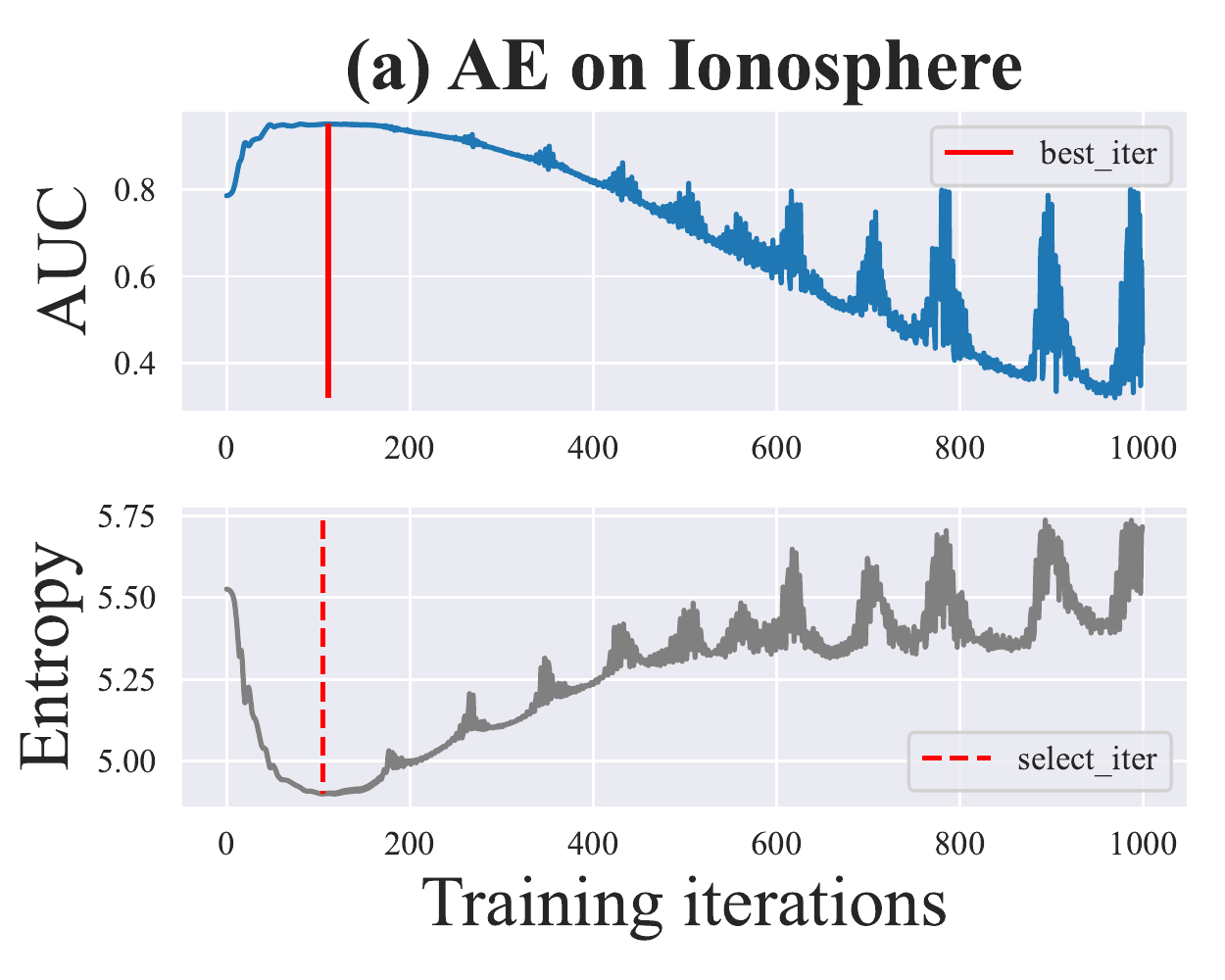}
  \includegraphics[width=95.00pt]{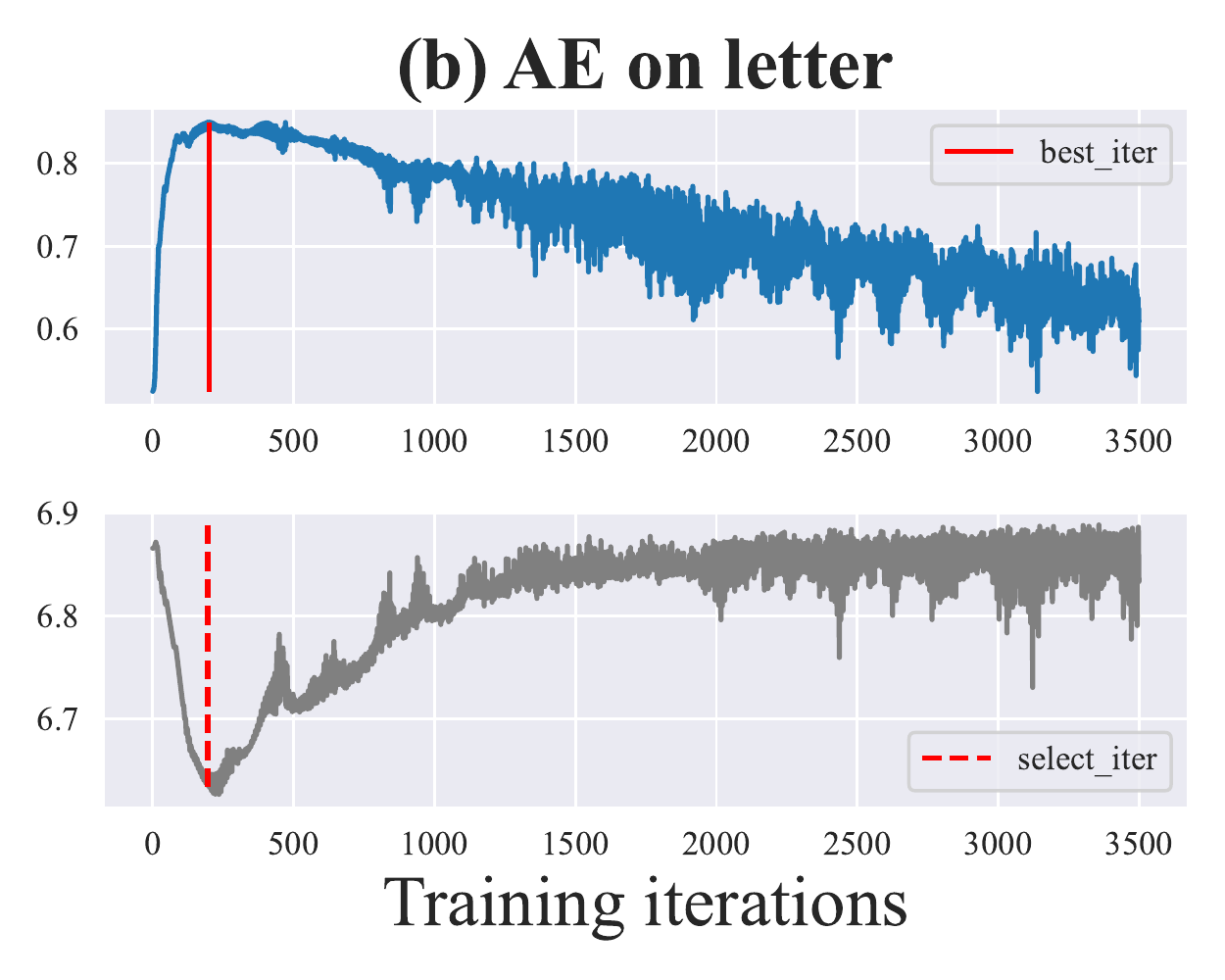}
  \includegraphics[width=95.00pt]{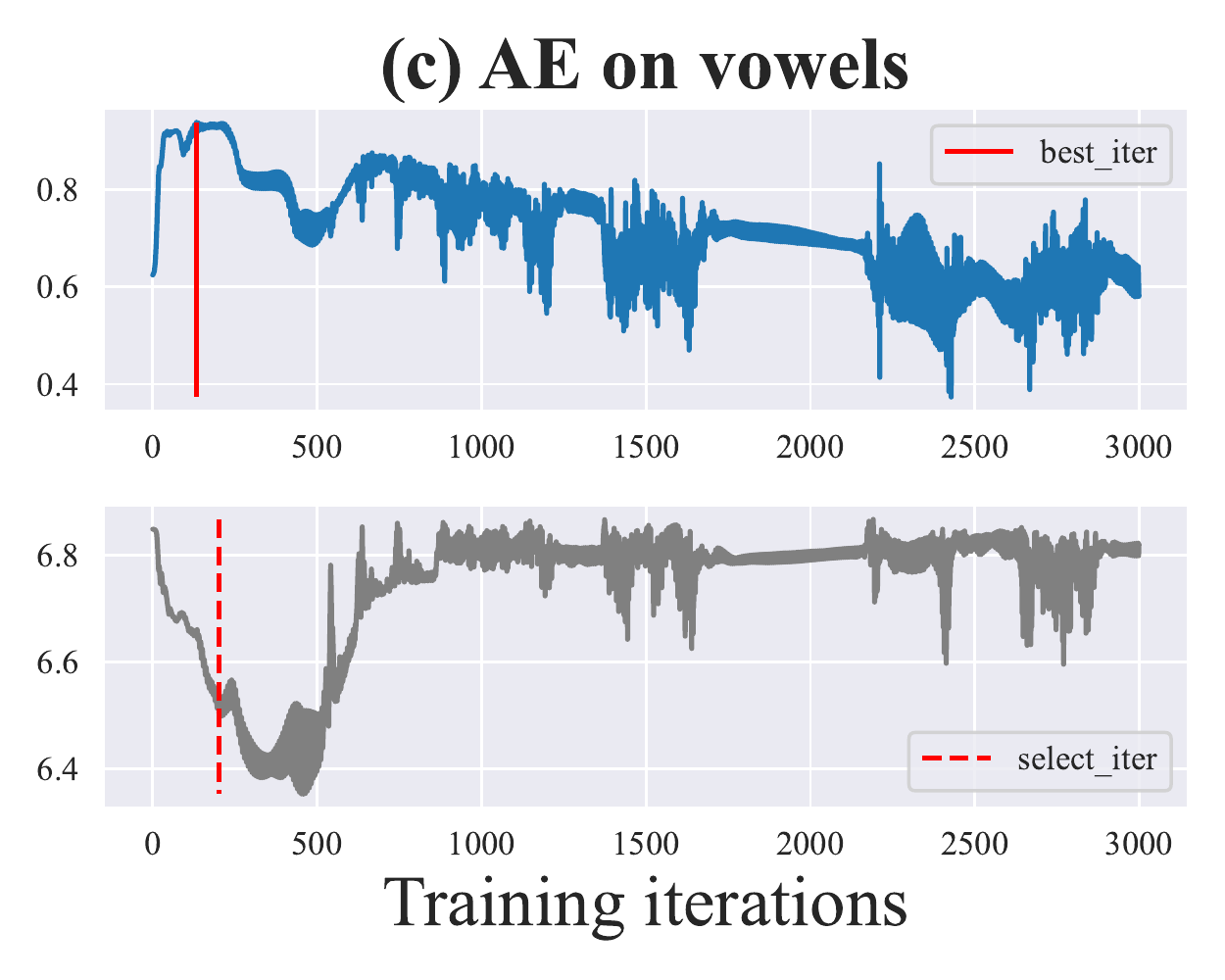}
  \includegraphics[width=95.00pt]{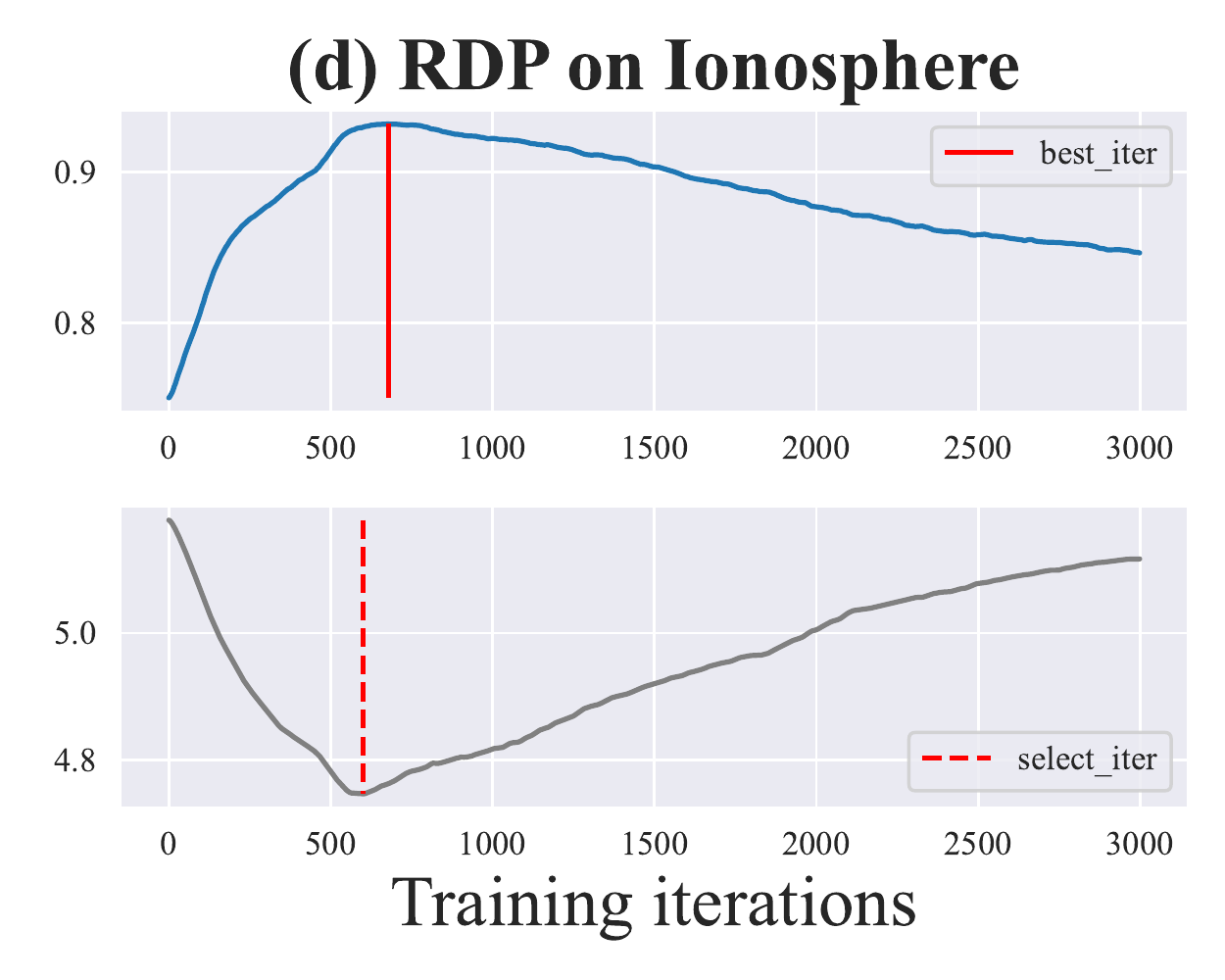}
  \includegraphics[width=95.00pt]{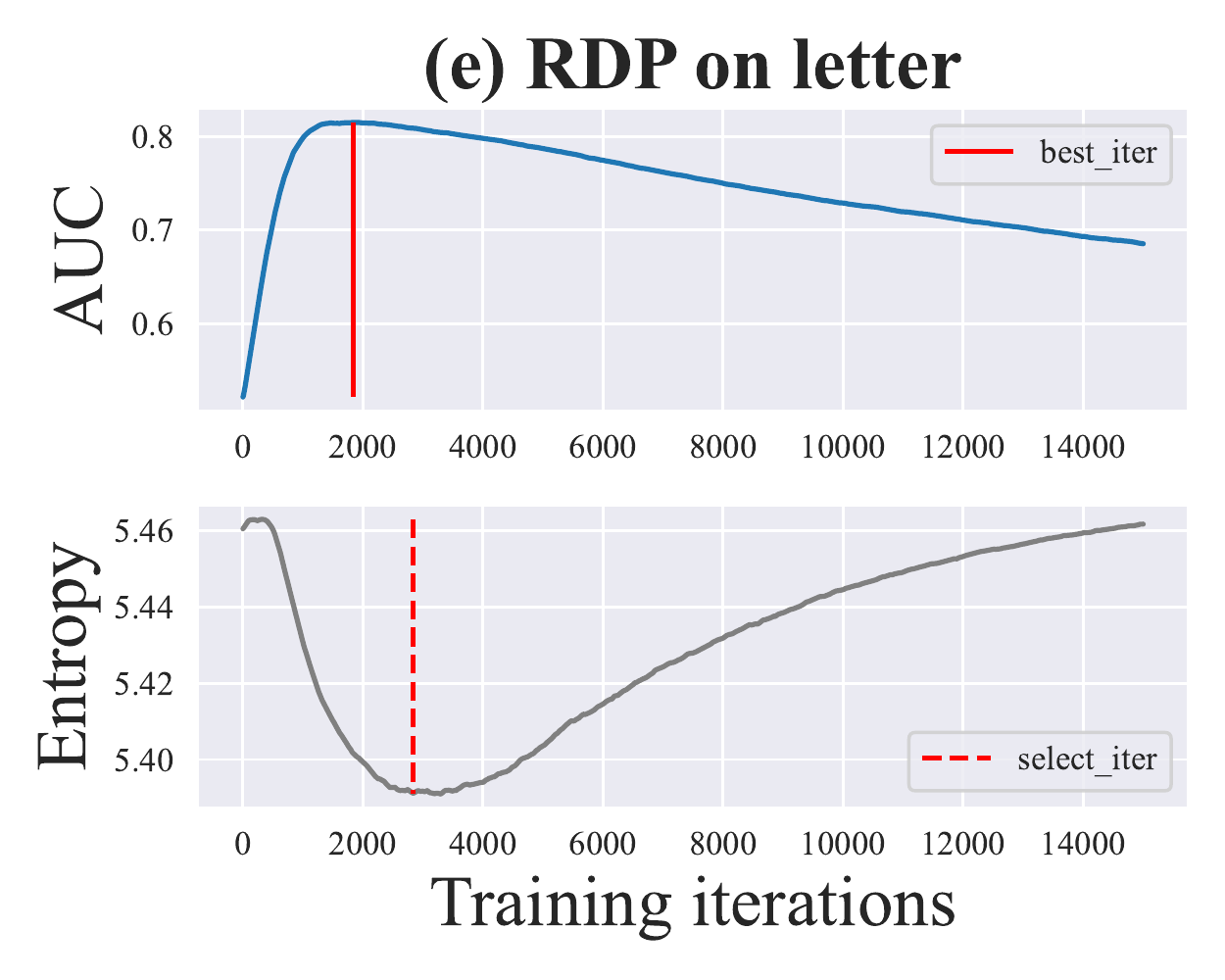}
  \includegraphics[width=95.00pt]{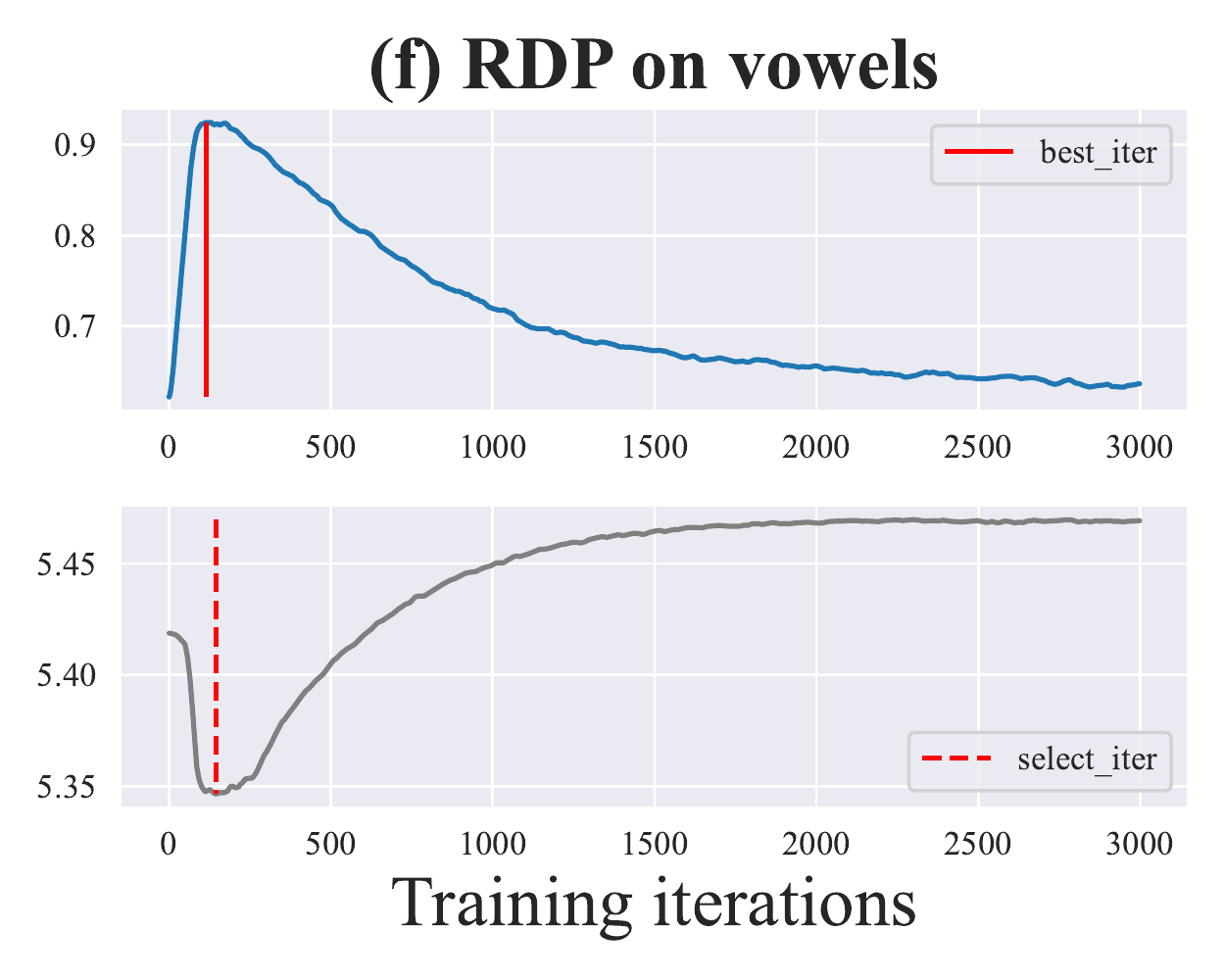}
  \includegraphics[width=95.00pt]{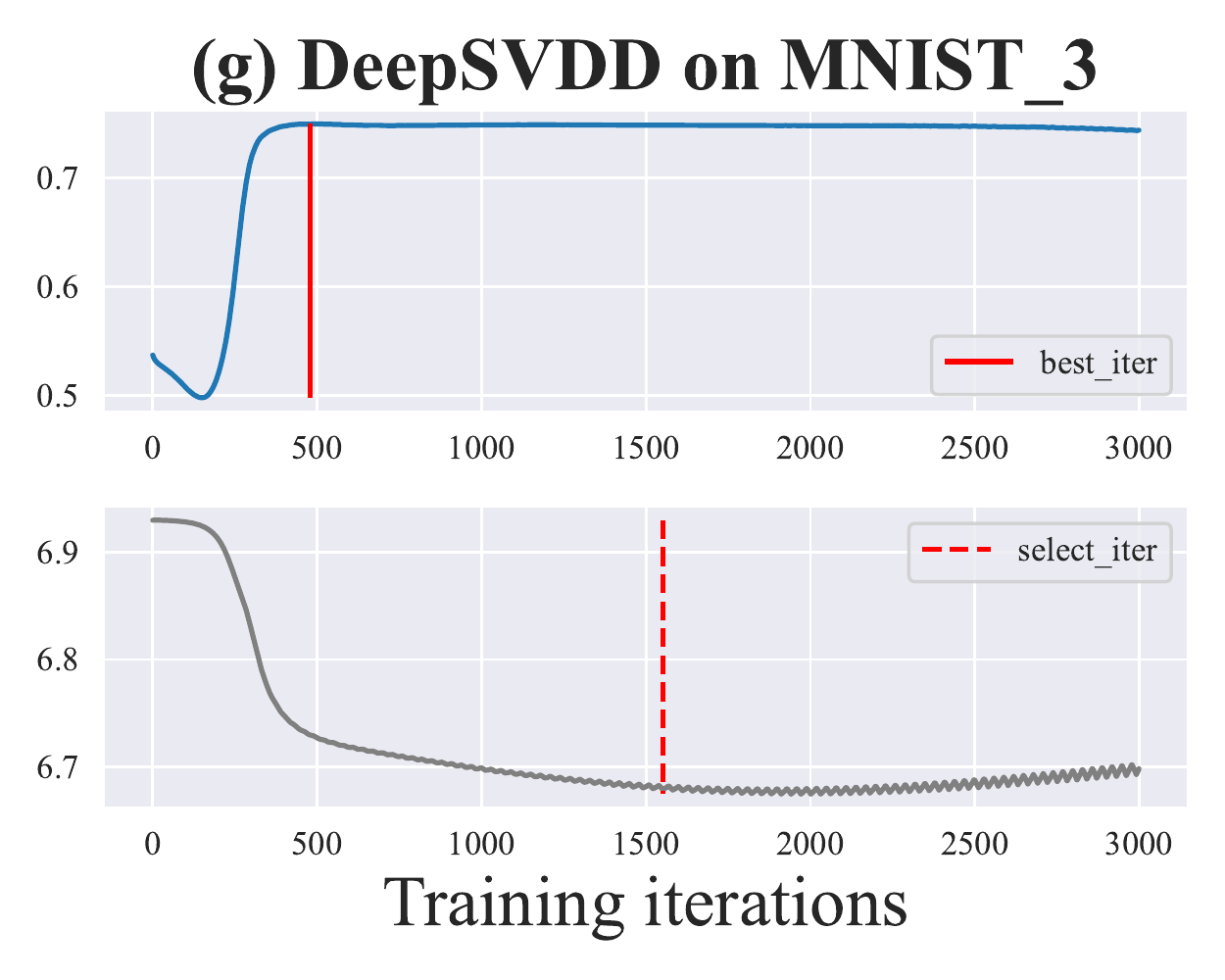}
  \includegraphics[width=95.00pt]{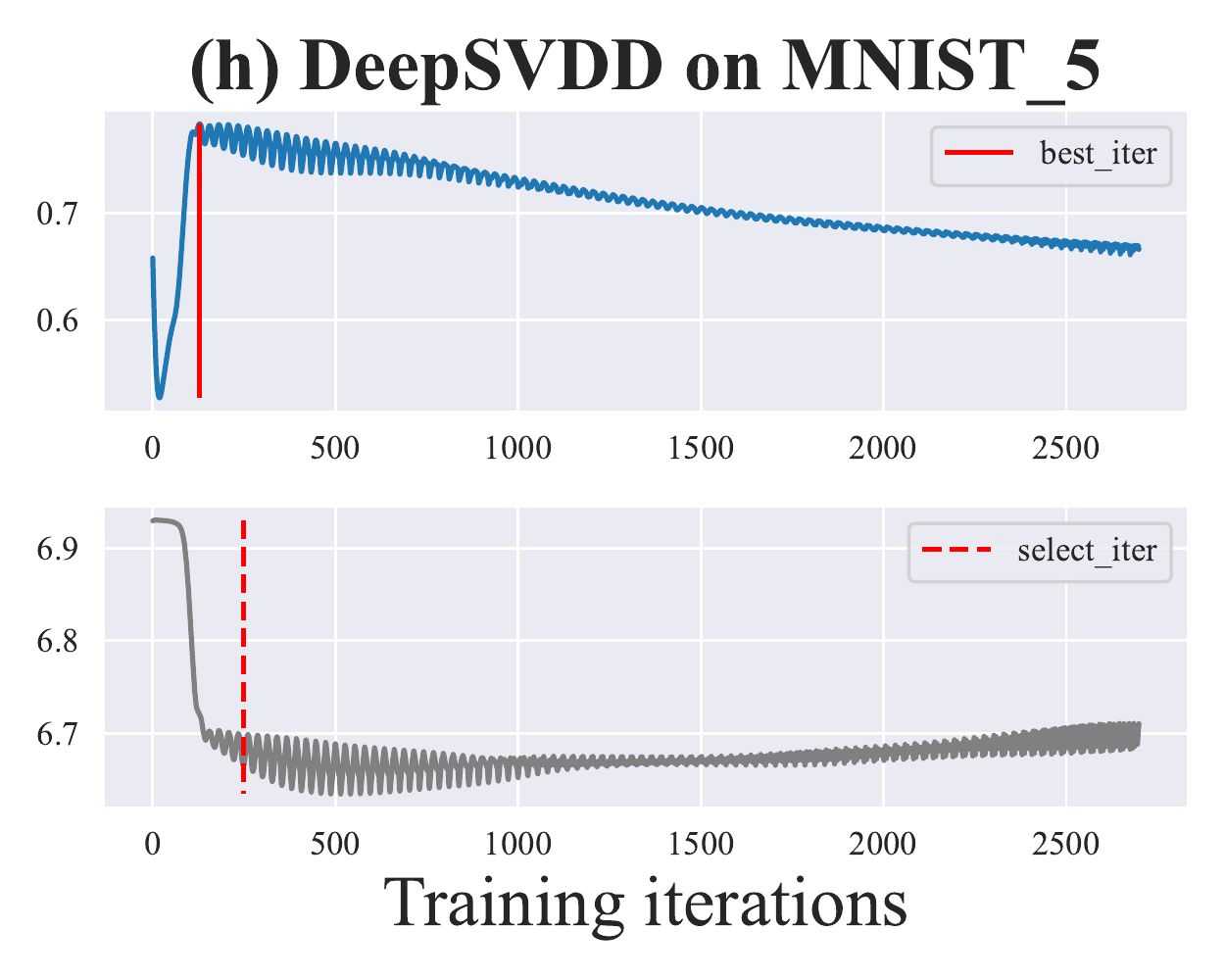}
  \caption{Examples of AUC and loss entropy curves during the training. The ``select\_iter'' denotes the iteration selected by \textit{EntropyStop}. }
  \label{Fig:auc-entropy-corre}
  
\end{figure*}

To achieve efficient and consistent entropy measurement, we randomly sample $N_{eval}$ instances  from $D$ to create  an evaluation set $D_{eval}$ before  training and calculate $entropy(\mathcal{P}^{loss})$ on $D_{eval}$ per iteration.  To ensure accuracy, techniques that introduce randomness (e.g. dropout) are disabled when computing $entropy(\mathcal{P}^{loss})$, although they can still be used during training.
Finally,  \textit{loss entropy} is expected to have a negative correlation with variation in detection performance. 
% If model $M$ learns mainly the useful signals, then the loss distribution become steeper and entropy will decrease. On the other hand, if loss entropy increases, it implies that the distribution becomes flatter due to that $M$ learns  more harmful signals.  
Several examples in Figure \ref{Fig:auc-entropy-corre} show that \textit{loss entropy} provides a reliable measure of variations in AUC.

\subsection{EntropyStop: Automated Training Stopping Algorithm}
\label{entropy-stop-sec}
In this subsection, we will illustrate the primary idea for selecting the optimal iteration  from the entropy curve (see Fig \ref{Fig:auc-entropy-corre}), and propose a  training stopping algorithm to automate this process.

Loss entropy primarily reflects the direction of variation in detection performance, rather than the precise amount of variation. In most cases, the model prioritizes learning useful signals from inliers, leading to a gradual decrease in loss entropy. However, when the model shifts its focus on outliers, the entropy starts to rise, suggesting that training should be terminated. Although the model has a small potential to re-focus on inliers to improve detection performance, our metric cannot measure the extent of this improvement. Therefore, we opt to stop training as soon as the entropy stops decreasing. We present our problem formulation for selecting the optimal iteration below.

\textbf{Problem Formulation.} Suppose $\mathcal{E} = \{e_j\}_{j=0}^E$ denotes the  entropy curve of model $M$. 
When $M$ finishs its $i^{th}$ training iteration, only the subcurve $\{e_j\}_{j=0}^i$ is available.
The goal is to select a point $e_i \in \mathcal{E}$ as early as possible that  (1) $\forall j < i, e_i < e_j$; (2) the subcurve $\{e_j\}_{j=0}^i$  has a downtrend; (3) $\forall q \in (i,k + i)$, the subcurve $\{e_j\}_{j=i}^q$  has no  downtrend.

% \begin{wrapfigure}{r}{0.5\textwidth}
%   \begin{center}
% \includegraphics[width=1.0\textwidth]{fig/alg-figure.pdf}
%   \end{center}
%   \caption{The whole process of \textit{EnttropyStop}}
%   \label{fig:alg}
% \end{wrapfigure}
\textbf{Algorithm.} In above formulation, $k$ is the patience parameter of algorithm. 
As an overview, our algorithm continuously explores new points within  $k$ iterations of the current lowest entropy point $e_i$, and tests whether the subcurve between the new point  and $e_i$ exhibits a downtrend. Specifically,  when encountering a new point $e_q$,  we calculate $G = \sum_{j=i+1}^q (|e_j - e_{j-1}|)$ 
 as the total  variations of the subcurve $\{e_j\}_{j=i}^q$   and the downtrend of the subcurve is then quantified by $\frac{e_i - e_q}{G}$. Particularly, if the subcurve is monotonically decreasing, then $\frac{e_i - e_q}{G} = 1$. To test for a significant downtrend, we use a threshold parameter $R_{down} \in (0,1)$. Only when $\frac{e_i - e_q}{G}$ exceeds $R_{down}$ will $e_q$ be considered as the new lowest entropy point. The complete process  is shown in Algorithm \ref{alg:online}.

\RestyleAlgo{ruled}
\SetKwComment{Comment}{/* }{ */}
\SetKwInput{kwInput}{Input}
\SetKwInput{kwOutput}{Output}
\SetKwInput{kwReturn}{Return}
\SetKw{Break}{break}

\begin{algorithm}
\small
\caption{EntropyStop: An automated training stopping algorithm}
\label{alg:online}
\kwInput{ Model $M$ with  learnable parameters $\Theta$, patience parameter $k$, downtrend threshold  $R_{down}$, dataset $D$,  iterations T, evaluation set size $N_{eval}$ }
\kwOutput{Outlier score list $\textbf{O}$}
Initialize  the parameter $\Theta$ of Model $M$\; Random sample $N_{eval}$ instances from  $D$ as the evaulation set $D_{eval}$ \;
$G \gets 0;$  $patience \gets 0; $ $ \Theta_{best} \gets \Theta; $ \; 

  $e_0 \gets entropy^{loss}(\{\mathcal{J}(\textbf{x}_i;\Theta), \textbf{x}_i \in D_{eval}\});$ $e^{min} \gets e_0$\;
  \Comment*[r]{Model Training}
\For {$j:= 1 \rightarrow T$}{
    Random sample a batch of training  data $D_{batch}$ \;
    Calculate $\mathcal{L}_{train} =  \mathcal{J}(D_{batch};\Theta) + \mathcal{R}(D_{batch};\Theta)$ \;
    Optimize the parameters $\Theta$  by minimizing $\mathcal{L}_{train}$\;
  $ e_j \gets entropy^{loss}(\{\mathcal{J}(\textbf{x}_i;\Theta), \textbf{x}_i \in D_{eval}\})$\;
    $G \gets G + |e_j - e_{j-1}|$\;
\eIf{
$e_j < e^{min}$ and $\frac{e^{min}-e_j}{G} > R_{down}$
}{
$e^{min} \gets e_j;$ $G \gets 0;$  $patience \gets 0;$ $ \Theta_{best} \gets \Theta; $
}{
$patience \gets patience + 1$\;
}
\If{$patience = k$}{\Break}
}
\kwReturn{$M(D;\Theta_{best})$}
\end{algorithm}

Two new parameters are introduced, namely $k$ and $R_{down}$. $k$ represents the patience for searching the optimal iteration, with larger value improving accuracy at the expense of longer training time. Then, $R_{down}$ sets the requirement for the significance of downtrend. Apart from these two parameters, learning rate is also critical as it can significantly impact the training time. We recommend setting $R_{down}$ within the range of $[0.01,0.1]$, while the optimal value of $k$ and learning rate is associated with the actual entropy curve. We provide a guidance on tuning these HPs in Appx \ref{appx:guide-for-tuning}.

\textbf{Time Complexity.} The time complexity for each iteration is $O(\mathcal{J}(D_{eval};\Theta)) + O(|D_{eval}|)$, including the computation of $\{\mathcal{J}(\textbf{x}_i;\Theta), \textbf{x}_i \in D_{eval}\}$ and $entropy(\mathcal{P}^{loss}_{eval})$.

\subsection{Inapplicability}
Our metric and algorithm are based on  inlier priority, but it is crucial to acknowledge that outliers may exihibit similar patterns to inliers. For instance, local outliers may exist in the peripheral region of the normal class, resulting in gradient directions that are comparable to those of normal data points. To study this, we conducted a  investigation of various outlier types in Sec \ref{sec-further-invest}, which indicates that our approach has limited effectiveness for local outliers. Nevertheless, for other types of outliers, such as cluster and global outliers, our approach can significantly boost performance by halting training at the optimal iteration. 

Our method cannot be combined with pre-training techniques. As the learnable parameters $\Theta$ are significantly influenced by pre-training, the entropy curves  during training may be inaccurate. 

\section{Experiments}
\label{sec:exp}
% \textbf{Overview.} 
% In this section, our objective is to demonstrate the effectiveness and applicability of our approach. Firstly, we conduct a comprehensive analysis of the hyperparameter(s) (HP) sensitivity and utmost potential of deep models to showcase the motivation and efficacy of our method. Secondly, we investigate the effectiveness of our approach in detecting various types of outliers under varying outlier ratios to gain a better understanding of its application scope. Lastly, we investigate the sensitivity of our approach for \textit{batch size} and $N_{eval}$. The experiments are all conducted under the Polluted (i.e., transductive) setting where  the train data is the same as the test data, containing the inliers as well as outliers.

\textbf{Evaluation Metrics.} We evaluate the detection performance by AUC \cite{auc}.We adopt $Score_{AUC}$ \cite{dcfod} to indicate the aggregated scores across all datasets:
\[
\small
    Score_{AUC}(A^{(i)}) = \frac{1}{j}\sum_j \frac{AUC(A^{(i)},D^{(j)})}{
    max_p AUC(A^{(p)},D^{(j)})}
\]
where $A^{(i)}$,$D^{(j)}$ represent the $i^{th}$ algorithm and the $j^{th}$ dataset, respectively.

\textbf{Computing Infrastructures.} All experiments are conducted on a computer with Ubuntu 18.04 OS, AMD Ryzen 9 5900X CPU, 62GB memory, and an RTX A5000 (24GB GPU memory) GPU.

% The objective of this experiment is twofold. Firstly, we aim to demonstrate the crucial role of training time in mitigating the HP sensitivity issue. Secondly, we investigate the effectiveness of our approach compared to Unsupervised Outlier Model Selection (UOMS) \cite{Internal-evaluation-paper} solutions and Ensemble solutions. 
% The former selects a model from a pool of trained models, while the latter trains only one ensemble model for any dataset.

\subsection{TestBed}
\label{sec-hp-testbed}
\begin{wraptable}{r}{4.5cm}
  \centering
  \scriptsize
  \caption{Dataset statistics}
  \setlength{\tabcolsep}{0.5mm}{
  
    \begin{tabular}{ccccc}
    \toprule
    \textbf{Name}  & \textbf{Type}  & \textbf{\#} \textbf{Pts} &\textbf{ Dim}   & \textbf{\% Outl} \\
    \midrule
    Ionosphere & tabular & 351   & 32    & 35.9 \\
    letter & tabular & 1600  & 32    & 6.25 \\
    vowels & tabular & 1456  & 12    & 3.43 \\
    \midrule
    MNIST-3 & image & 7357  & $28\times28$ & 16.6 \\
    MNIST-5 & image & 6505  & $28\times28$ & 16.6 \\
    \bottomrule
    \end{tabular}%
    }
  \label{tab:Dataset-statistics}%
  \vspace{-0.6cm}
\end{wraptable}%
\textbf{Datasets.} Experiments are based on 3 tabular datasets from ADbenchmark\footnote{https://github.com/Minqi824/ADBench/tree/main/datasets/Classical} \cite{ADbench}  as well as 2 image datasets from MNIST (see Table \ref{tab:Dataset-statistics}). For image datasets, we pick one class as the inliers and downsample the other classes to constitute the outliers. Details on dataset description and preparation can be found in Appx. \ref{appx:dataset-desc-for-hp}.

\textbf{Models.} We analyze the sensitivity of HPs for  AE \cite{AE}, RDP \cite{RDP},  DeepSVDD \cite{deep-svdd}. We list the studied HPs  in Table \ref{tab:model-hp}, and provide detailed descriptions in Appx. \ref{appx:hp-config}. We define a grid of 2-3 values for each HP and train each deep OD method with all possible combinations. To evaluate the utmost potential of model during its training, we evaluate its AUC on the whole dataset after each training iteration while the loss entropy is computed on a pre-sampled evaluation set at the same time.  

\begin{table}[htbp]
    \small
  \centering
  \caption{Deep OD  for studying HP sensitivity. (See Appx. \ref{appx:hp-config} for list of grid values per HP.)}
    \begin{tabular}{lllr}
    \toprule
    \textbf{Method} & \textbf{Type}  &\textbf{List of HPs} & \textbf{\# models} \\
    \midrule
    AE \cite{AE}  & tabluar \& image  & act\_func $\cdot$ dropout, h\_dim $\cdot$  lr $\cdot$  layers $\cdot$ epoch & 64 \\
    RDP \cite{RDP} & tabluar   & out\_c $\cdot$ lr $\cdot$ dropout $\cdot$ filter $\cdot$ epoch & 72 \\
    DeepSVDD \cite{deep-svdd} & image & conv\_dim $\cdot$ fc\_dim $\cdot$ relu\_slope $\cdot$ epoch,lr $\cdot$ wght\_dc & 64 \\
    % GLAM \cite{glam} & graph & lr $\cdot$ wght\_dc $\cdot$ epoch $\cdot$ layers $\cdot$ h\_dim & 72 \\
    \bottomrule
    \end{tabular}%
  \label{tab:model-hp}%
\end{table}%

\begin{wrapfigure}{0}{0.35\textwidth}
\vspace{-0.5cm}
  \begin{center}
\includegraphics[width=1.0\textwidth]{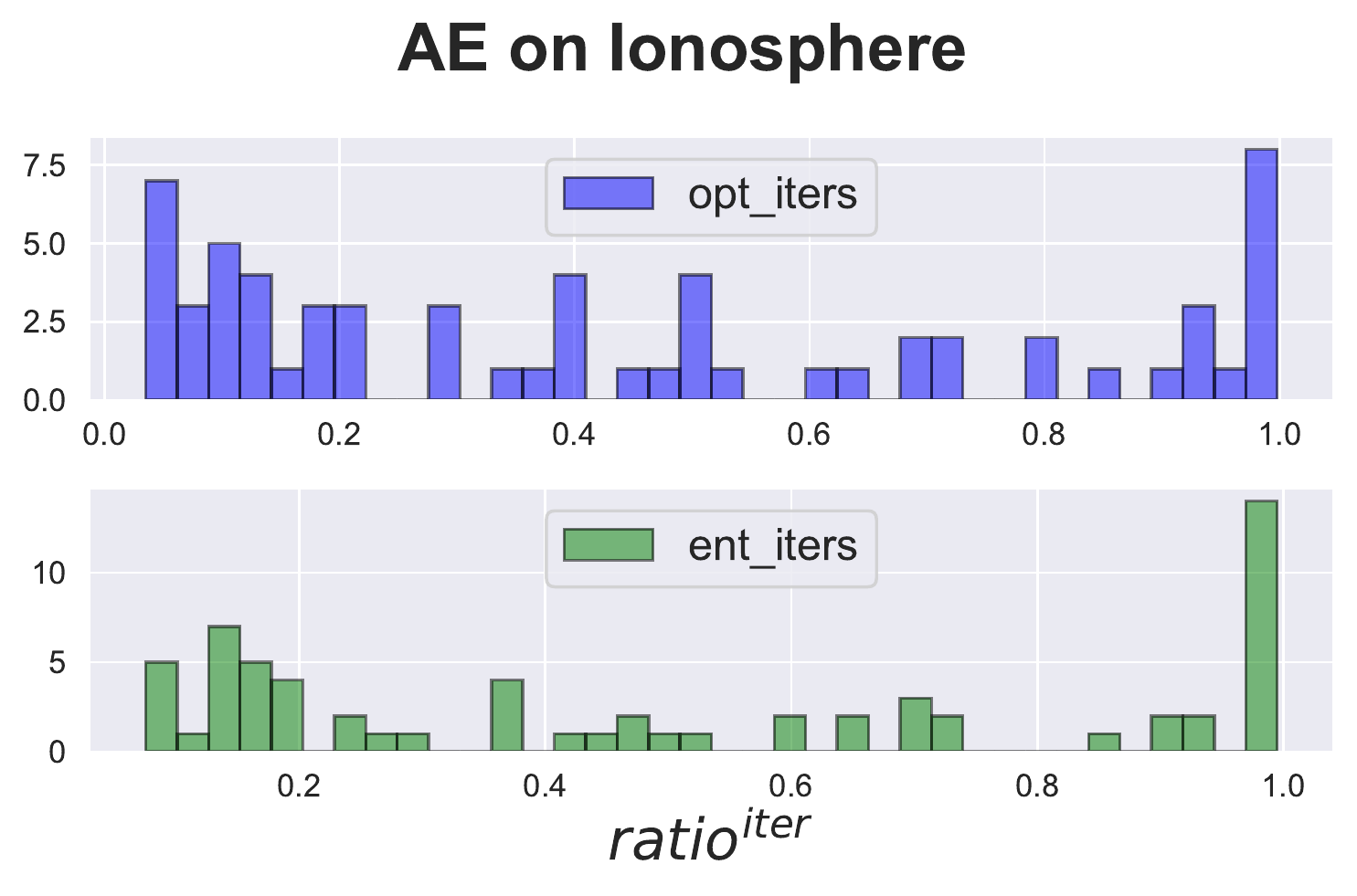}
  \end{center}
  \caption{Iterations selected by $Optimal$ and $Entropy$ accross 64 HP configs. The value is normalized by iterations of Naive training.}
  \label{fig:auc-iteration}
\end{wrapfigure}
\textbf{Baselines.} We report the end-of-training AUC and the  optimal AUC during training as \textit{Naive} and \textit{Optimal}, respectively. We denote the AUC obtained by our \textit{EntropyStop}  as \textit{Entropy}. 
For UOMS solutions,
Xie-Beni index (XB) \cite{xb}, ModelCentrality (MC) \cite{MC}, and HITS
 \cite{Internal-evaluation-paper} are the baselines for comparison.
For ensemble solutions, we select the deep hyper-ensemble AE model ROBOD \cite{robod} and the SOTA traditional detector IsolationForest (IF) \cite{IsolationForest}.
Note that \textit{Naive}, \textit{Optimal}, and \textit{Entropy} have AUC scores  for each HP configuration, whereas the UOMS solutions only produce a single AUC score for a given dataset and algorithm  pair. For \textit{EntropyStop}, we set $R_{down}$ to 0.1 while $k$ varies according to  model and dataset. Since ROBOD is essentially an AE ensemble, we use different layers, hidden units as well as random seeds  to create an ensemble of AE models while analyzing its performance on the remaining HPs of AE. We keep  IF at a default setting. More details are available in Appx. \ref{appx:hp-config}. 

\begin{figure*}
  \centering
  \includegraphics[width=130.00pt]{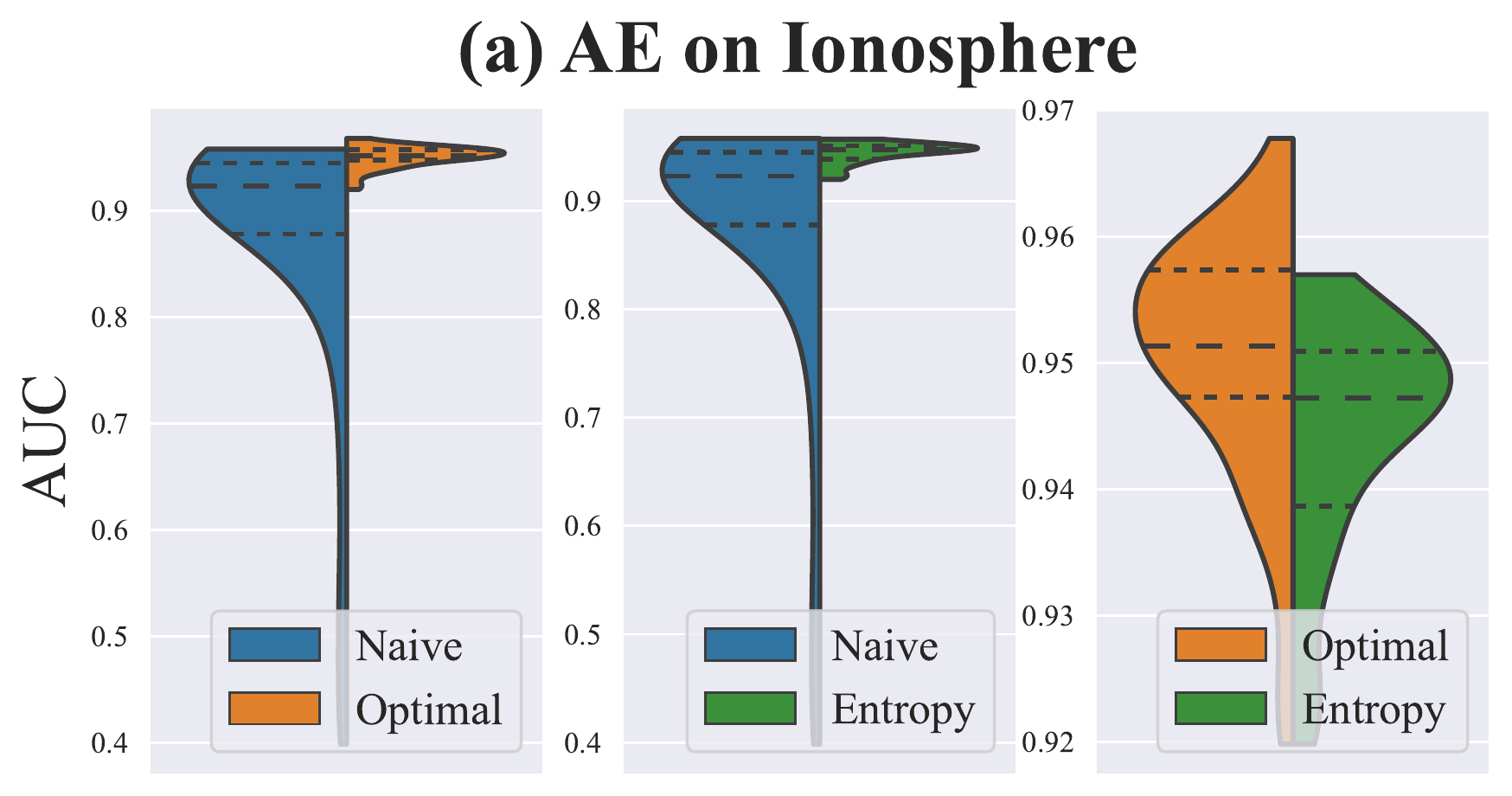}
  \includegraphics[width=130.00pt]{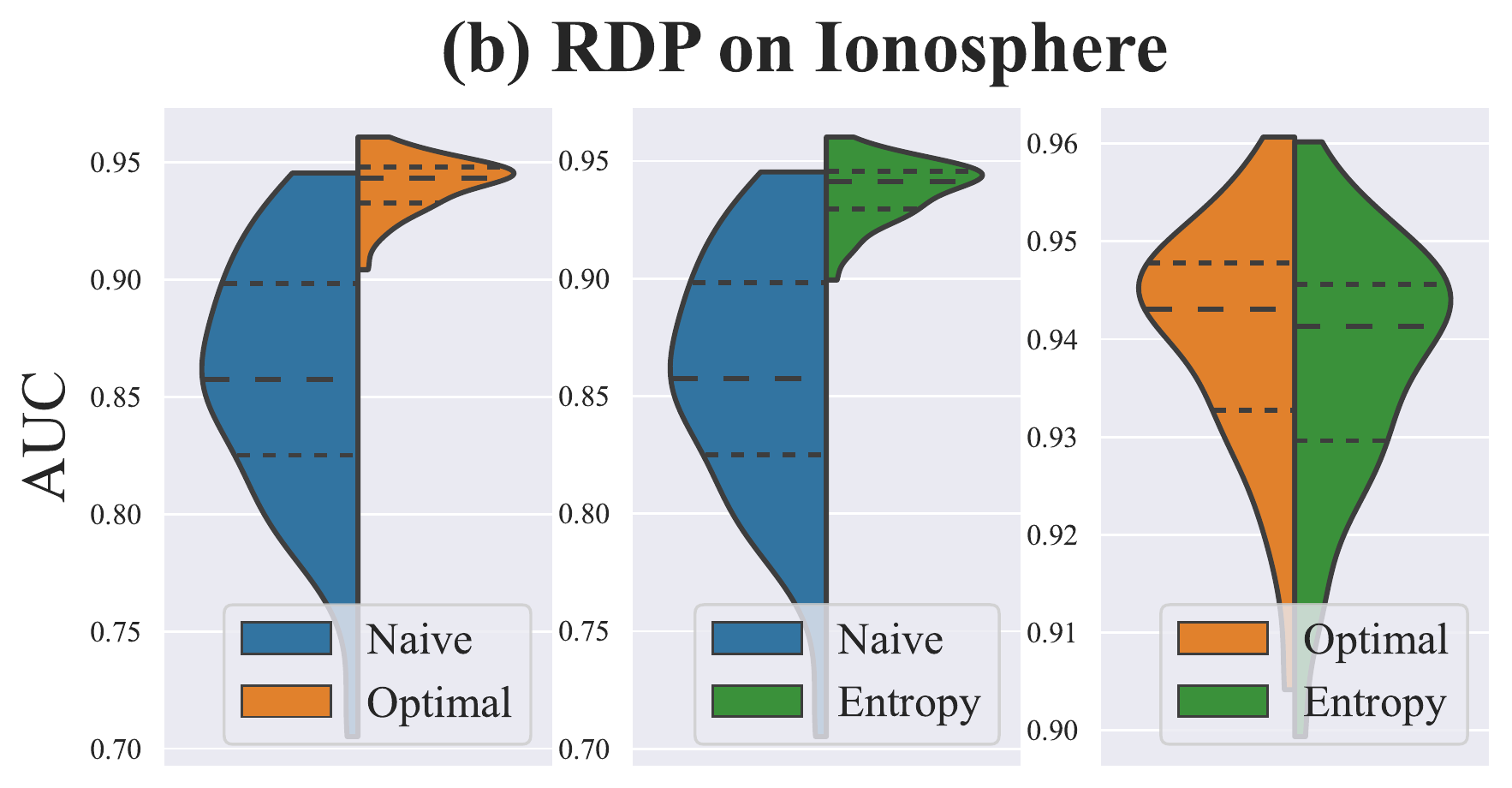}
  \includegraphics[width=130.00pt]{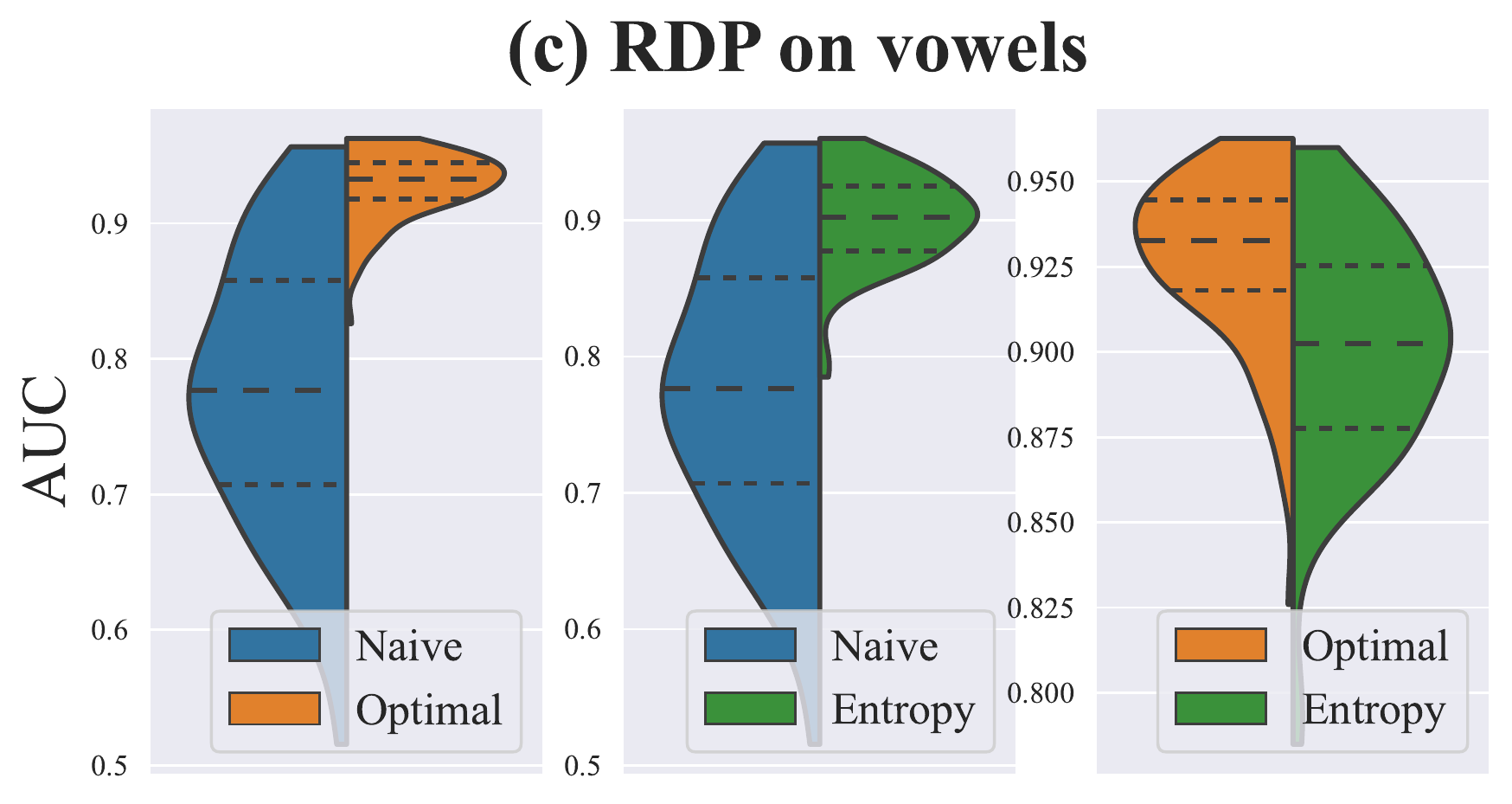}
  \includegraphics[width=130.00pt]{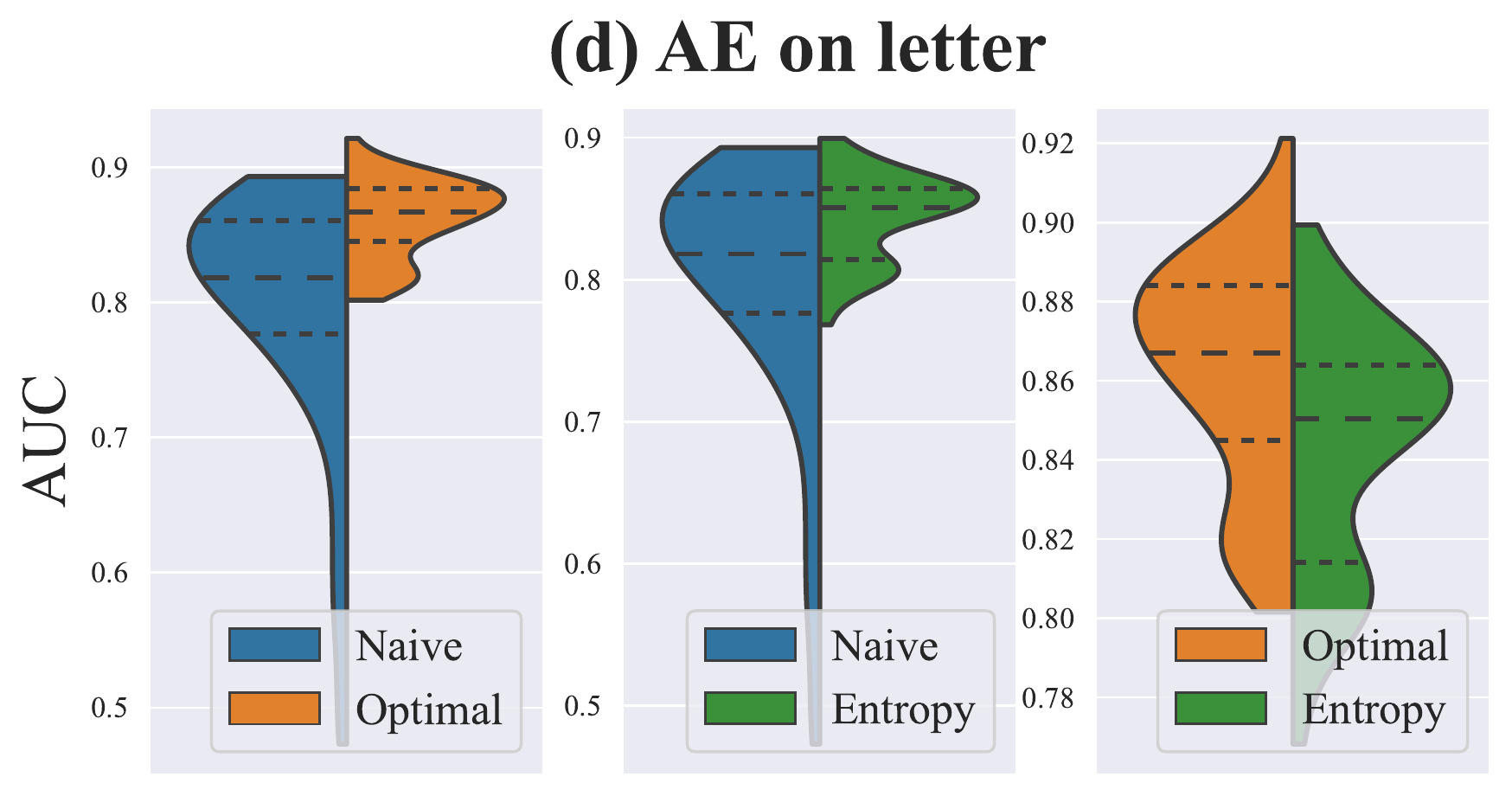}
  \includegraphics[width=130.00pt]{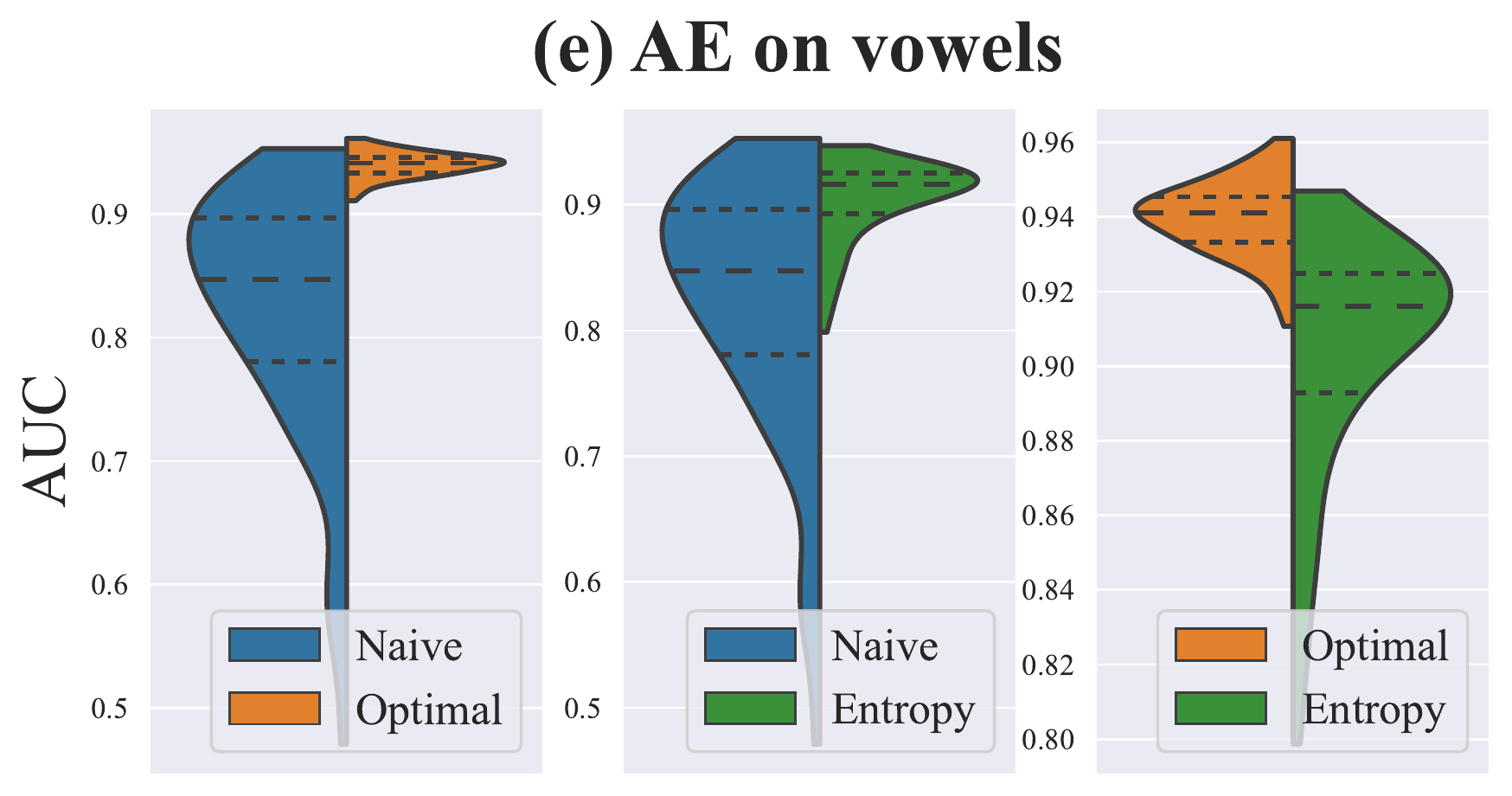}
  \includegraphics[width=130.00pt]{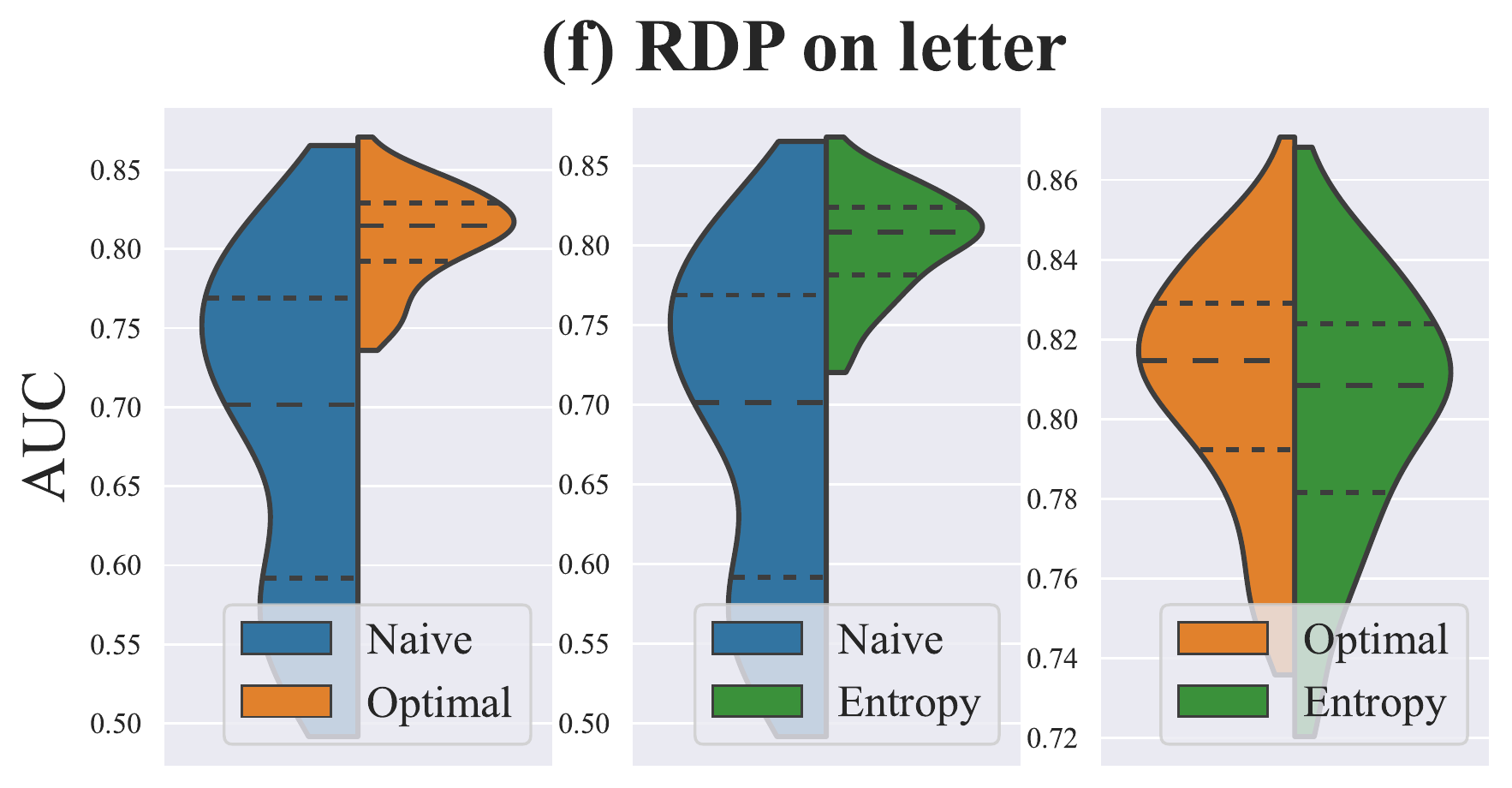}
  \caption{Comparison of AUC distribution  of \textit{Naive}, \textit{Optimal}, and \textit{Entropy} across varing HP configurations. Our \textit{Entropy} achieves a similar effect as \textit{Optimal} \textbf{without the  ground-truth labels} in (a), (b),(c),(d) and (f) (totally 344 different trainings). Remaining figures in Fig \ref{Fig:appx-hp-auc-distribution} Appx. \ref{appx:additonal-fig}.}
  \label{Fig:HP-sensivity-exp} 
\end{figure*}

\subsection{Results}

The experiments are conducted  three times and the average results are reported in Table \ref{tab:hp-sen-exp}.
\textbf{Q1: Does training time significantly impact the model’s performance?} For training AE/RDP on tabular datasets (see Fig \ref{Fig:HP-sensivity-exp}), by solely attaining the optimal performance during training, the HP sensitivity issue is greatly mitigated, revealing that training time is the primary factor to cause HP sensitivity.
For training DeepSVDD on  MNIST (see Fig \ref{Fig:appx-hp-auc-distribution} in Appx. \ref{appx:additonal-fig}), due to the existence of other sensitive HPs, \textit{Optimal} only has a slight reduction in AUC variation,  but it greatly improves the overall AUC.
These results demonstrate the significant  impact of training time on model's performance. In Fig \ref{fig:auc-iteration}, the optimal AUC iteration has a diverse distribution, indicating that the model with different HPs requires varying  training iterations to achieve its optimal performance.

\begin{wraptable}{r}{5.2cm}

\scriptsize
  \centering
  \caption{Expected training time on each dataset, normalized by  training time of \textit{Naive} AE. (Smaller is better)}
   \setlength{\tabcolsep}{0.45mm}{
    \begin{tabular}{lcccc}
    \toprule
    \multicolumn{1}{l}{\textbf{Dataset}} & \multicolumn{1}{l}{\textbf{EntropyAE}} & \multicolumn{1}{l}{\textbf{ROBOD}} & \multicolumn{1}{l}{\textbf{UOMS(AE)}} & \multicolumn{1}{l}{\textbf{IF}} \\
    \midrule
    Ionosphere & 0.39  & 4.16  & 64.00  & 0.13  \\
    letter & 0.27  & 4.15  & 64.00  & 0.04  \\
    vowels & 0.15  & 4.08  & 64.00  & 0.04  \\
    MNIST-3 & 0.04  & 4.79  & 64.00  & 0.01  \\
    MNIST-5 & 0.05  & 4.80  & 64.00  & 0.01  \\
    \bottomrule
    \end{tabular}%
    }
  \label{tab:ae-time-saving}%
   \vspace{-0.2cm}
\end{wraptable}
\textbf{Q2: Does our approach can achieve a similar effect to \textit{Optimal}?}
For RDP, \textit{Entropy} achieves a surprisingly similar effect to \textit{Optimal} on 3 datasets with 72 diverse HP configs (totally 216 trainings) in the absence of labels, highlighting its effectiveness.
For AE, the similar outstanding  effect is observed on Ionosphere and letter (totally 128 trainings). Though there is a gap between \textit{Optimal} and \textit{Entropy} on vowels for AE, \textit{Entropy} still significantly reduces the HP sensitivity of \textit{Naive} and improve the overall AUC.  We acknowledge  its limitation in capturing the optimal performance of AE on MNIST. However, more than 95\% of training time is saved on MNIST due to the early detection of training convergence, as shown in Table \ref{tab:ae-time-saving}.
For DeepSVDD, \textit{Entropy} also achieves a similar effect to \textit{Optimal}.
Overall, \textit{Entropy} improves the AUC of \textit{Naive} on all models and datasets, as shown in Table \ref{tab:hp-sen-exp}, while save 61-96\% training time according to Table \ref{tab:ae-time-saving}.  The similar time-saving results  are also observed in RDP and DeepSVDD. In fact, it averagely saves more time on a larger dataset as the same training epochs will result in more training iterations under a fixed batch size.
In Fig \ref{fig:auc-iteration}, it shows that \textit{Entropy} adopts to different HP configurations to select the optimal iterations. 
\begin{table}[htbp]
\scriptsize
  \centering
  \caption{AUC performance of baselines for Unsupervised OD. 
Highlighted in red bold and blue italics are the \textcolor[rgb]{ 1,  0,  0}{\textbf{best}} and \textcolor[rgb]{ 0,  .439,  .753}{\textit{runner-up}}. Standard deviation of \textit{Naive}, \textit{Entropy} and ROBOD denotes performance variance across  HP configs while that of UOMS (i.e. XB, MC and HITS) and IF denotes the performance variance across different random seeds.}
   \setlength{\tabcolsep}{1.05mm}{
    \begin{tabular}{ccccccccc}
    \toprule
    \textbf{Dataset} & \textbf{Model} & \textbf{Naive} & \textbf{Entropy (Ours)} & \textbf{XB \cite{Internal-evaluation-paper}} & \textbf{MC  \cite{Internal-evaluation-paper}} & \textbf{HITS  \cite{Internal-evaluation-paper}} & \textbf{ROBOD \cite{robod}} & \textbf{IF \cite{IsolationForest}} \\
    \midrule
    Ionosphere & AE    & 0.878$\pm$0.123 & \textcolor[rgb]{ 0,  .439,  .753}{\textit{0.944$\pm$0.009}} & 0.884$\pm$0.056 & \textcolor[rgb]{ 1,  0,  0}{\textbf{0.946$\pm$0.004}} & 0.939$\pm$0.006 & 0.919$\pm$0.028 & 0.853$\pm$0.008 \\
          & RDP \cite{RDP}  & 0.859$\pm$0.053 & \textcolor[rgb]{ 1,  0,  0}{\textbf{0.938$\pm$0.013}} & \textcolor[rgb]{ 0,  .439,  .753}{\textit{0.871$\pm$0.054}} & 0.810$\pm$0.039 & 0.856$\pm$0.004 &       &  \\
    \midrule
    letter & AE    & 0.799$\pm$0.091 & 0.838$\pm$0.032 & 0.770$\pm$0.005 & \textcolor[rgb]{ 0,  .439,  .753}{\textit{0.868$\pm$0.007}} & \textcolor[rgb]{ 1,  0,  0}{\textbf{0.870$\pm$0.005}} & 0.815$\pm$0.037 & 0.616$\pm$0.039 \\
          & RDP   & 0.689$\pm$0.103 & \textcolor[rgb]{ 1,  0,  0}{\textbf{0.801$\pm$0.032}} & 0.601$\pm$0.106 & 0.597$\pm$0.048 & \textcolor[rgb]{ 0,  .439,  .753}{\textit{0.72$\pm$0.027}} &       &  \\
    \midrule
    vowels & AE    & 0.824$\pm$0.101 & \textcolor[rgb]{ 0,  .439,  .753}{\textit{0.902$\pm$0.033}} & 0.701$\pm$0.184 & \textcolor[rgb]{ 1,  0,  0}{\textbf{0.912$\pm$0.006}} & 0.899$\pm$0.015 & 0.872$\pm$0.034 & 0.749$\pm$0.008 \\
          & RDP   & 0.779$\pm$0.099 & \textcolor[rgb]{ 1,  0,  0}{\textbf{0.901$\pm$0.034}} & 0.669$\pm$0.054 & 0.731$\pm$0.048 & \textcolor[rgb]{ 0,  .439,  .753}{\textit{0.791$\pm$0.028}} &       &  \\
    \midrule
    MNIST-3 & AE    & \textcolor[rgb]{ 0,  .439,  .753}{\textit{0.826$\pm$0.009}} & \textcolor[rgb]{ 1,  0,  0}{\textbf{0.835$\pm$0.009}} & 0.819$\pm$0.001 & 0.825$\pm$0.002 & 0.822$\pm$0.002 & 0.830$\pm$0.004 & 0.755$\pm$0.021 \\
          & DeepSVDD \cite{deep-svdd} & 0.748$\pm$0.047 & \textcolor[rgb]{ 0,  .439,  .753}{0.805$\pm$0.025} & 0.665$\pm$0.022 & 0.801$\pm$0.014 & \textcolor[rgb]{ 1,  0,  0}{\textbf{0.818$\pm$0.012}} &       &  \\
    \midrule
    MNIST-5 & AE    & 0.758$\pm$0.006 & \textcolor[rgb]{ 0,  .439,  .753}{\textit{0.762$\pm$0.009}} & 0.762$\pm$0.011 & 0.759$\pm$0.002 & 0.755$\pm$0.001 & \textcolor[rgb]{ 1,  0,  0}{\textbf{0.763$\pm$0.009}} & 0.660$\pm$0.008 \\
          & DeepSVDD & 0.684$\pm$0.053 & \textcolor[rgb]{ 1,  0,  0}{\textbf{0.748$\pm$0.042}} & 0.668$\pm$0.007 & 0.725$\pm$0.039 & \textcolor[rgb]{ 0,  .439,  .753}{\textit{0.747$\pm$0.019}} &       &  \\
    \midrule
    \multicolumn{2}{c}{\textbf{Score}} & 0.920  & \textcolor[rgb]{ 1,  0,  0}{\textbf{0.994}} & 0.870  & 0.935  & 0.964  & \textcolor[rgb]{ 0,  .439,  .753}{\textit{0.971}} & 0.840  \\
    \bottomrule
    \end{tabular}%
    }
  \label{tab:hp-sen-exp}%
\end{table}%

\textbf{Q3: How does our approach compare to UOMS solutions?}
Though our approach does not consistently outperform UOMS solutions, it is important to note that the comparison is somewhat unfair.{ \textbf{The entry of \textit{Entropy} in  Table \ref{tab:hp-sen-exp} represents an expected AUC for a model in a random HP configuration, while UOMS (i.e., XB, MC, HITS) selects the ``best`` one from a pool of all models}}, resulting in that their training time is several magnitudes longer than ours (see Table \ref{tab:ae-time-saving}).  Additionally, \textit{Entropy} exhibits superior performance when dealing with RDP that employs random-distance as its outlier score function.  Our goal is to emphasize the significance of training time in HP sensitivity and offer a lightweight solution to mitigate the HP sensitivity issue. By solely determining the optimal  iteration, our approach surpasses UOMS  in terms of overall effectiveness.

\textbf{Q4: how does our approach compare to ensemble solutions?}
Since ROBOD is essentially an AE ensemble, we compare it to EntropyAE (i.e. AE with  \textit{EntropyStop}). As shown in Table \ref{tab:hp-sen-exp}, EntropyAE exceeds ROBOD on  4 datasets (i.e. three tabular datasets and MNIST-3) and takes only 1-10\% training time of ROBOD (see Table \ref{tab:ae-time-saving}). For MNIST-5, they have the same performance (only 0.001 difference in AUC). This result verifies the effectiveness of our  approach in unleashing the potential of AE. In terms of IF, its performance falls far behind the other baselines on these datasets.

\subsection{Further Investigation}
\label{sec-further-invest}
\begin{wraptable}{r}{4.5cm}
\scriptsize
  \centering
  \caption{Effectiveness of \textit{EntropyStop} on heterogeneous outliers at varing outlier ratio. $P_{auc} \leq 0.05$ means the existence of significant improvement. (See detailed results in Appx. \ref{appx:outlier-type-results})}
   \setlength{\tabcolsep}{0.5mm}{
    \begin{tabular}{lccccc}
    \toprule
    \textbf{Type} & \multicolumn{1}{l}{\textbf{Outl}} & \multicolumn{1}{l}{\textbf{Naive}} & \multicolumn{1}{l}{\textbf{Entropy}} & \multicolumn{1}{l}{\textbf{$P_{auc}$}} & \multicolumn{1}{l}{\textbf{time}} \\
    \midrule
    cluster & 0.1   & 0.646  & \textcolor[rgb]{ 1,  0,  0}{\textbf{0.864}} & {\textbf{0.000}} & 0.203  \\
    cluster & 0.4   & 0.559  & \textcolor[rgb]{ 1,  0,  0}{\textbf{0.836}} & {\textbf{0.000}} & 0.168  \\
    \midrule
    global & 0.1  & 0.944  & \textcolor[rgb]{ 1,  0,  0}{\textbf{0.981}} & {\textbf{0.000}}  & 0.479  \\
    global & 0.4  & 0.917  & \textcolor[rgb]{ 1,  0,  0}{\textbf{0.967}} & {\textbf{0.000}} & 0.335  \\
    \midrule
    local & 0.1   & 0.891  & \textcolor[rgb]{ 1,  0,  0}{\textbf{0.901}} & 0.522  & 0.351  \\
    local & 0.4   & \textcolor[rgb]{ 1,  0,  0}{\textbf{0.910}} & 0.900  & 0.998  & 0.390  \\
    \bottomrule
    \end{tabular}%
    }
  \label{tab:outlier-type-study}%
   \vspace{-0.4cm}
\end{wraptable}
\textbf{Applicability and Limitations.} To gain a more comprehensive understanding of our approach, we conduct investigations on its effectiveness on heterogeneous outliers at varying outlier ratios.
We inject local, global, and cluster outliers separately into 19 datasets (see Appx. \ref{appx:dataset-inject}) from \cite{ADbench}, using the default settings of injection algorithms in \cite{ADbench}, with outlier ratios set to 0.1 and 0.4. The main results are presented in Table \ref{tab:outlier-type-study}, along with the p-values obtained from one-sided paired Wilcoxon signed rank tests.
Our findings demonstrate that our approach is highly  effective on cluster outliers, significantly improving performance on  almost all 19 datasets (see Table \ref{tab:0.1-cluster}-\ref{tab:0.4-cluster} in Appx. \ref{appx:outlier-type-results}). For global outliers, our approach also significantly improves performance, albeit to a lesser extent. However, we observed limited effectiveness in detecting local outliers, as they exhibit similar patterns and gradients to those of inliers, leading to the failure of inlier priority.
On the other hand, our method continues to enhance model detection even at an outlier ratio of 0.4, highlighting its robustness to varying outlier ratios. Thus, the type of outlier is more critical for our approach. We recommend that newly developed deep OD methods integrate our approach while verifying the  type of outliers and the discrepancy of outlier gradients from inlier gradients.

\begin{wrapfigure}{r}{0.5\textwidth}
% {0.3\textwidth}
\vspace{-0.4cm}
    \begin{center}
\includegraphics[width=0.5\textwidth]{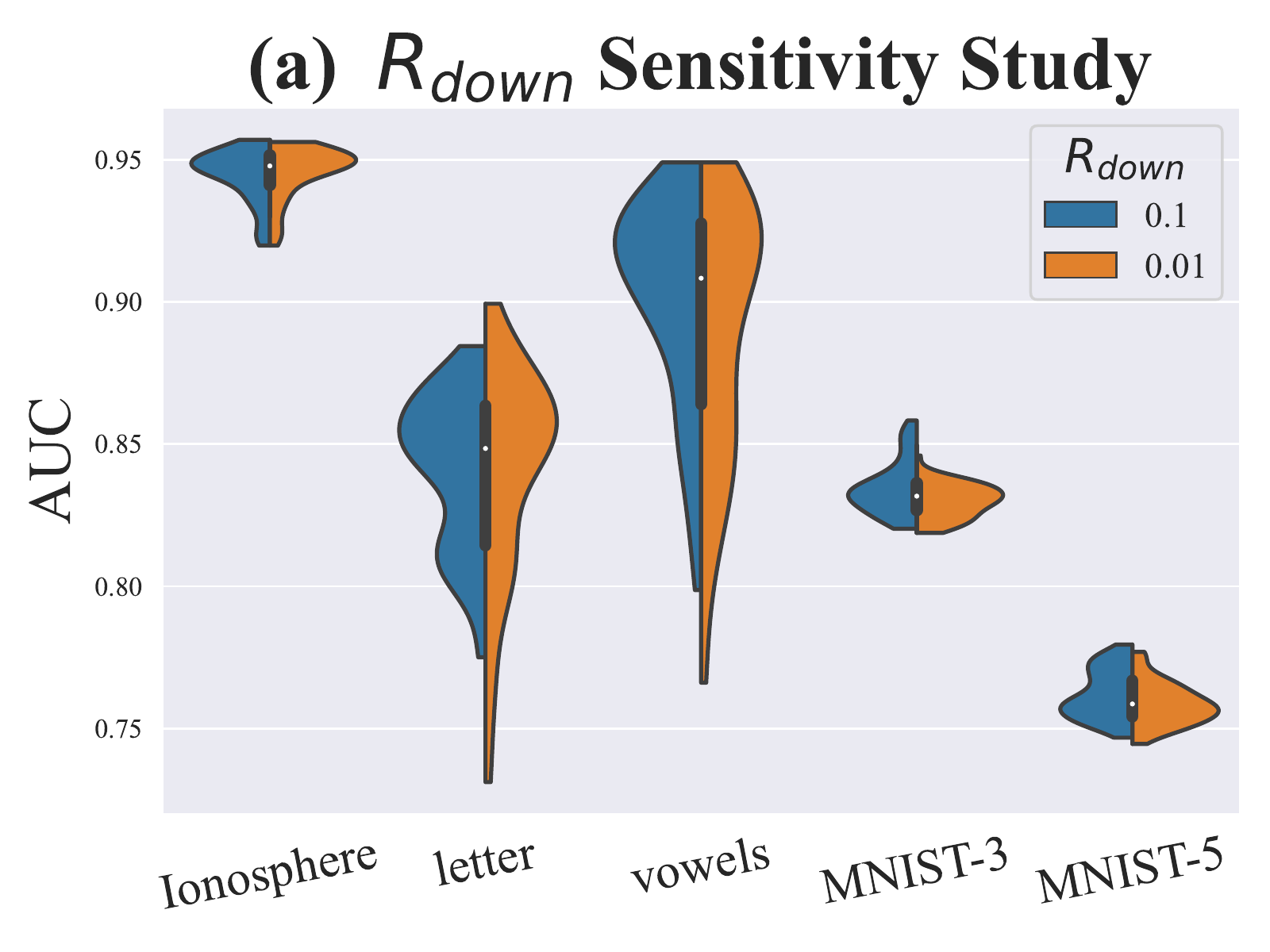}
\includegraphics[width=0.45\textwidth]{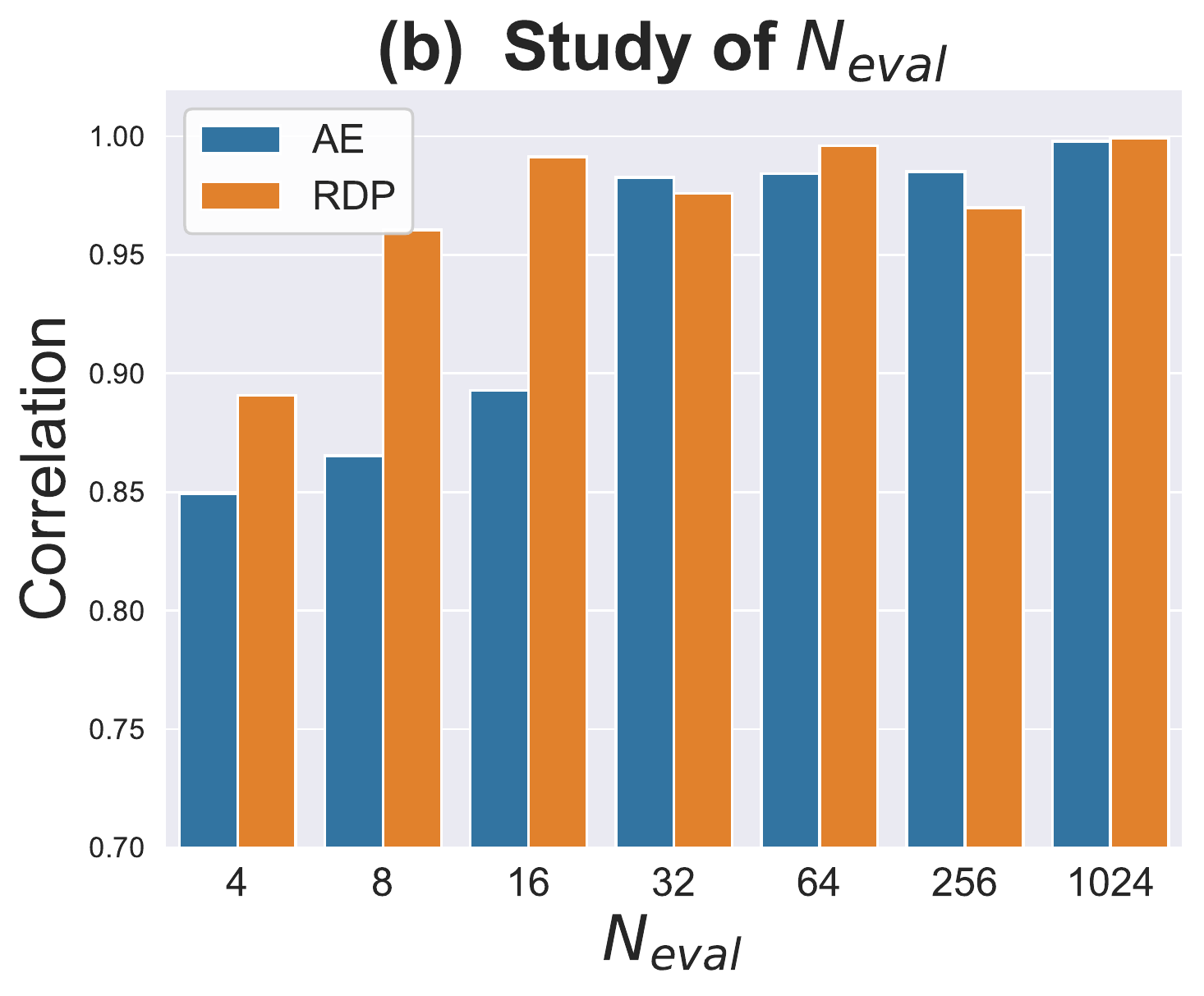}
    \end{center}
    \caption{(a) The AUC distributions of EntropyAE with two different $R_{down}$ on 5 datasets. (b) The pearson correlation between entropy curve computed on the whole dataset and the evaluation set with size  $N_{eval}$.  Fig \ref{Fig:appx-ae-n_eval} in Appx. \ref{appx:additonal-fig}  offers a more intuitive display.}
    \label{fig:para-study}
% \end{wrapfigure}
\end{wrapfigure}
\textbf{Parameter Sensitive Study.} We study the sensitivity of our approach to $R_{down}$ and $N_{eval}$. Firstly, we set $R_{down}$ to 0.1 and 0.01, exploring the corresponding performance of EntropAE (i.e. AE with \textit{EntropyStop}) on 5 datasets in Fig \ref{fig:para-study} (a). It shows that the two AUC distributions of EntropAE  are similar on all datasets, indicating that  $R_{down}$ does not significantly influence the effectiveness. Next, we investigate how large the evaluation set needs to be to ensure accurate loss entropy calculation. To measure accuracy, we compute the pearson correlation of the entropy curve computed on the entire dataset and the evaluation set with size $N_{eval}$. We train  AE and RDP on \textit{vowels} dataset. As depicted in Fig \ref{fig:para-study} (b), when $N_{eval} \geq 32$, the correlation can exceed 0.96, indicating that loss entropy can be accurately computed without the requirement of a large $N_{eval}$.

% \begin{figure}[H]
%   \begin{center}
% \includegraphics[width=0.335\textwidth]{fig/study_n_eval.pdf}
%   \end{center}
%   \caption{}
%   \label{auc-iteration}
% \end{figure}

% \begin{figure}[H]
%   \begin{center}

%   \end{center}
%   \caption{}
%   \label{auc-iteration}
% \end{figure}

 \section{Conclusion}
In this paper, we systematically study the significance of training time in unsupervised OD and propose a novel, efficient metric as well as a training stopping algorithm that can  automatically identify the optimal training iterations  without requiring any labels. Our research demonstrates that hyperparameter (HP) sensitivity in deep OD can be  attributed to training time in certain cases. By halting the training at the optimal moment, the HP sensitivity issue can be greatly mitigated.  We hope our insights would benefit the future research in unsupervised OD.

The limitation of our work is the absence of exhaustive theoretical justification for inlier priority and inadequate analysis for the effectiveness of detecting different types of outliers. However, the main focus of our paper is to emphasize the importance of training time in HP sensitivity. In the future, we are committed to conducting an in-depth investigation of cases where inlier priority does not hold and providing a more comprehensive analysis of detecting heterogeneous outliers.

% \medskip

{
\small
\bibliographystyle{acm}
\bibliography{ref.bib}

}

%%%%%%%%%%%%%%%%%%%%%%%%%%%%%%%%%%%%%%%%%%%%%%%%%%%%%%%%%%%%

\appendix
\renewcommand\thefigure{\Alph{section}\arabic{figure}}    
\setcounter{figure}{0}  
\renewcommand\thetable{\Alph{section}\arabic{table}}    
\setcounter{table}{0}
\renewcommand{\theequation}{\Alph{section}.\arabic{equation}}
\setcounter{equation}{0}

\newpage

\section{Appendix}

\subsection{Guidelines for tuning parameters of  \textit{EntropyStop}}
\label{appx:guide-for-tuning}
In this  section, we provide guidelines on how to tune the hyperparameters (HPs) of EntropyStop when working with unlabeled data. The three key parameters are the learning rate, 
$k$, and $R_{down}$. The learning rate is a crucial factor as it significantly impacts the training time. $k$ represents the patience for finding the optimal iteration, with a larger value improving accuracy but also resulting in a longer training time. $R_{down}$ sets the requirement for the significance of the downtrend.

\begin{figure*}[h]
  \centering
  \includegraphics[width=0.24\textwidth]{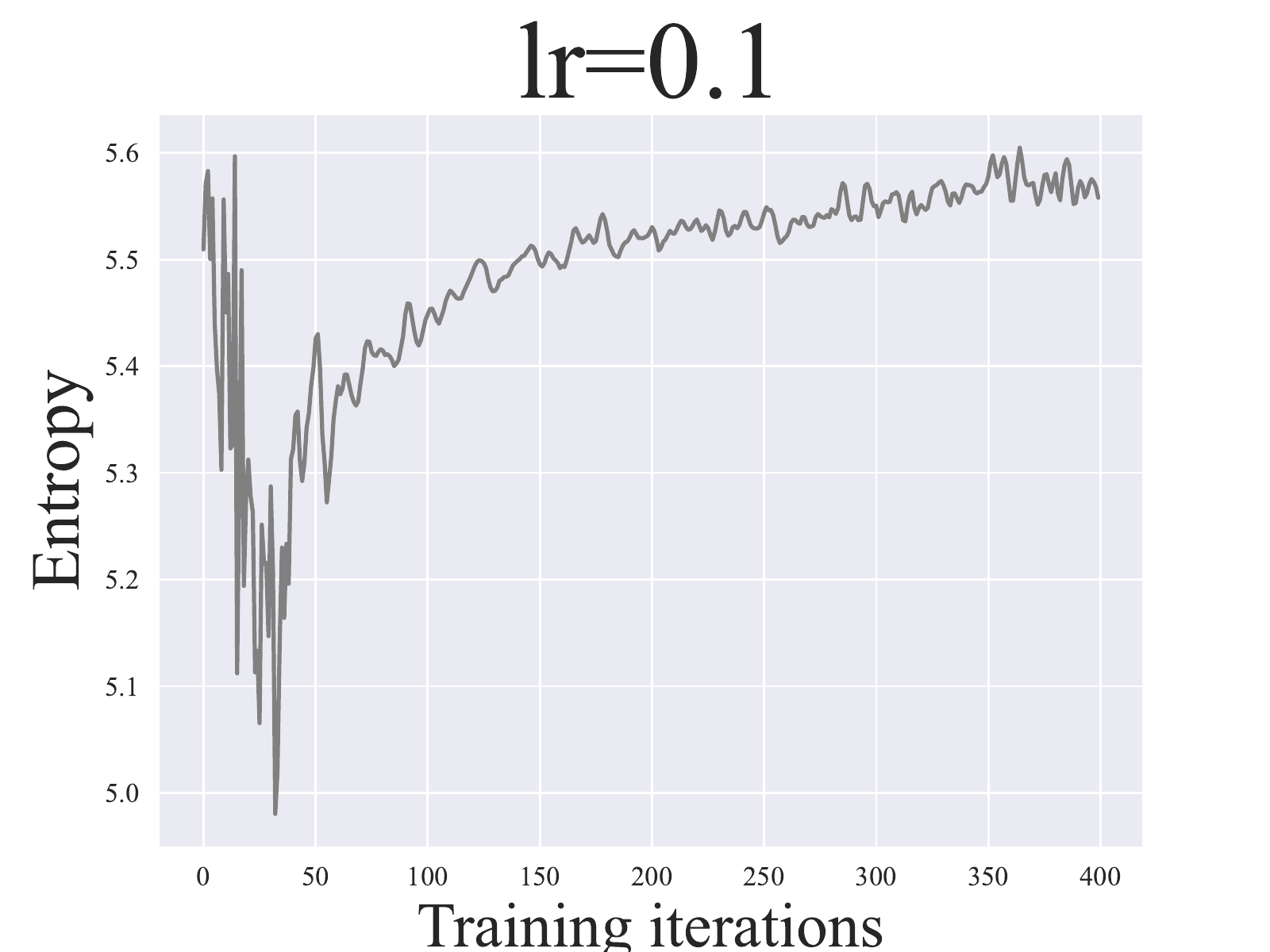}
  \includegraphics[width=0.24\textwidth]{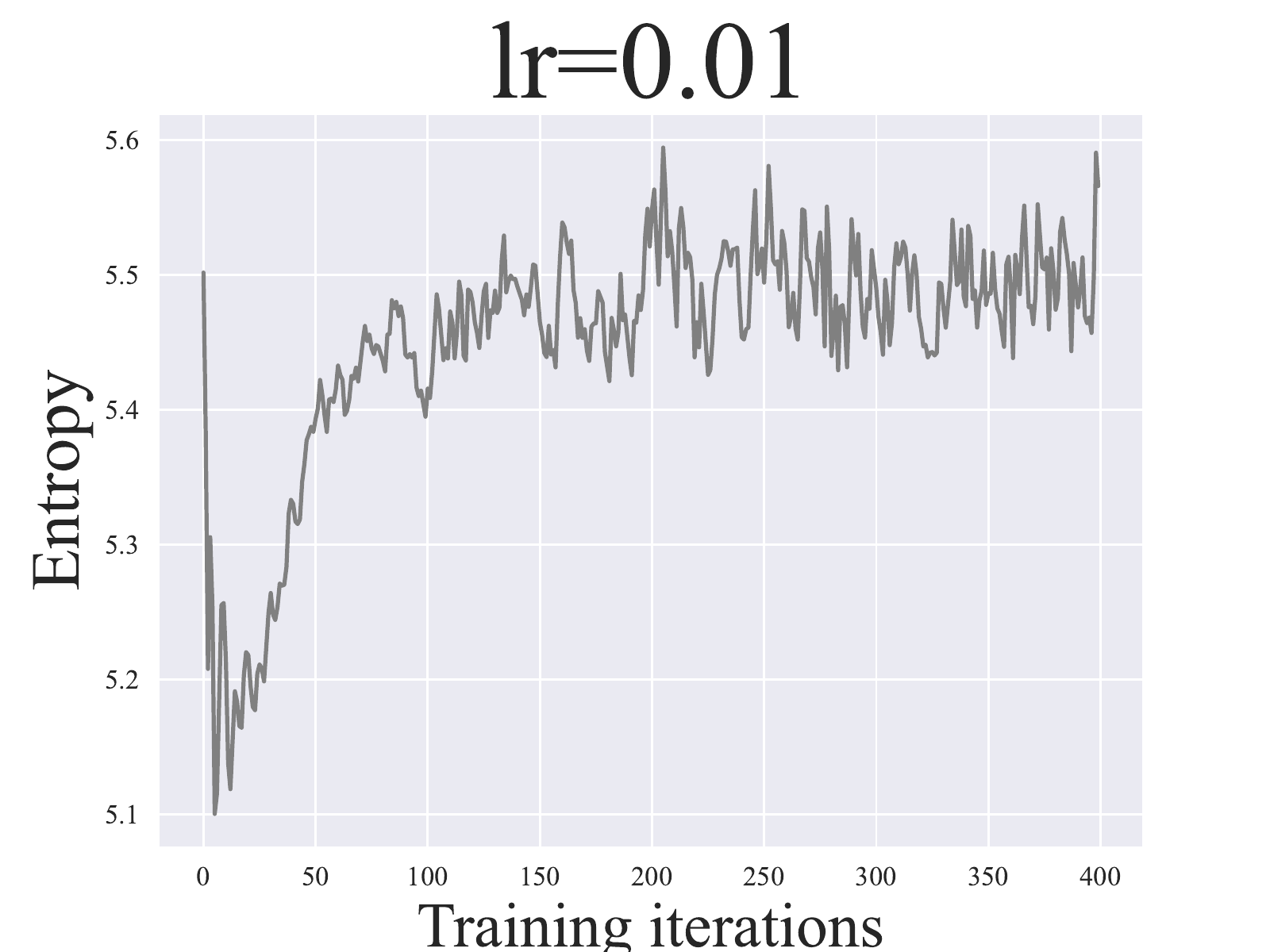}
  \includegraphics[width=0.24\textwidth]{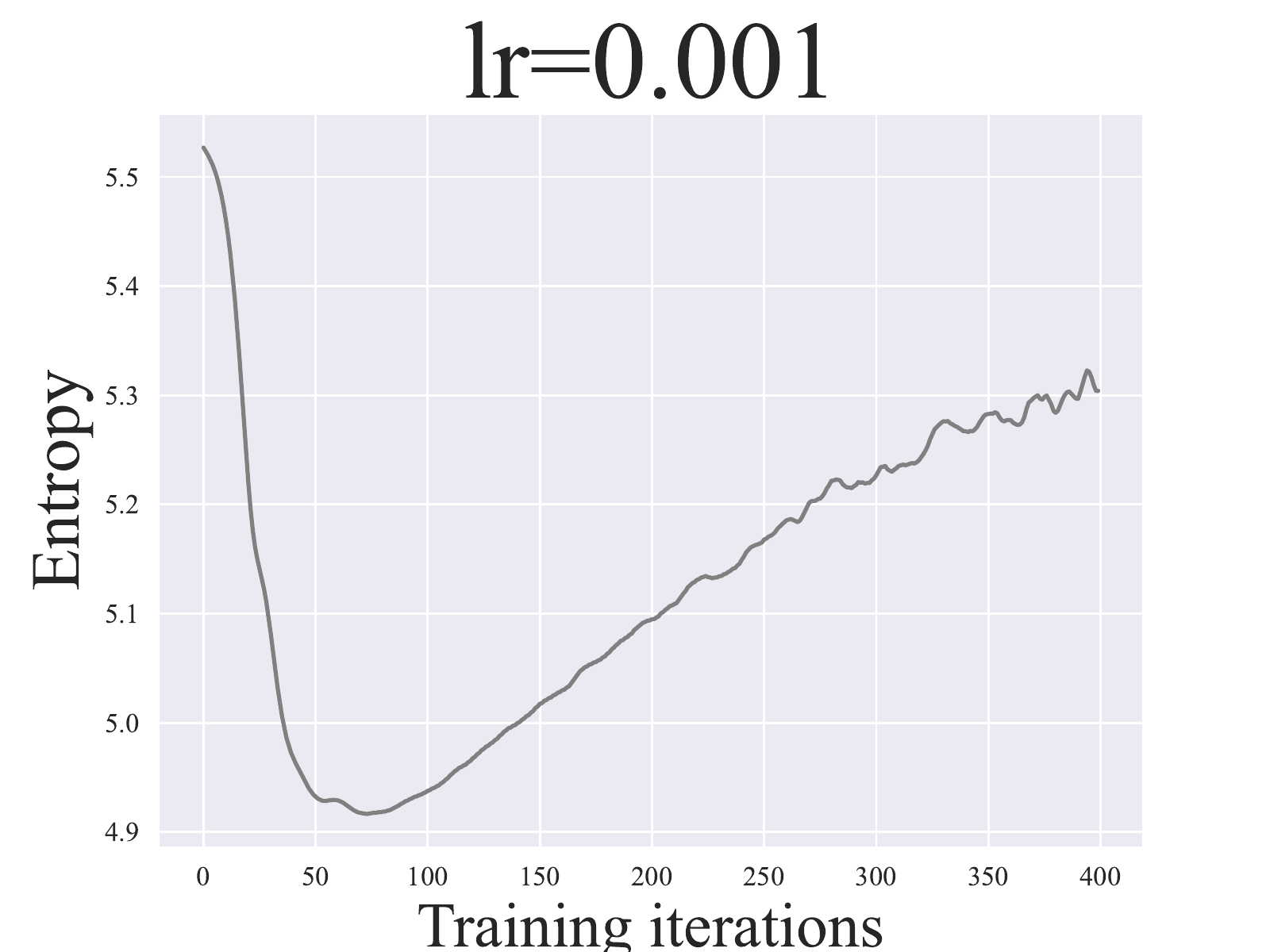}
  \includegraphics[width=0.24\textwidth]{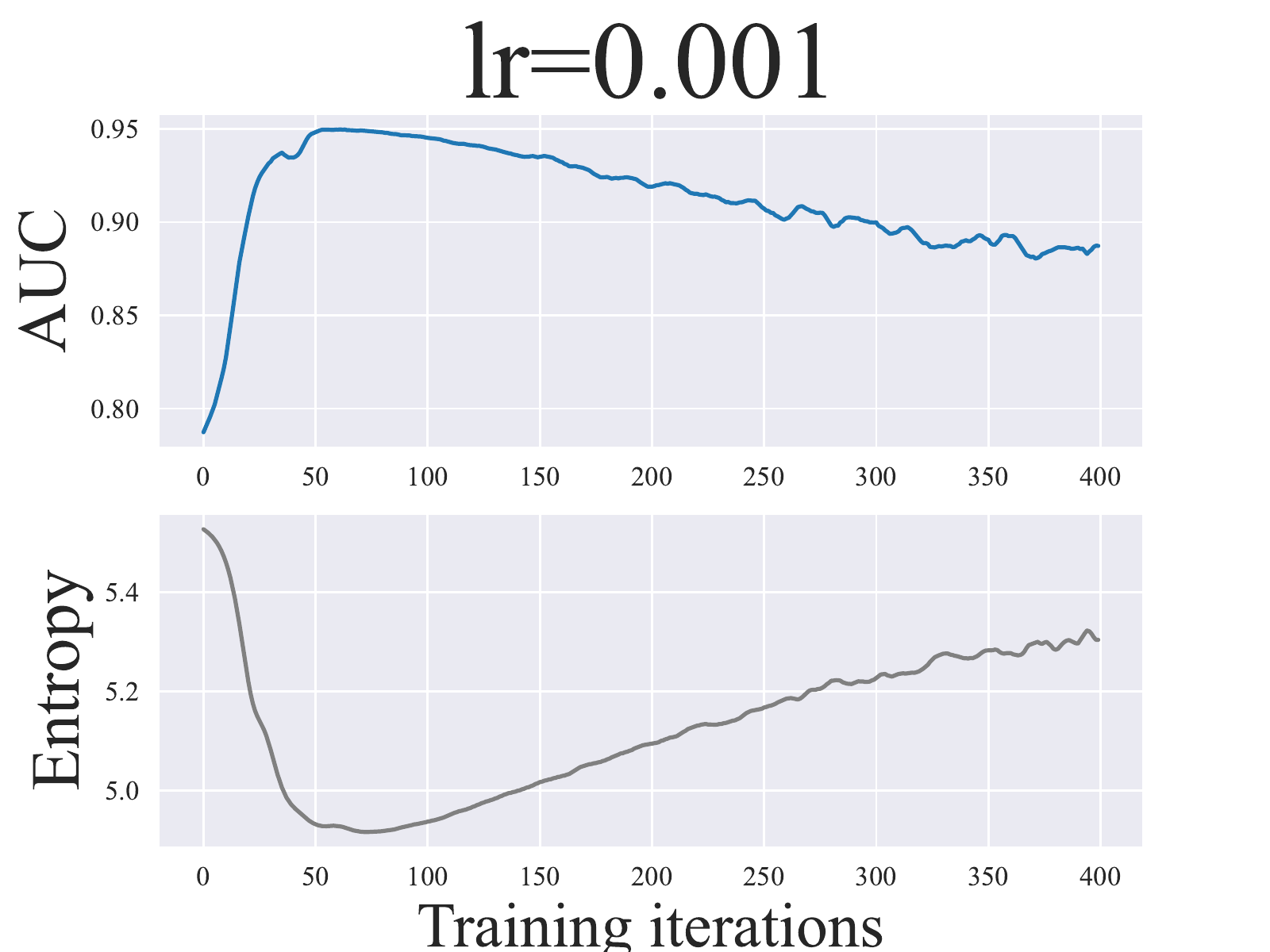}
  \caption{The loss entropy curve of training Autoencoder (AE) on dataset \textit{Ionosphere} with different learning rate.}
  \label{Fig:tune-lr-ae-iono}
\end{figure*}

\textbf{Tuning learning rate.} When tuning these parameters, the learning rate should be the first consideration, as its value will determine the shape of the entropy curve, as shown in Figure \ref{Fig:tune-lr-ae-iono}. For illustration purposes, we first set a large learning rate, such as 0.1, which is too large for  training  autoencoder (AE). This will result in a sharply fluctuating entropy curve, indicating that the learning rate is too large. By reducing the learning rate to 0.01, a less fluctuating curve during the first 50 iterations is obtained, upon which an obvious trend of first falling and then rising can be observed.
Based on the observed entropy curve, we can infer that the training process reaches convergence after approximately 50 iterations. Meanwhile, the optimal iteration for achieving the best performance may occur within the first 25 iterations. However, the overall curve remains somewhat jagged, indicating that the learning rate may need to be further reduced. After reducing the learning rate to 0.001, we observe a significantly smoother curve compared to the previous two, suggesting that the learning rate is now at an appropriate level.

A good practice for tuning learning rate is to begin with a large learning rate to get a overall view of the whole training process while the optimal iteration can be located. For the example in Fig \ref{Fig:tune-lr-ae-iono}, it is large enough to set learning rate to 0.01 for AE model. Then, zoom out the learning rate to obtain a smoother curve and employ EntropyStop to automatically select the optimal iteration.

\begin{figure*}[h]
  \centering
  \includegraphics[width=0.6\textwidth]{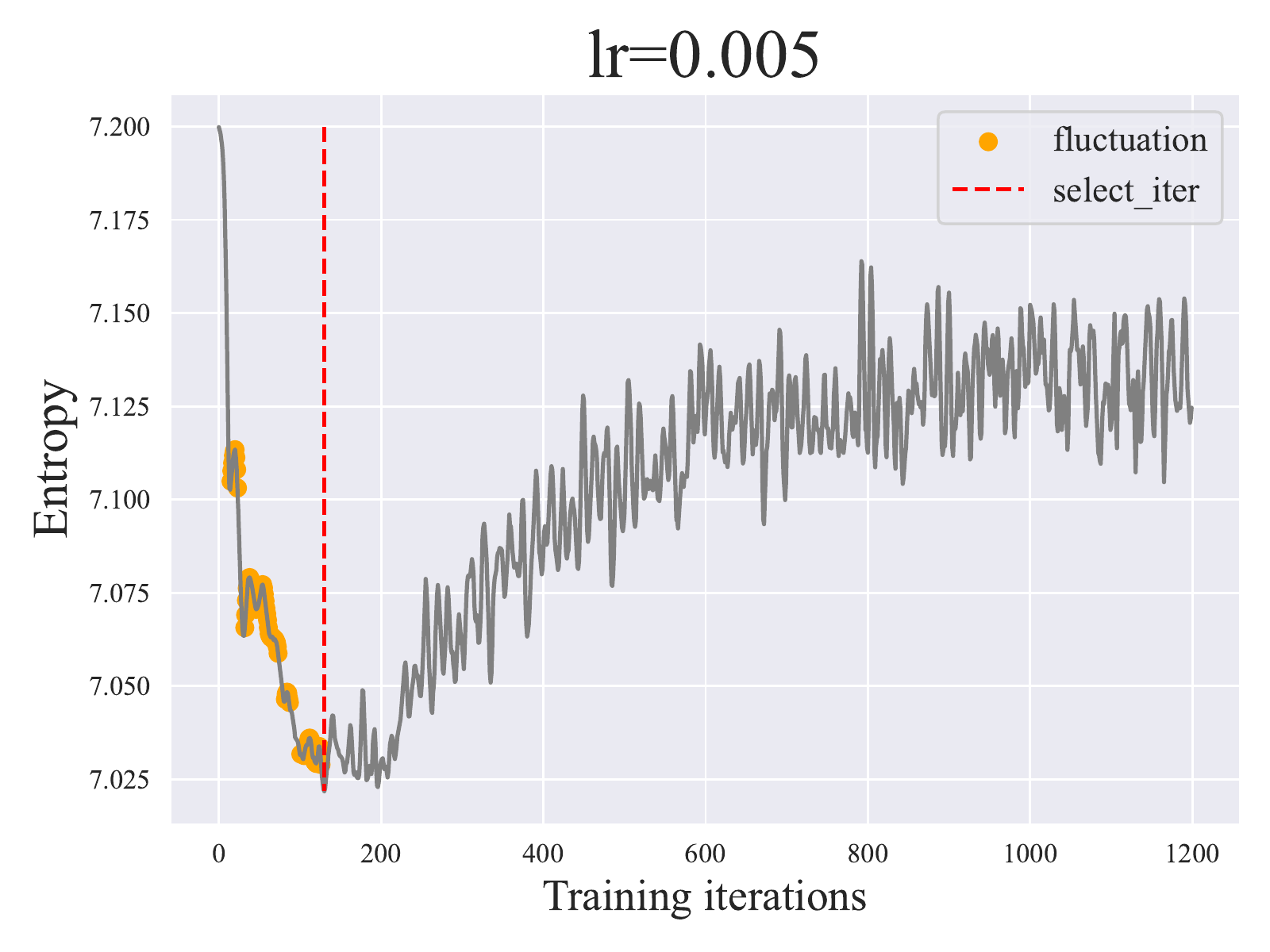}
  \caption{The example of fluctuations (or rises) during the downtrend of entropy curve when training  AE on  dataset \textit{vowels}. The value of $k$ should be set larger than the width of all fluctuations.}
  \label{Fig:fluctuation}
\end{figure*}
\textbf{Tuning $k$ and $R_{down}$.} 
After setting the learning rate, the next step is to tune $k$. If the entropy curve is monotonically decreasing throughout the downtrend, then $k=1$ and $R_{down}=1$ will suffice. However, this is impossible for most cases. Thus, an important role of $k$ and $R_{down}$ is to tolerate the existence of small rise or fluctuation during the downtrend of curve. Essentially, the value of $k$ is determined by the maximum width of the fluctuations or small rises before encounting the opitmal iteration. As shown in Fig \ref{Fig:fluctuation}, the orange color marks the fluctuation area of the curve before our target iteration. The value of $k$ should be set larger than the width of all these fluctuations. For  the example in Fig \ref{Fig:fluctuation}, as long as $k \geq 50$ , EntropyStop can select the target iteration.

\begin{figure*}[h]
  \centering
  \includegraphics[width=0.45\textwidth]{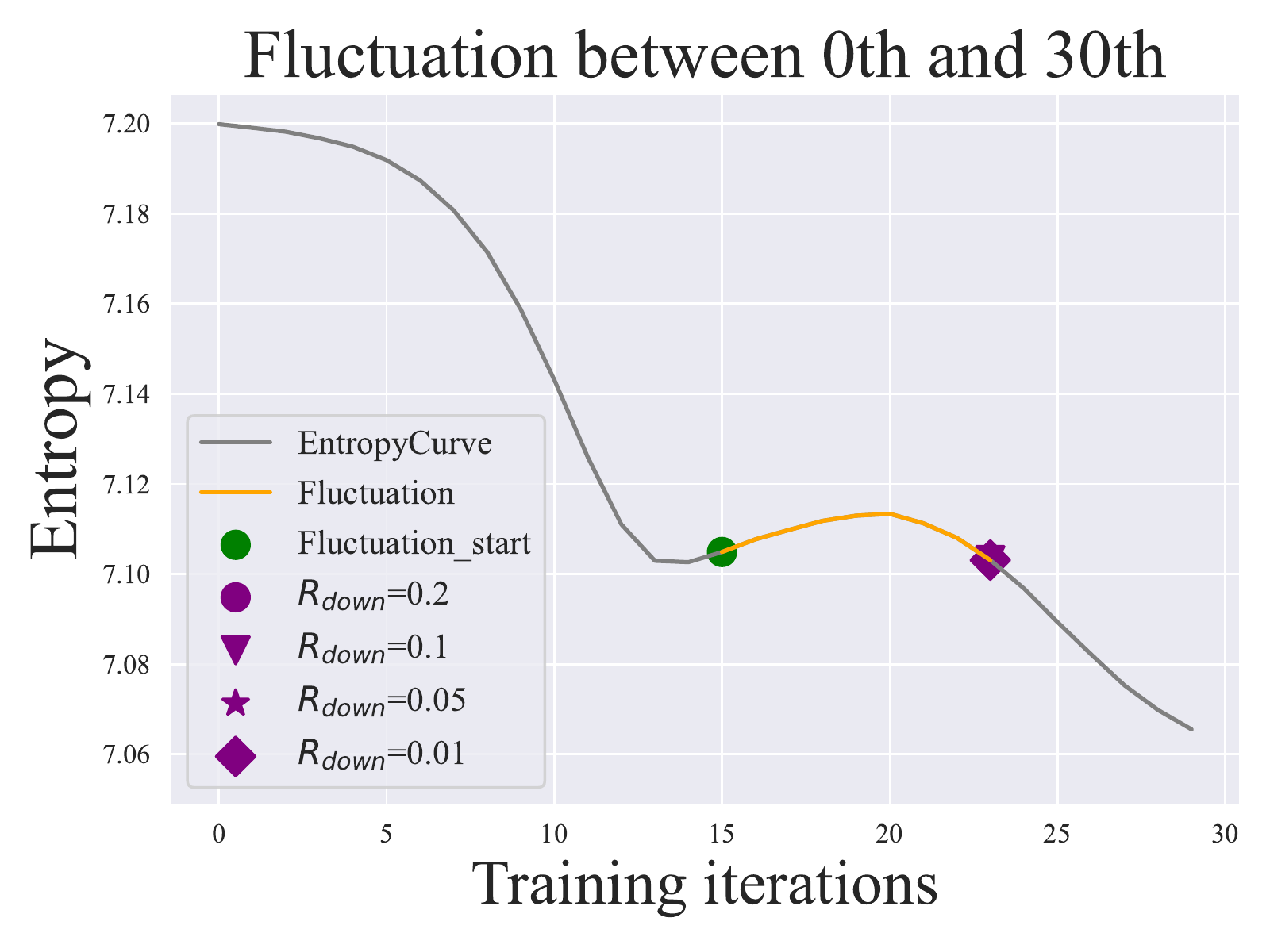}
  \includegraphics[width=0.45\textwidth]{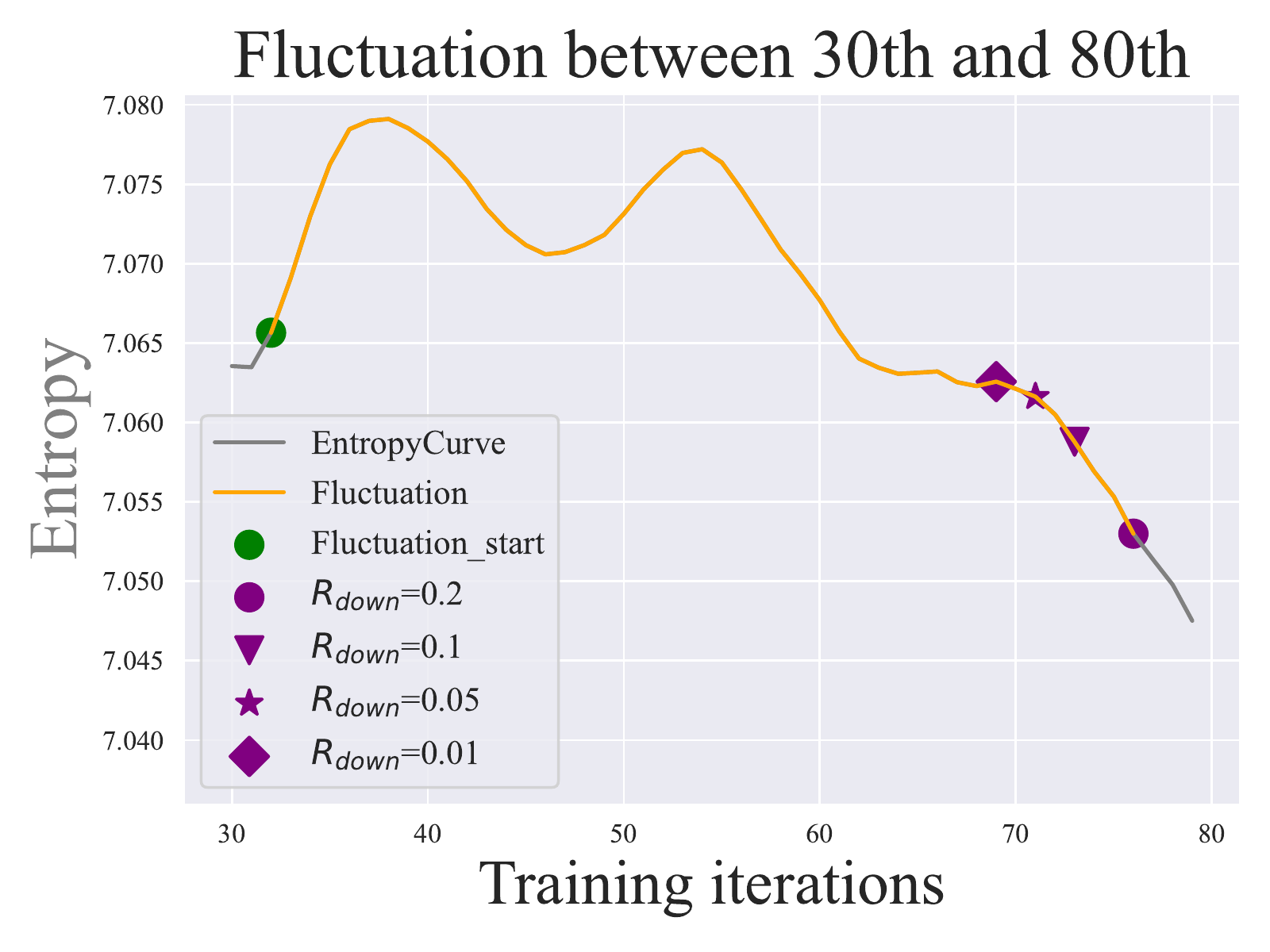}
  \caption{Explanation of the effect of $R_{down}$ in tolerating the existence of fluctuations in the entropy curve shown in Figure \ref{Fig:fluctuation}.}
  \label{Fig:r-down-explanation}
\end{figure*}

Regarding $R_{down}$, setting it to 0.1 is sufficient in most cases. In Sec. \ref{sec-further-invest} of our primary paper, our sensitivity study results verify that the EntropyStop algorithm is not sensitive to $R_{down}$ when its value lies in the range of $[0.01,0.1]$. A visualization of the effect of $R_{down}$ is depicted in Figure \ref{Fig:r-down-explanation}. When a small fluctuation (or rise) occurs during the downtrend of the curve, suppose $e_i$ is the start of this fluctuation. Then, the new lowest entropy points $e_q$ that satisfies the downtrend test of varying $R_{down}$ is close to each other. This explains the robustness of EntropyStop to $R_{down}$.

\begin{figure}[h]
  \centering
% \minipage{1.0\textwidth}
    \includegraphics[width=130.pt]{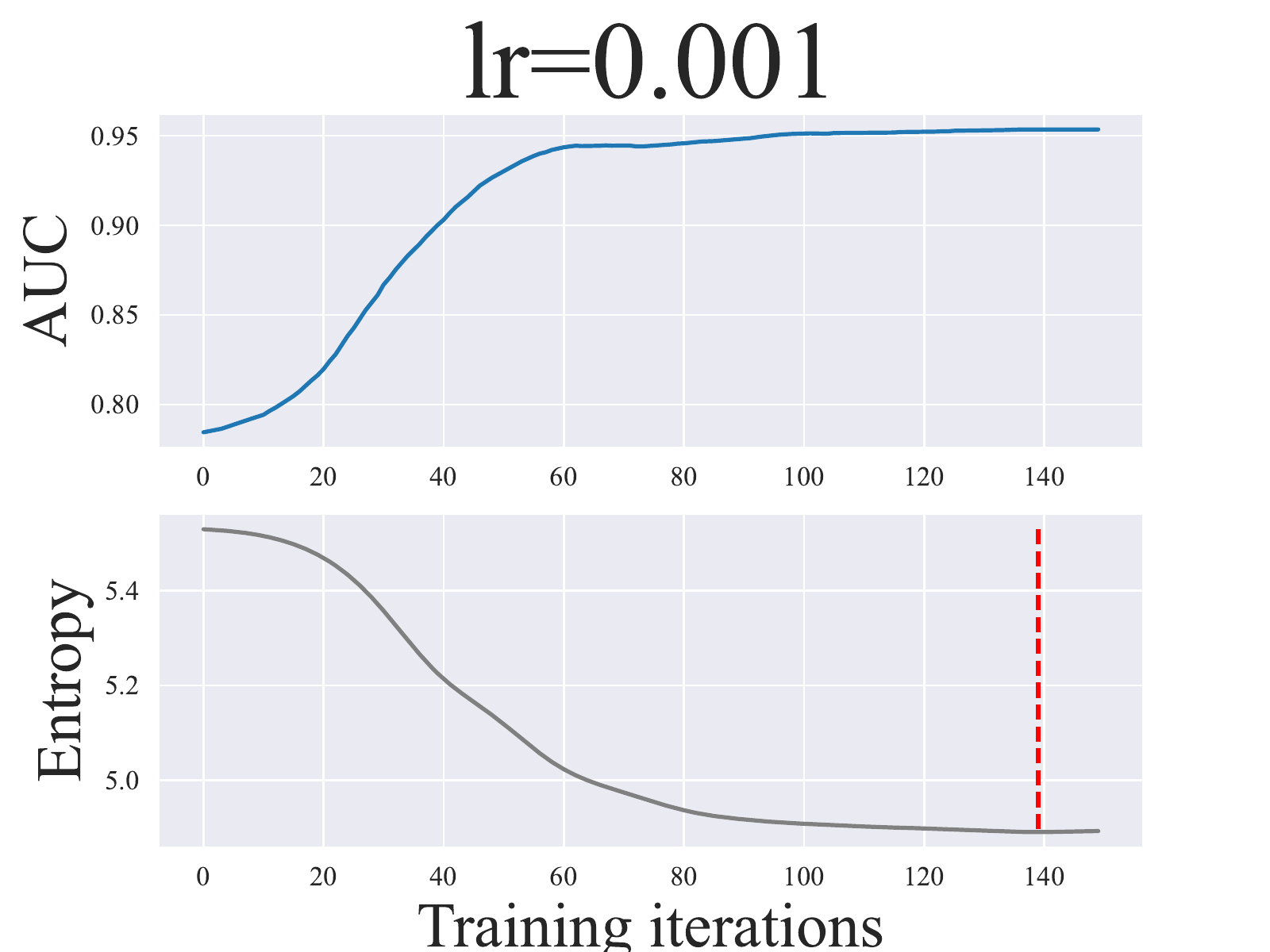}
  \includegraphics[width=130.pt]{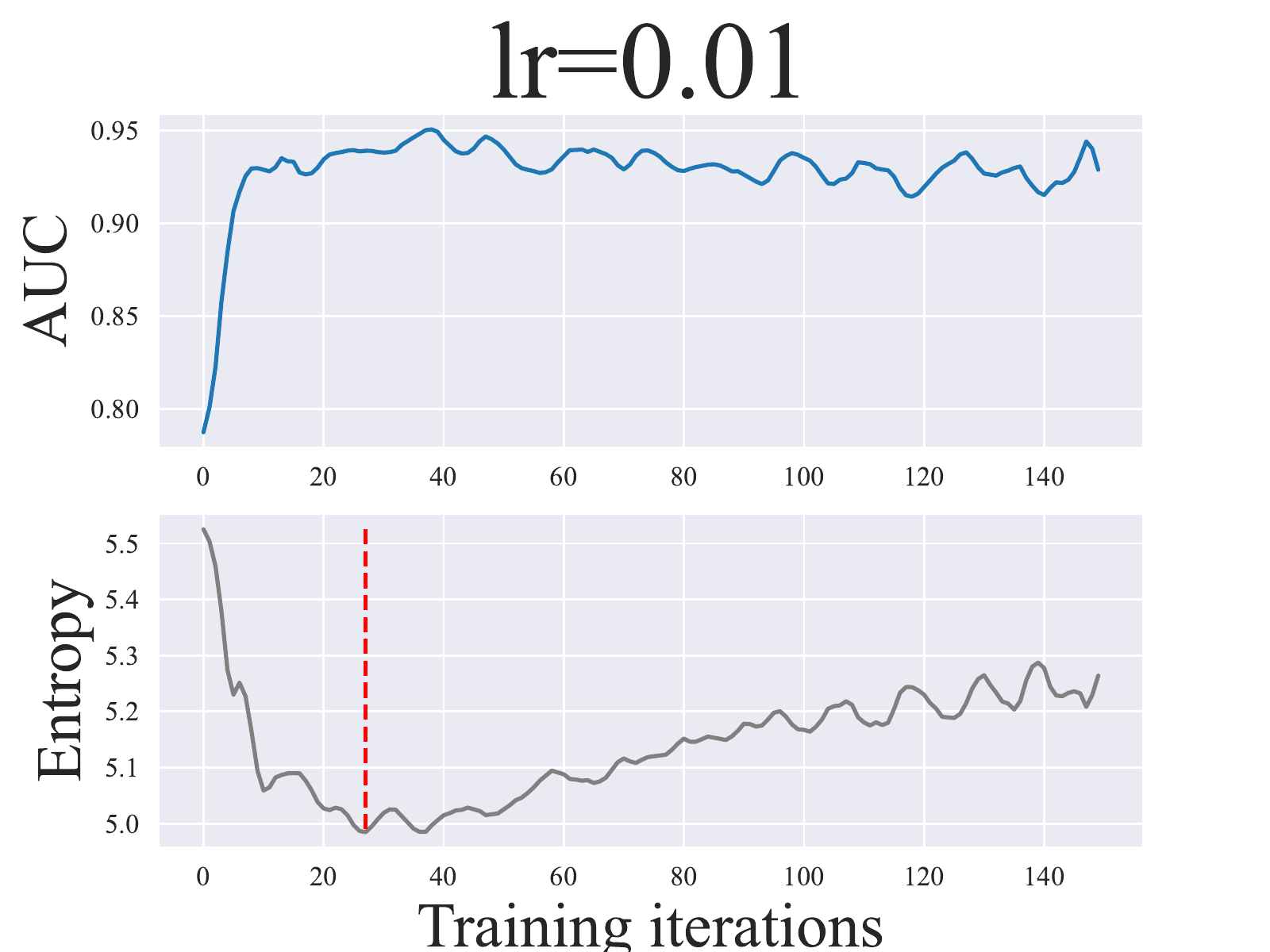}
    \includegraphics[width=130.pt]{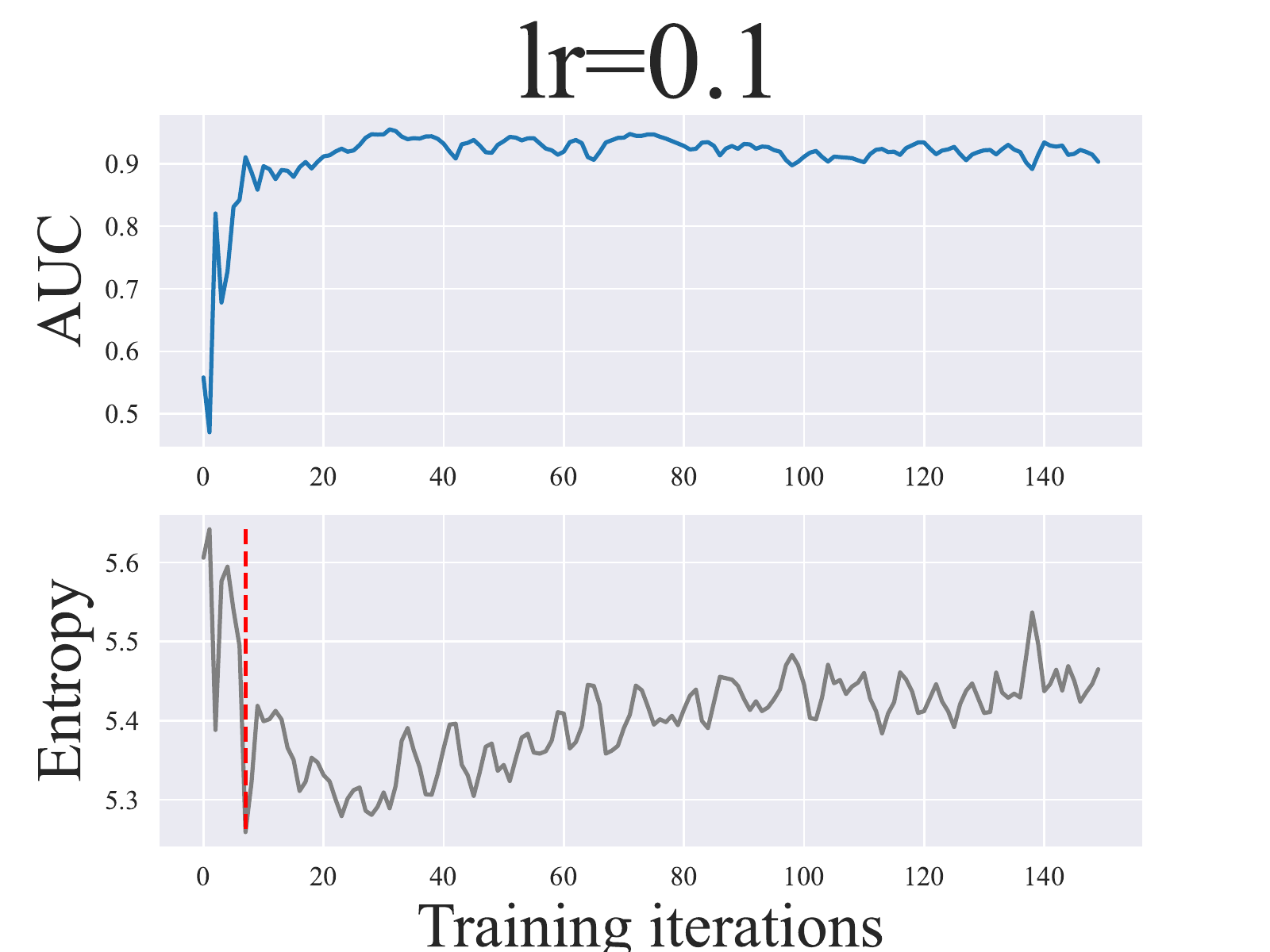}
  \caption{The effect of tolerating the fluctuations when $k$ is set to 50 and $R_{down}$ is set to 0.1 for the training of AE on the dataset \textit{Ionosphere}. The displayed training process only includes the first 150 iterations. The red dashed line marks the iteration selected by \textit{EntropyStop}.}
  \label{Fig:tolerate-to-lr}
\end{figure}

Owning to the effectiveness of $k$ and $R_{down}$ in tolerating the fluctuations, even the entropy curve is not smooth enough due to a large learning rate, the target iteration can still be selected by EntropyStop.(see Fig \ref{Fig:tolerate-to-lr}). Nevertheless, we still recommend fine-tuning the learning rate to achieve a smooth entropy curve, which will ensure a stable and reliable training process.

In Sec. \ref{sec-hp-testbed} of our primary paper, it is sufficient for AE on all 5 datasets when $k$=100. In finer granularity, $k$ can be set to 25 for the dataset \textit{Ionosphere} and MNIST to further save training time. 

% The setting of $k$ in our experiments is available at Table \ref{tab:k-for-entropy}.

\subsection{Details on Experiment Setup
}

\subsubsection{Dataset Description for Primary Experiment}
\label{appx:dataset-desc-for-hp}
We conduct our experiment in Sec. \ref{sec-hp-testbed} based on 3 tabular datasets from \cite{ADbench} (available at \url{https://github.com/Minqi824/ADBench}) and 2 image datasets from MNIST. For tabular datasets, we employ \verb|StandardScaler| in \verb|sklearn.preprocessing| to standardize the feature of each dataset. For image dataset, we employ the open-source code from \cite{robod} (available at \url{https://github.com/xyvivian/robod}) to preprocess the image dataset and generate the outliers. In specific, the global contrast normalization is employed to individual images. The inliers are assigned the label 0 and all classes other than the inlier class will be marked with 1, indicating the outlier class. We choose Digit ‘3’ and ‘5’ as the inlier-class, respectively, while the remaining classes are down-sampled to constitute the outliers.

\subsubsection{Hyperparameter Configurations}
\label{appx:hp-config}

Regarding the implementation of  deep OD method, we utilize open-source code available for RDP and DeepSVDD, two well-established algorithms in outlier detection research. Specifically, for RDP, we use the original open-source code\footnote{https://github.com/billhhh/RDP}. In the case of DeepSVDD, we employ the open-source code\footnote{https://github.com/xyvivian/robod} provided by the authors of the ROBOD paper \cite{robod}. We develop our own implementation for AE.

For RDP, we employ a training configuration where 30 mini-batch iterations are considered to be a single training epoch. The reason is the existence of parameter ``filter'' in RDP , which will filter out possible outliers per epoch. Therefore, we keep the consistency training configuration with the original implementation provided in the RDP open-source code.

\textbf{Model HP Descriptions and Grid of Values.}
\begin{table}[htbp]
  \centering
  \caption{We define a grid of 2-3 unique values for each hyperparameter (HP) of each deep OD method studied.The author-recommended value or the default values in their code is marked in bold  and underlined. }
    \begin{tabular}{rllr}
    \toprule
    \multicolumn{1}{l}{Method (\#models)} & Hyperparameter & Grid  & \multicolumn{1}{l}{\#values} \\
    \midrule
    \multicolumn{1}{l}{AE (64)} & act\_func & [relu, sigmoid] & 2 \\
          & dropout & [0.0, 0.2] & 2 \\
          & h\_dim & [64, 256] & 2 \\
          & lr    & [0.005, 0.001] & 2 \\
          & layers & [2, 4] & 2 \\
          & epoch & [100, 500] & 2 \\
    \midrule
    \multicolumn{1}{l}{RDP (72)} & out\_c & [25,\underline{\textbf{50}},100] & 3 \\
          & lr    & [\underline{\textbf{0.1}}, 0.01] & 2 \\
          & dropout & [0.0, \underline{\textbf{0.1}}] & 2 \\
          & filter & [0.0, \underline{\textbf{0.05}}, 0.1] & 3 \\
          & epoch & [100, 500] & 2 \\
    \midrule
    \multicolumn{1}{l}{DeepSVDD (64)} & conv\_dim & [ \underline{\textbf{8}},16] & 2 \\
          & fc\_dim & [\underline{\textbf{16}}, 64 & 2 \\
          & relu\_slope & [\underline{\textbf{0.1}}, 0.001] & 2 \\
          & epoch & [100, \underline{\textbf{250}}] & 2 \\
          & lr    & [1e-4, 1e-3] & 2 \\
          & wght\_dc & [1e-5, \underline{\textbf{1e-6}}] & 2 \\
    \bottomrule
    \end{tabular}%
  \label{tab:model-hp-grid-values}%
\end{table}%

\begin{itemize}
    \item AE: \begin{enumerate}
        \item \verb|act_func|: the activation function employed after the first layer of encoder
        \item \verb|dropout|: the possibility of dropping out the nodes in hidden layer during the training
        \item \verb|h_dim|: the dimension of hidden layers 
        \item \verb|lr|: the learning rate during the training
        \item \verb|layers|: the number of hidden layers
        \item \verb|epoch|: number of training epochs
        
    \end{enumerate}
    \item RDP: \begin{enumerate}
        \item \verb|out_c|: the dimension of random projection space
        \item \verb|lr|: the learning rate during the training
        \item \verb|dropout|: the possibility of dropping out the nodes in hidden layer during the training
        \item \verb|filter|: the percent of outliers to be filtered out per epoch
        \item \verb|epoch|: number of training epochs
        
    \end{enumerate}
    \item DeepSVDD: \begin{enumerate}
        \item \verb|conv_dim|: the number of output channels generated by the first convolutional encoder layer. Following this layer, the number of channels will expand at a rate of 2.
        \item \verb|fc_dim|: the output dimension of the fully connected layer between convolutional
encoder layers and decoder layers, in the LeNet structure \cite{lenet}.
        \item \verb|relu_slope|: DeepSVDD utilizes leaky-relu activation to avoid the trivial, uninformative solutions \cite{deep-svdd}. Here we alter the leakiness of the relu sloping.
        \item \verb|epoch|: number of training epochs
        \item \verb|lr|: the learning rate during the training
        \item \verb|wght_dc|: weight decay rate
    \end{enumerate}
\end{itemize}

\textbf{HP Configuration of Baselines.}

For \textit{EntropyStop}, we set $N_{eval} = 1024$ and $R_{down} = 0.1$ for all deep OD methods and datasets.  The values of $k$  are listed in Table \ref{tab:k-for-entropy}.

\begin{table}[htbp]
  \centering
  \caption{The setting of $k$ for \textit{EntropyStop}. There are 2 different learning rates for each deep OD method and the value of $k$ is set according to different learning rate as well.}
    \begin{tabular}{cccc}
    \toprule
    \textbf{Method} & \textbf{Dataset} & \textbf{$k$ for small lr} & \textbf{$k$ for large  lr} \\
    \midrule
    AE    & Ionosphere & 25    & 25 \\
          & letter & 100   & 100 \\
          & vowels & 100    & 100 \\
          & MNIST-3 & 25    & 25 \\
          & MNIST-5 & 25    & 25 \\
    \midrule
    RDP   & Ionosphere & 50    & 50 \\
          & letter & 750   & 150 \\
          & vowels & 750  & 150 \\
    \midrule
    DeepSVDD & MNIST-3 & 100   & 100 \\
          & MNIST-5 & 100   & 100 \\
    \bottomrule
    \end{tabular}%
  \label{tab:k-for-entropy}%
\end{table}%

% For Unsupervised Outlier Model Selection (UOMS) solutions, we provide the following definition to facilitate readers' comprehension. 

% \textbf{Preliminaries.} For a given OD algorithm (e.g. autoencoder), suppose $\mathcal{M} = \{M_i\}_{i=0}^N$ 
% denotes all  models with $N$ different HP configurations of this OD algorithm. When training $M_i$ on outlier dataset $D = \{\textbf{x}_j\}_{j=1}^n$, the corresponding outlier score list $s_i \in \mathbb{R}^n$ is obtained.

% \textbf{Problem (UOMS)} \textit{Given an unsupervised outlier detection task (i.e. dataset) $D = \{\textbf{x}_j\}_{j=1}^n$ and an OD algorithm (e.g. autoencoder) as well as its  models in $\mathcal{M}$ trained in $D$ with corresponding  outlier score lists $\{s_i\}_{i=0}^N$,  select a model $M_i \in \mathcal{M}$ where $s_i$ yields a good performance on $D$.}

% Note that our definition differs slightly from that of \cite{Internal-evaluation-paper}. In their definition, each UOMS solution attempts to select an OD algorithm along with its corresponding HP configuration for a given task. In contrast, our definition simplifies the problem by employing UOMS solely for the selection of HP configuration for a given OD algorithm and task, thus allowing for a more reasonable comparison.

For the implementation of Unsupervised Outlier Model Selection (UOMS) \cite{Internal-evaluation-paper} solutions (i.e. baseline XB, MC and HITS),  we use the open-source code\footnote{http://bit.ly/UOMSCODE} from \cite{Internal-evaluation-paper}. For XB, we give it the advantage to know the outlier ratio of dataset. For MC, the  Kendall $\tau$ coefficient is employed to calculate the similarity of outlier score list.

For AE ensemble model ROBOD, we use the open-source code\footnote{https://github.com/xyvivian/robod} from its original paper \cite{robod}. We employ their original code implementation of ensembling varying widths and depths of AE, alongside different random seeds, resulting in a total of 16 models for ensemble. The remaining HPs of AE are used to analyze its expected performance.

For Isolation Forest (IF), we use  its implementation from PYOD \cite{pyod} and adopt its default HP setting, which ensembles 100 different trees.

\subsubsection{Experiment setting for 
 Further Investigation}
\label{appx:dataset-inject}

For our experiment in Sec. \ref{sec-further-invest} of primary paper, we use 19 datasets  as well as outlier injection approach from ADbench\footnote{https://github.com/Minqi824/ADBench/tree/main/datasets/Classical} \cite{ADbench} to study the effectiveness in detecting heterogeneous outliers under varying outlier ratio. The original dataset statistics of 19 datasets are shown in Table \ref{tab:dataset-19}.  We remove the original outliers in the dataset and separately inject cluster, global and local outliers into the dataset, while controlling the outlier ratio to 0.1 and 0.4. We adopt the default value of parameters for the injection approach (see below definition). In Sec. \ref{sec-further-invest} Table \ref{tab:outlier-type-study}, we report the results of EntropyAE (i.e. AE with our \textit{EntropyStop}) with a specific HP configuration (listed in  Table \ref{tab:entropyae-outlier-type}) on detecting different types of outliers. The detailed detection results of Table \ref{tab:outlier-type-study} are available in Appx. \ref{appx:outlier-type-results}.

\begin{minipage}{1.2\textwidth}
\begin{minipage}[l]{0.42\textwidth}
% \begin{table}[htbp]
\small
  \centering
  \makeatletter\def\@captype{table}\makeatother \caption{19 datasets for heterogeneous outliers injection.}
  \vspace{0.3cm}
    \begin{tabular}{lrrr}
    \toprule
    \textbf{Dataset} & \textbf{Num Pts} & \textbf{Dim} & \textbf{\% Outlier} \\
    \midrule
    breastw & 683   & 9     & 34.99  \\
    cardio & 1831  & 21    & 9.61  \\
    Cardiotocography & 2114  & 21    & 22.04  \\
    fault & 1941  & 27    & 34.67  \\
    glass & 214   & 7     & 4.21  \\
    Hepatitis & 80    & 19    & 16.25  \\
    InternetAds & 1966  & 1555  & 18.72  \\
    Ionosphere & 351   & 32    & 35.90  \\
    letter & 1600  & 32    & 6.25  \\
    Lymphography & 148   & 18    & 4.05  \\
    Pima  & 768   & 8     & 34.90  \\
    Stamps & 340   & 9     & 9.12  \\
    vertebral & 240   & 6     & 12.50  \\
    vowels & 1456  & 12    & 3.43  \\
    WBC   & 223   & 9     & 4.48  \\
    WDBC  & 367   & 30    & 2.72  \\
    wine  & 129   & 13    & 7.75  \\
    WPBC  & 198   & 33    & 23.74  \\
    yeast & 1484  & 8     & 34.16  \\
    \bottomrule
    \end{tabular}%
  \label{tab:dataset-19}%
% \end{table}%
\end{minipage}
\hspace{0.8cm}
\begin{minipage}[r]{0.3\textwidth}
% \begin{table}[htbp]
\small
  \centering
\makeatletter\def\@captype{table}\makeatother 
\caption{The hyperparameter of EntropyAE for the experiment in Sec. \ref{sec-further-invest} and Table \ref{tab:outlier-type-study}.}
\vspace{0.3cm}
    \begin{tabular}{cc}
    \toprule
    \textbf{Hyperparameter} & \textbf{value} \\
    \midrule
    act\_func & relu \\
    dropout & 0.2 \\
    h\_dim & 64 \\
    num\_layer & 2 \\
    lr    & 0.001 \\
    epoch & 250 \\
    batch size & 256 \\
    \midrule
    $N_{eval}$ & 1024 \\
    $k$     & 100 \\
    $R_{down}$ & 0.1 \\
    \bottomrule
    \end{tabular}%
  \label{tab:entropyae-outlier-type}%
% \end{table}%
\end{minipage}
\vspace{0.3cm}
\end{minipage}

\textbf{Definition and Generation Process of Three Types of Outliers Used in ADBench \cite{ADbench} :}
\begin{itemize}
    \item  \textbf{Local Outlier} refer to the outliers that are deviant from their local neighborhoods \cite{lof}. The injection method follows the GMM procedure \cite{GMM} to generate synthetic normal samples, and then scale the covariance matrix $\hat{\Sigma} = \alpha \hat{\Sigma}$ by a scaling parameter $\alpha = 5$ to generate local outlier.
    \item  \textbf{Global Outlier} are more different from the normal data \cite{global-outlier}, generated from 
    a uniform distribution $Unif(\alpha \cdot min(\textbf{X}^k),\alpha \cdot max(\textbf{X}^k))$, where the boundaries are defined as the $min$ and $max$ of an input feature, e.g., $k$-th feature $\textbf{X}^k$, and $\alpha = 1.1$ controls the outlyingness.
    \item  \textbf{Cluster Outlier}, also known as group outliers \cite{cluster-outlier-1}, exhibit similar characteristics \cite{cluster-2}. The injection method scales the mean feature vector of normal samples by $\alpha = 5$, i.e., $\hat{\mu} = \alpha \hat{\mu}$, where $\alpha$ controls the distance between outlier clusters and the normal, and use the scaled GMM to generate outliers.
\end{itemize}

\subsection{Detailed Experiment Results}

\subsubsection{Detailed Results for heterogeneous outliers at varying outlier ratios}
\label{appx:outlier-type-results}

In this section, we provide the detailed results of Table \ref{tab:outlier-type-study} in Sec. \ref{sec-further-invest}, which are shown in Table \ref{tab:0.1-cluster} - \ref{tab:0.4-local}.

\textbf{Results.} Our results demonstrate the effectiveness of our proposed approach in mitigating the impact of cluster outliers on  AE performance. Specifically, our approach successfully prevents AE from learning harmful signals from cluster outliers on all 19 datasets, which would otherwise lead to a significant decline in performance when using the \textit{Naive} training method. For global outliers, our approach improves the overall performance with a less significant level. This is due to the fact that the in-training performance of AE on detecting global outliers is not sensitive to the presence of harmful signals, resulting in only a slight improvement in performance with our approach. However, it should be noted that our approach is not applicable to detecting local outliers, as they behave similarly to inliers when the reconstruction loss of AE is computed.

\begin{table}[htbp]
\small
  \centering
  \caption{AUC Performance of EntropyAE and NaiveAE on detecting \textbf{cluster} outliers at \textbf{0.1} outlier ratio. Column \textit{nai\_time} and \textit{en\_time} denotes the training time of \textit{Naive} and \textit{Entropy} in seconds.}
    \begin{tabular}{rcccc}
    \toprule
    \textbf{Dataset} & \textbf{Entropy} & \textbf{Naive} & \textbf{en\_time} & \textbf{nai\_time} \\
    \midrule
    Cardiotocography & \textcolor[rgb]{ 1,  0,  0}{\textbf{0.799}} & 0.760  & \textcolor[rgb]{ 1,  0,  0}{\textbf{0.286}} & 2.278  \\
    Hepatitis & \textcolor[rgb]{ 1,  0,  0}{\textbf{1.000}} & 0.397  & \textcolor[rgb]{ 1,  0,  0}{\textbf{0.152}} & 0.295  \\
    InternetAds & 0.047  & \textcolor[rgb]{ 1,  0,  0}{\textbf{0.159}} & \textcolor[rgb]{ 1,  0,  0}{\textbf{0.376}} & 2.249  \\
    Ionosphere & \textcolor[rgb]{ 1,  0,  0}{\textbf{0.953}} & 0.685  & \textcolor[rgb]{ 1,  0,  0}{\textbf{0.154}} & 0.290  \\
    Lymphography & \textcolor[rgb]{ 1,  0,  0}{\textbf{0.997}} & 0.531  & \textcolor[rgb]{ 1,  0,  0}{\textbf{0.152}} & 0.289  \\
    Pima  & \textcolor[rgb]{ 1,  0,  0}{\textbf{0.997}} & 0.873  & \textcolor[rgb]{ 1,  0,  0}{\textbf{0.183}} & 0.887  \\
    Stamps & \textcolor[rgb]{ 1,  0,  0}{\textbf{0.941}} & 0.764  & \textcolor[rgb]{ 1,  0,  0}{\textbf{0.155}} & 0.569  \\
    WBC   & \textcolor[rgb]{ 1,  0,  0}{\textbf{0.926}} & 0.705  & \textcolor[rgb]{ 1,  0,  0}{\textbf{0.155}} & 0.286  \\
    WDBC  & \textcolor[rgb]{ 1,  0,  0}{\textbf{0.923}} & 0.579  & \textcolor[rgb]{ 1,  0,  0}{\textbf{0.167}} & 0.576  \\
    WPBC  & \textcolor[rgb]{ 1,  0,  0}{\textbf{0.484}} & 0.251  & \textcolor[rgb]{ 1,  0,  0}{\textbf{0.153}} & 0.287  \\
    breastw & \textcolor[rgb]{ 1,  0,  0}{\textbf{1.000}} & 0.713  & \textcolor[rgb]{ 1,  0,  0}{\textbf{0.181}} & 0.567  \\
    cardio & 0.692  & \textcolor[rgb]{ 1,  0,  0}{\textbf{0.744}} & \textcolor[rgb]{ 1,  0,  0}{\textbf{0.372}} & 2.275  \\
    fault & \textcolor[rgb]{ 1,  0,  0}{\textbf{1.000}} & 0.754  & \textcolor[rgb]{ 1,  0,  0}{\textbf{0.161}} & 1.770  \\
    glass & \textcolor[rgb]{ 1,  0,  0}{\textbf{0.959}} & 0.889  & \textcolor[rgb]{ 1,  0,  0}{\textbf{0.161}} & 0.308  \\
    letter & \textcolor[rgb]{ 1,  0,  0}{\textbf{1.000}} & 0.778  & \textcolor[rgb]{ 1,  0,  0}{\textbf{0.180}} & 2.017  \\
    vertebral & \textcolor[rgb]{ 1,  0,  0}{\textbf{0.944}} & 0.660  & \textcolor[rgb]{ 1,  0,  0}{\textbf{0.161}} & 0.287  \\
    vowels & \textcolor[rgb]{ 1,  0,  0}{\textbf{1.000}} & 0.798  & \textcolor[rgb]{ 1,  0,  0}{\textbf{0.157}} & 1.987  \\
    wine  & \textcolor[rgb]{ 1,  0,  0}{\textbf{1.000}} & 0.438  & \textcolor[rgb]{ 1,  0,  0}{\textbf{0.161}} & 0.291  \\
    yeast & 0.763  & \textcolor[rgb]{ 1,  0,  0}{\textbf{0.803}} & \textcolor[rgb]{ 1,  0,  0}{\textbf{0.378}} & 1.414  \\
    \bottomrule
    \end{tabular}%
  \label{tab:0.1-cluster}%
\end{table}%

% Table generated by Excel2LaTeX from sheet 'inject-0.4-cluster'
\begin{table}[htbp]
\small
  \centering
  \caption{AUC Performance of EntropyAE and NaiveAE on detecting \textbf{cluster} outliers at \textbf{0.4} outlier ratio. Column \textit{nai\_time} and \textit{en\_time} denotes the training time of \textit{Naive} and \textit{Entropy} in seconds.}
    \begin{tabular}{rcccc}
    \toprule
    \textbf{Dataset} & \textbf{Entropy} & \textbf{Naive} & \textbf{en\_time} & \textbf{nai\_time} \\
    \midrule
    Cardiotocography & \textcolor[rgb]{ 1,  0,  0}{\textbf{0.656}} & 0.592  & \textcolor[rgb]{ 1,  0,  0}{\textbf{0.337}} & 3.141  \\
    Hepatitis & \textcolor[rgb]{ 1,  0,  0}{\textbf{0.914}} & 0.507  & \textcolor[rgb]{ 1,  0,  0}{\textbf{0.203}} & 0.301  \\
    InternetAds & 0.172  & \textcolor[rgb]{ 1,  0,  0}{\textbf{0.290}} & \textcolor[rgb]{ 1,  0,  0}{\textbf{0.369}} & 3.346  \\
    Ionosphere & \textcolor[rgb]{ 1,  0,  0}{\textbf{0.932}} & 0.622  & \textcolor[rgb]{ 1,  0,  0}{\textbf{0.230}} & 0.563  \\
    Lymphography & \textcolor[rgb]{ 1,  0,  0}{\textbf{0.996}} & 0.550  & \textcolor[rgb]{ 1,  0,  0}{\textbf{0.154}} & 0.292  \\
    Pima  & \textcolor[rgb]{ 1,  0,  0}{\textbf{0.973}} & 0.723  & \textcolor[rgb]{ 1,  0,  0}{\textbf{0.243}} & 1.143  \\
    Stamps & \textcolor[rgb]{ 1,  0,  0}{\textbf{0.931}} & 0.551  & \textcolor[rgb]{ 1,  0,  0}{\textbf{0.257}} & 0.852  \\
    WBC   & \textcolor[rgb]{ 1,  0,  0}{\textbf{0.690}} & 0.605  & \textcolor[rgb]{ 1,  0,  0}{\textbf{0.365}} & 0.595  \\
    WDBC  & \textcolor[rgb]{ 1,  0,  0}{\textbf{0.930}} & 0.493  & \textcolor[rgb]{ 1,  0,  0}{\textbf{0.162}} & 0.870  \\
    WPBC  & \textcolor[rgb]{ 1,  0,  0}{\textbf{0.598}} & 0.364  & \textcolor[rgb]{ 1,  0,  0}{\textbf{0.155}} & 0.299  \\
    breastw & \textcolor[rgb]{ 1,  0,  0}{\textbf{0.868}} & 0.593  & \textcolor[rgb]{ 1,  0,  0}{\textbf{0.239}} & 0.849  \\
    cardio & \textcolor[rgb]{ 1,  0,  0}{\textbf{0.686}} & 0.594  & \textcolor[rgb]{ 1,  0,  0}{\textbf{0.318}} & 3.148  \\
    fault & \textcolor[rgb]{ 1,  0,  0}{\textbf{1.000}} & 0.598  & \textcolor[rgb]{ 1,  0,  0}{\textbf{0.162}} & 2.606  \\
    glass & \textcolor[rgb]{ 1,  0,  0}{\textbf{0.948}} & 0.706  & \textcolor[rgb]{ 1,  0,  0}{\textbf{0.264}} & 0.597  \\
    letter & \textcolor[rgb]{ 1,  0,  0}{\textbf{1.000}} & 0.575  & \textcolor[rgb]{ 1,  0,  0}{\textbf{0.200}} & 2.972  \\
    vertebral & \textcolor[rgb]{ 1,  0,  0}{\textbf{0.945}} & 0.485  & \textcolor[rgb]{ 1,  0,  0}{\textbf{0.156}} & 0.560  \\
    vowels & \textcolor[rgb]{ 1,  0,  0}{\textbf{0.955}} & 0.669  & \textcolor[rgb]{ 1,  0,  0}{\textbf{0.255}} & 2.851  \\
    wine  & \textcolor[rgb]{ 1,  0,  0}{\textbf{1.000}} & 0.463  & \textcolor[rgb]{ 1,  0,  0}{\textbf{0.153}} & 0.293  \\
    yeast & \textcolor[rgb]{ 1,  0,  0}{\textbf{0.684}} & 0.644  & \textcolor[rgb]{ 1,  0,  0}{\textbf{0.362}} & 1.986  \\
    \bottomrule
    \end{tabular}%
  \label{tab:0.4-cluster}%
\end{table}%

% Table generated by Excel2LaTeX from sheet 'inject-0.1-global'
\begin{table}[htbp]
\small
  \centering
  \caption{AUC Performance of EntropyAE and NaiveAE on detecting \textbf{global} outliers at \textbf{0.1} outlier ratio. Column \textit{nai\_time} and \textit{en\_time} denotes the training time of \textit{Naive} and \textit{Entropy} in seconds.}
    \begin{tabular}{rcccc}
    \toprule
    \textbf{Dataset} & \textbf{Entropy} & \textbf{Naive} & \textbf{en\_time} & \textbf{nai\_time} \\
    \midrule
    Cardiotocography & 0.999  & 0.999  & \textcolor[rgb]{ 1,  0,  0}{\textbf{0.668}} & 2.431  \\
    Hepatitis & \textcolor[rgb]{ 1,  0,  0}{\textbf{0.881}} & 0.510  & \textcolor[rgb]{ 1,  0,  0}{\textbf{0.253}} & 0.293  \\
    InternetAds & 1.000  & 1.000  & \textcolor[rgb]{ 1,  0,  0}{\textbf{1.723}} & 2.241  \\
    Ionosphere & 1.000  & 1.000  & 0.372  & \textcolor[rgb]{ 1,  0,  0}{\textbf{0.293}} \\
    Lymphography & \textcolor[rgb]{ 1,  0,  0}{\textbf{0.986}} & 0.858  & \textcolor[rgb]{ 1,  0,  0}{\textbf{0.161}} & 0.309  \\
    Pima  & 0.909  & \textcolor[rgb]{ 1,  0,  0}{\textbf{0.940}} & \textcolor[rgb]{ 1,  0,  0}{\textbf{0.614}} & 0.889  \\
    Stamps & \textcolor[rgb]{ 1,  0,  0}{\textbf{0.986}} & 0.953  & \textcolor[rgb]{ 1,  0,  0}{\textbf{0.158}} & 0.621  \\
    WBC   & \textcolor[rgb]{ 1,  0,  0}{\textbf{0.986}} & 0.940  & \textcolor[rgb]{ 1,  0,  0}{\textbf{0.155}} & 0.290  \\
    WDBC  & 1.000  & 1.000  & \textcolor[rgb]{ 1,  0,  0}{\textbf{0.478}} & 0.570  \\
    WPBC  & 1.000  & 1.000  & 0.349  & \textcolor[rgb]{ 1,  0,  0}{\textbf{0.305}} \\
    breastw & \textcolor[rgb]{ 1,  0,  0}{\textbf{0.990}} & 0.955  & \textcolor[rgb]{ 1,  0,  0}{\textbf{0.234}} & 0.596  \\
    cardio & 0.998  & 0.998  & \textcolor[rgb]{ 1,  0,  0}{\textbf{0.661}} & 2.361  \\
    fault & 1.000  & 1.000  & \textcolor[rgb]{ 1,  0,  0}{\textbf{0.921}} & 1.767  \\
    glass & 0.954  & \textcolor[rgb]{ 1,  0,  0}{\textbf{0.979}} & \textcolor[rgb]{ 1,  0,  0}{\textbf{0.237}} & 0.292  \\
    letter & 1.000  & 1.000  & \textcolor[rgb]{ 1,  0,  0}{\textbf{0.606}} & 2.122  \\
    vertebral & 0.968  & 0.968  & 0.445  & \textcolor[rgb]{ 1,  0,  0}{\textbf{0.301}} \\
    vowels & 0.999  & 0.999  & \textcolor[rgb]{ 1,  0,  0}{\textbf{0.789}} & 2.103  \\
    wine  & \textcolor[rgb]{ 1,  0,  0}{\textbf{0.988}} & 0.910  & 0.373  & \textcolor[rgb]{ 1,  0,  0}{\textbf{0.299}} \\
    yeast & \textcolor[rgb]{ 1,  0,  0}{\textbf{0.990}} & 0.930  & \textcolor[rgb]{ 1,  0,  0}{\textbf{0.154}} & 1.428  \\
    \bottomrule
    \end{tabular}%
  \label{tab:0.1-global}%
\end{table}%

% Table generated by Excel2LaTeX from sheet 'inject-0.4-global'
\begin{table}[htbp]
\small
  \centering
  \caption{AUC Performance of EntropyAE and NaiveAE on detecting \textbf{global} outliers at \textbf{0.4} outlier ratio. Column \textit{nai\_time} and \textit{en\_time} denotes the training time of \textit{Naive} and \textit{Entropy} in seconds.}
    \begin{tabular}{rcccc}
 \toprule
    \textbf{Dataset} & \textbf{Entropy} & \textbf{Naive} & \textbf{en\_time} & \textbf{nai\_time} \\
    \midrule
    Cardiotocography & \textcolor[rgb]{ 1,  0,  0}{\textbf{0.656}} & 0.592  & \textcolor[rgb]{ 1,  0,  0}{\textbf{0.337}} & 3.141  \\
    Hepatitis & \textcolor[rgb]{ 1,  0,  0}{\textbf{0.914}} & 0.507  & \textcolor[rgb]{ 1,  0,  0}{\textbf{0.203}} & 0.301  \\
    InternetAds & 0.172  & \textcolor[rgb]{ 1,  0,  0}{\textbf{0.290}} & \textcolor[rgb]{ 1,  0,  0}{\textbf{0.369}} & 3.346  \\
    Ionosphere & \textcolor[rgb]{ 1,  0,  0}{\textbf{0.932}} & 0.622  & \textcolor[rgb]{ 1,  0,  0}{\textbf{0.230}} & 0.563  \\
    Lymphography & \textcolor[rgb]{ 1,  0,  0}{\textbf{0.996}} & 0.550  & \textcolor[rgb]{ 1,  0,  0}{\textbf{0.154}} & 0.292  \\
    Pima  & \textcolor[rgb]{ 1,  0,  0}{\textbf{0.973}} & 0.723  & \textcolor[rgb]{ 1,  0,  0}{\textbf{0.243}} & 1.143  \\
    Stamps & \textcolor[rgb]{ 1,  0,  0}{\textbf{0.931}} & 0.551  & \textcolor[rgb]{ 1,  0,  0}{\textbf{0.257}} & 0.852  \\
    WBC   & \textcolor[rgb]{ 1,  0,  0}{\textbf{0.690}} & 0.605  & \textcolor[rgb]{ 1,  0,  0}{\textbf{0.365}} & 0.595  \\
    WDBC  & \textcolor[rgb]{ 1,  0,  0}{\textbf{0.930}} & 0.493  & \textcolor[rgb]{ 1,  0,  0}{\textbf{0.162}} & 0.870  \\
    WPBC  & \textcolor[rgb]{ 1,  0,  0}{\textbf{0.598}} & 0.364  & \textcolor[rgb]{ 1,  0,  0}{\textbf{0.155}} & 0.299  \\
    breastw & \textcolor[rgb]{ 1,  0,  0}{\textbf{0.868}} & 0.593  & \textcolor[rgb]{ 1,  0,  0}{\textbf{0.239}} & 0.849  \\
    cardio & \textcolor[rgb]{ 1,  0,  0}{\textbf{0.686}} & 0.594  & \textcolor[rgb]{ 1,  0,  0}{\textbf{0.318}} & 3.148  \\
    fault & \textcolor[rgb]{ 1,  0,  0}{\textbf{1.000}} & 0.598  & \textcolor[rgb]{ 1,  0,  0}{\textbf{0.162}} & 2.606  \\
    glass & \textcolor[rgb]{ 1,  0,  0}{\textbf{0.948}} & 0.706  & \textcolor[rgb]{ 1,  0,  0}{\textbf{0.264}} & 0.597  \\
    letter & \textcolor[rgb]{ 1,  0,  0}{\textbf{1.000}} & 0.575  & \textcolor[rgb]{ 1,  0,  0}{\textbf{0.200}} & 2.972  \\
    vertebral & \textcolor[rgb]{ 1,  0,  0}{\textbf{0.945}} & 0.485  & \textcolor[rgb]{ 1,  0,  0}{\textbf{0.156}} & 0.560  \\
    vowels & \textcolor[rgb]{ 1,  0,  0}{\textbf{0.955}} & 0.669  & \textcolor[rgb]{ 1,  0,  0}{\textbf{0.255}} & 2.851  \\
    wine  & \textcolor[rgb]{ 1,  0,  0}{\textbf{1.000}} & 0.463  & \textcolor[rgb]{ 1,  0,  0}{\textbf{0.153}} & 0.293  \\
    yeast & \textcolor[rgb]{ 1,  0,  0}{\textbf{0.684}} & 0.644  & \textcolor[rgb]{ 1,  0,  0}{\textbf{0.362}} & 1.986  \\
    \bottomrule
    \end{tabular}%
  \label{tab:0.4-global}%
\end{table}%

% Table generated by Excel2LaTeX from sheet 'inject-0.1-local'
\begin{table}[htbp]
\small
  \centering
  \caption{AUC Performance of EntropyAE and NaiveAE on detecting \textbf{local} outliers at \textbf{0.1} outlier ratio. Column \textit{nai\_time} and \textit{en\_time} denotes the training time of \textit{Naive} and \textit{Entropy} in seconds.}
    \begin{tabular}{rcccc}
    \toprule
    \textbf{Dataset} & \textbf{Entropy} & \textbf{Naive} & \textbf{en\_time} & \textbf{nai\_time} \\
    \midrule
    Cardiotocography & 0.956  & \textcolor[rgb]{ 1,  0,  0}{\textbf{0.959}} & \textcolor[rgb]{ 1,  0,  0}{\textbf{0.802}} & 2.298  \\
    Hepatitis & \textcolor[rgb]{ 1,  0,  0}{\textbf{0.690}} & 0.561  & 0.339  & \textcolor[rgb]{ 1,  0,  0}{\textbf{0.281}} \\
    InternetAds & 1.000  & 1.000  & \textcolor[rgb]{ 1,  0,  0}{\textbf{0.169}} & 2.167  \\
    Ionosphere & 0.927  & \textcolor[rgb]{ 1,  0,  0}{\textbf{0.977}} & \textcolor[rgb]{ 1,  0,  0}{\textbf{0.248}} & 0.286  \\
    Lymphography & \textcolor[rgb]{ 1,  0,  0}{\textbf{0.878}} & 0.848  & \textcolor[rgb]{ 1,  0,  0}{\textbf{0.165}} & 0.323  \\
    Pima  & 0.887  & \textcolor[rgb]{ 1,  0,  0}{\textbf{0.897}} & \textcolor[rgb]{ 1,  0,  0}{\textbf{0.802}} & 0.877  \\
    Stamps & 0.831  & \textcolor[rgb]{ 1,  0,  0}{\textbf{0.868}} & \textcolor[rgb]{ 1,  0,  0}{\textbf{0.410}} & 0.575  \\
    WBC   & 0.830  & \textcolor[rgb]{ 1,  0,  0}{\textbf{0.833}} & \textcolor[rgb]{ 1,  0,  0}{\textbf{0.186}} & 0.284  \\
    WDBC  & 0.952  & \textcolor[rgb]{ 1,  0,  0}{\textbf{0.955}} & \textcolor[rgb]{ 1,  0,  0}{\textbf{0.185}} & 0.567  \\
    WPBC  & \textcolor[rgb]{ 1,  0,  0}{\textbf{0.980}} & 0.810  & \textcolor[rgb]{ 1,  0,  0}{\textbf{0.162}} & 0.305  \\
    breastw & \textcolor[rgb]{ 1,  0,  0}{\textbf{0.834}} & 0.813  & \textcolor[rgb]{ 1,  0,  0}{\textbf{0.240}} & 0.599  \\
    cardio & 0.947  & \textcolor[rgb]{ 1,  0,  0}{\textbf{0.949}} & \textcolor[rgb]{ 1,  0,  0}{\textbf{0.744}} & 2.292  \\
    fault & 0.945  & \textcolor[rgb]{ 1,  0,  0}{\textbf{0.961}} & \textcolor[rgb]{ 1,  0,  0}{\textbf{0.618}} & 1.709  \\
    glass & 0.736  & \textcolor[rgb]{ 1,  0,  0}{\textbf{0.849}} & \textcolor[rgb]{ 1,  0,  0}{\textbf{0.213}} & 0.284  \\
    letter & 0.990  & \textcolor[rgb]{ 1,  0,  0}{\textbf{0.995}} & \textcolor[rgb]{ 1,  0,  0}{\textbf{0.585}} & 2.050  \\
    vertebral & \textcolor[rgb]{ 1,  0,  0}{\textbf{0.877}} & 0.869  & 0.427  & \textcolor[rgb]{ 1,  0,  0}{\textbf{0.284}} \\
    vowels & \textcolor[rgb]{ 1,  0,  0}{\textbf{0.955}} & 0.954  & \textcolor[rgb]{ 1,  0,  0}{\textbf{0.922}} & 1.977  \\
    wine  & \textcolor[rgb]{ 1,  0,  0}{\textbf{0.977}} & 0.892  & \textcolor[rgb]{ 1,  0,  0}{\textbf{0.227}} & 3.738  \\
    yeast & 0.918  & \textcolor[rgb]{ 1,  0,  0}{\textbf{0.933}} & \textcolor[rgb]{ 1,  0,  0}{\textbf{0.408}} & 1.455  \\
    \bottomrule
    \end{tabular}%
  \label{tab:0.1-local}%
\end{table}%

% Table generated by Excel2LaTeX from sheet 'inject-0.4-local'
\begin{table}[htbp]
\small
  \centering
  \caption{AUC Performance of EntropyAE and NaiveAE on detecting \textbf{local} outliers at \textbf{0.4} outlier ratio. Column \textit{nai\_time} and \textit{en\_time} denotes the training time of \textit{Naive} and \textit{Entropy} in seconds.}
    \begin{tabular}{rcccc}
    \toprule
    \textbf{Dataset} & \textbf{Entropy} & \textbf{Naive} & \textbf{en\_time} & \textbf{nai\_time} \\
    \midrule
    Cardiotocography & 0.941  & \textcolor[rgb]{ 1,  0,  0}{\textbf{0.944}} & \textcolor[rgb]{ 1,  0,  0}{\textbf{0.834}} & 3.109  \\
    Hepatitis & \textcolor[rgb]{ 1,  0,  0}{\textbf{0.905}} & 0.806  & \textcolor[rgb]{ 1,  0,  0}{\textbf{0.242}} & 0.282  \\
    InternetAds & 1.000  & 1.000  & \textcolor[rgb]{ 1,  0,  0}{\textbf{0.610}} & 3.427  \\
    Ionosphere & 0.927  & \textcolor[rgb]{ 1,  0,  0}{\textbf{0.948}} & \textcolor[rgb]{ 1,  0,  0}{\textbf{0.448}} & 0.600  \\
    Lymphography & \textcolor[rgb]{ 1,  0,  0}{\textbf{0.886}} & 0.868  & \textcolor[rgb]{ 1,  0,  0}{\textbf{0.154}} & 0.285  \\
    Pima  & 0.865  & \textcolor[rgb]{ 1,  0,  0}{\textbf{0.882}} & \textcolor[rgb]{ 1,  0,  0}{\textbf{0.689}} & 1.148  \\
    Stamps & 0.850  & \textcolor[rgb]{ 1,  0,  0}{\textbf{0.860}} & 0.944  & \textcolor[rgb]{ 1,  0,  0}{\textbf{0.850}} \\
    WBC   & 0.837  & \textcolor[rgb]{ 1,  0,  0}{\textbf{0.876}} & \textcolor[rgb]{ 1,  0,  0}{\textbf{0.462}} & 0.561  \\
    WDBC  & 0.947  & \textcolor[rgb]{ 1,  0,  0}{\textbf{0.984}} & \textcolor[rgb]{ 1,  0,  0}{\textbf{0.179}} & 0.850  \\
    WPBC  & 0.981  & \textcolor[rgb]{ 1,  0,  0}{\textbf{0.998}} & \textcolor[rgb]{ 1,  0,  0}{\textbf{0.150}} & 0.286  \\
    breastw & 0.794  & \textcolor[rgb]{ 1,  0,  0}{\textbf{0.810}} & 0.893  & \textcolor[rgb]{ 1,  0,  0}{\textbf{0.843}} \\
    cardio & 0.920  & \textcolor[rgb]{ 1,  0,  0}{\textbf{0.928}} & \textcolor[rgb]{ 1,  0,  0}{\textbf{0.834}} & 3.130  \\
    fault & 0.954  & \textcolor[rgb]{ 1,  0,  0}{\textbf{0.964}} & \textcolor[rgb]{ 1,  0,  0}{\textbf{0.654}} & 2.562  \\
    glass & 0.652  & \textcolor[rgb]{ 1,  0,  0}{\textbf{0.774}} & \textcolor[rgb]{ 1,  0,  0}{\textbf{0.153}} & 0.573  \\
    letter & 0.989  & \textcolor[rgb]{ 1,  0,  0}{\textbf{0.995}} & \textcolor[rgb]{ 1,  0,  0}{\textbf{0.697}} & 3.067  \\
    vertebral & 0.849  & \textcolor[rgb]{ 1,  0,  0}{\textbf{0.845}} & 0.761  & \textcolor[rgb]{ 1,  0,  0}{\textbf{0.567}} \\
    vowels & 0.953  & \textcolor[rgb]{ 1,  0,  0}{\textbf{0.959}} & \textcolor[rgb]{ 1,  0,  0}{\textbf{1.220}} & 2.822  \\
    wine  & \textcolor[rgb]{ 1,  0,  0}{\textbf{0.973}} & 0.950  & \textcolor[rgb]{ 1,  0,  0}{\textbf{0.249}} & 0.283  \\
    yeast & 0.867  & \textcolor[rgb]{ 1,  0,  0}{\textbf{0.899}} & \textcolor[rgb]{ 1,  0,  0}{\textbf{0.434}} & 1.956  \\
    \bottomrule
    \end{tabular}%
  \label{tab:0.4-local}%
\end{table}%

\subsubsection{Additional Experiment Results}
\label{appx:additonal-fig}

\begin{wrapfigure}[11]{r}{0.4\textwidth}
  \centering
  \vspace{-1cm}
\includegraphics[width=120.00pt]{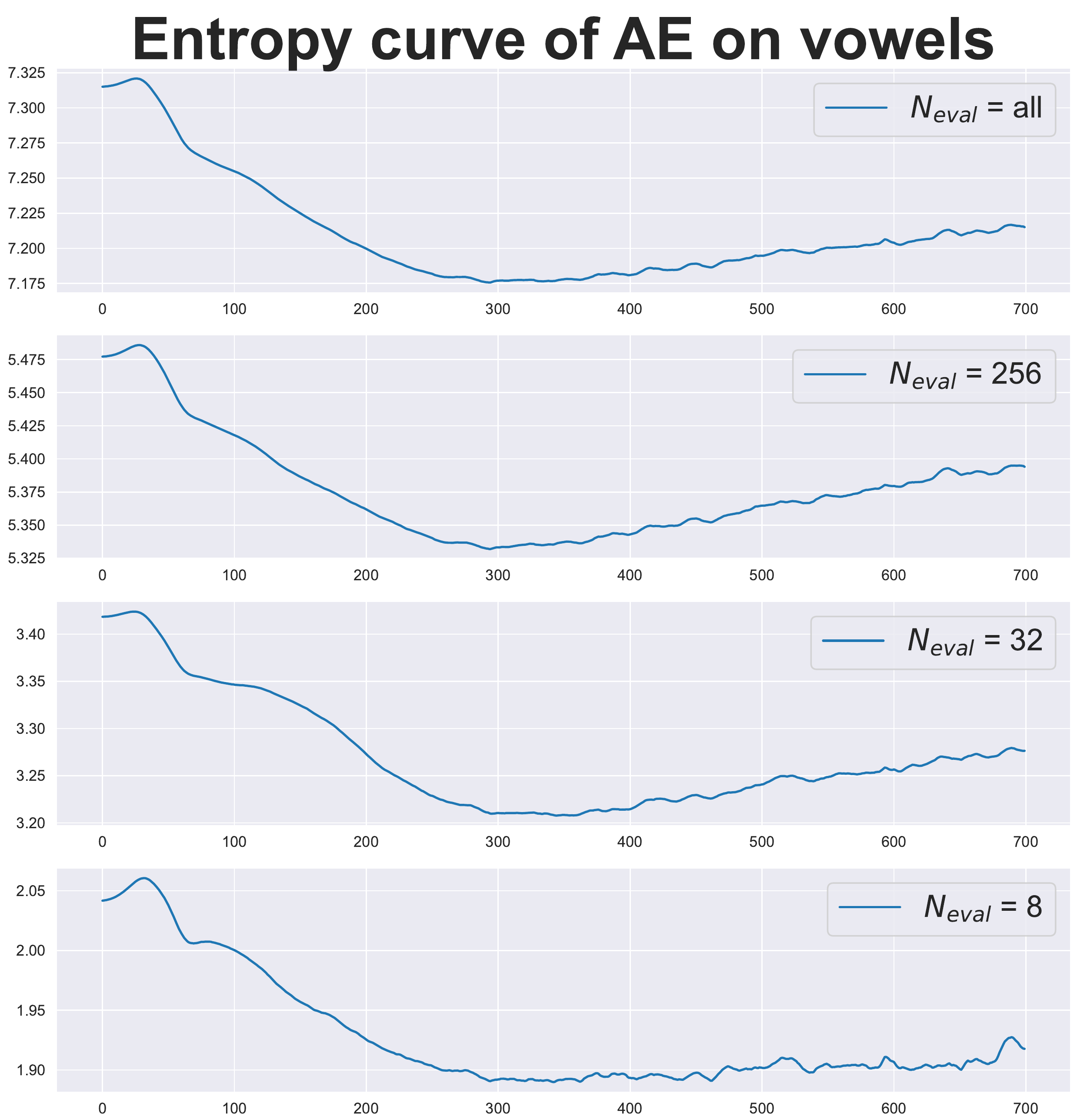}
  \caption{The loss entropy curves of the  training process of AE on the \textit{vowels} dataset with varying $N_{eval}$ values.}
  \label{Fig:appx-ae-n_eval}
\end{wrapfigure}
\begin{figure*}
  \centering
\includegraphics[width=130.00pt]{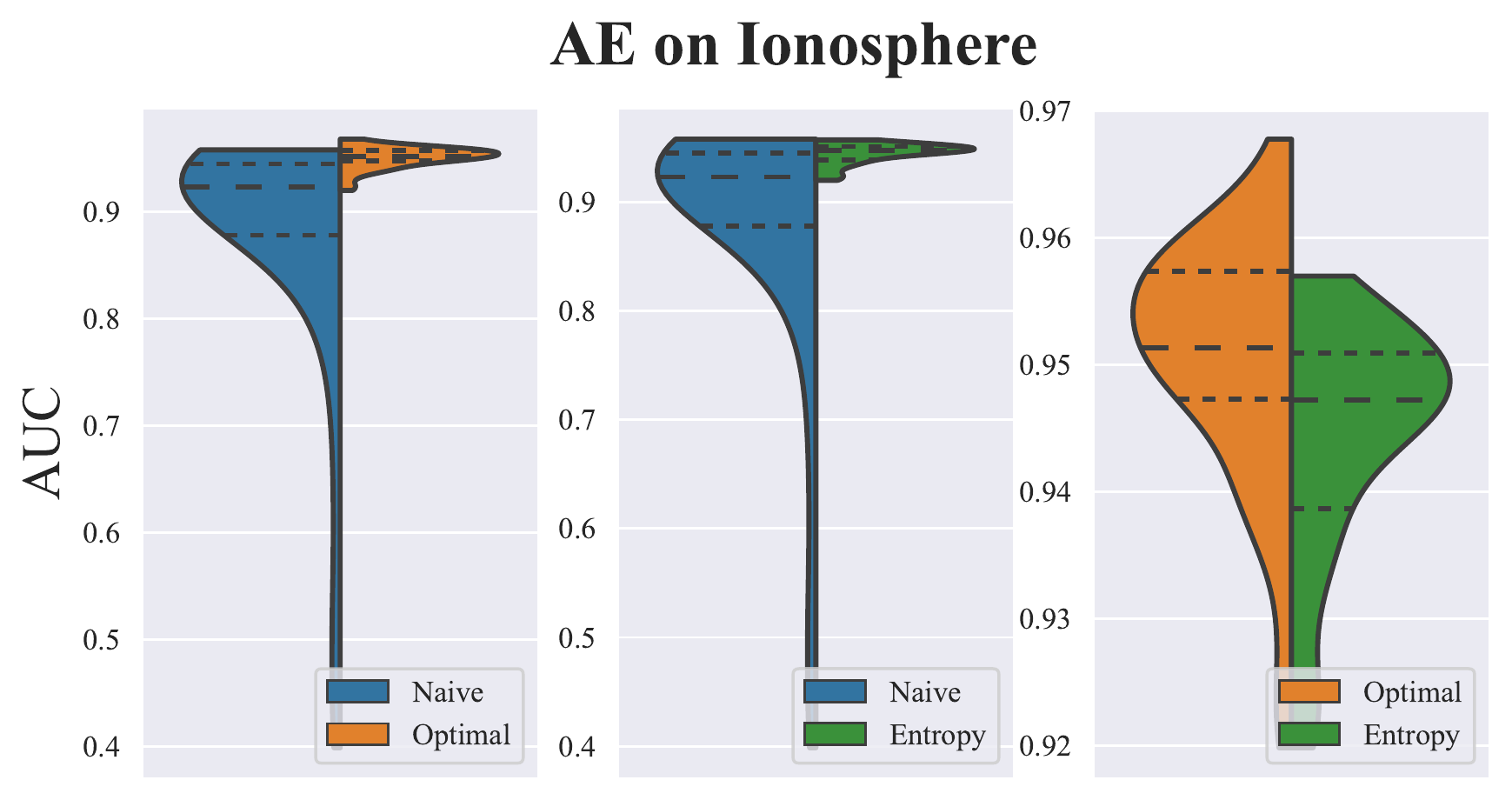}  \includegraphics[width=130.00pt]{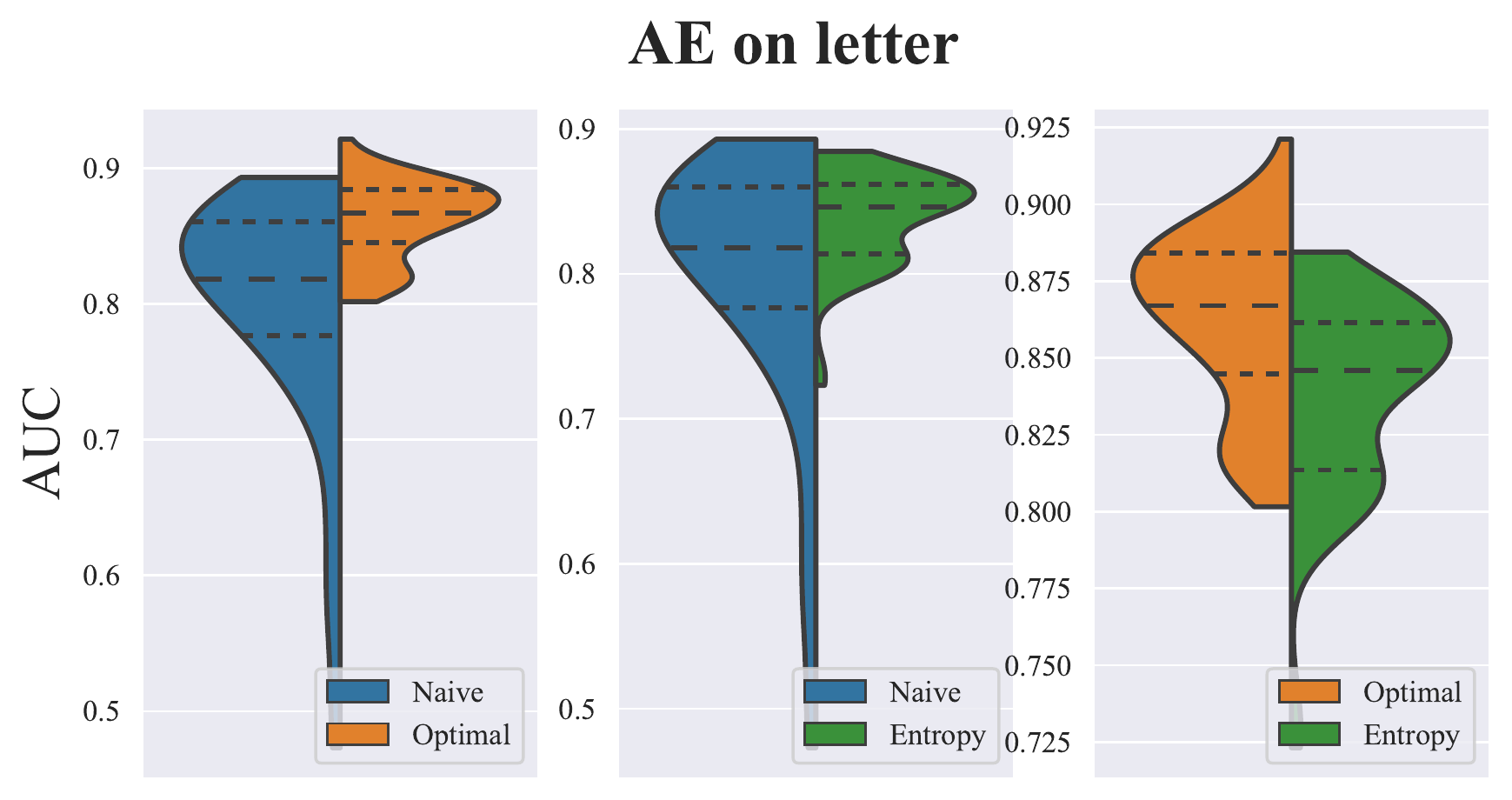}
   \includegraphics[width=130.00pt]{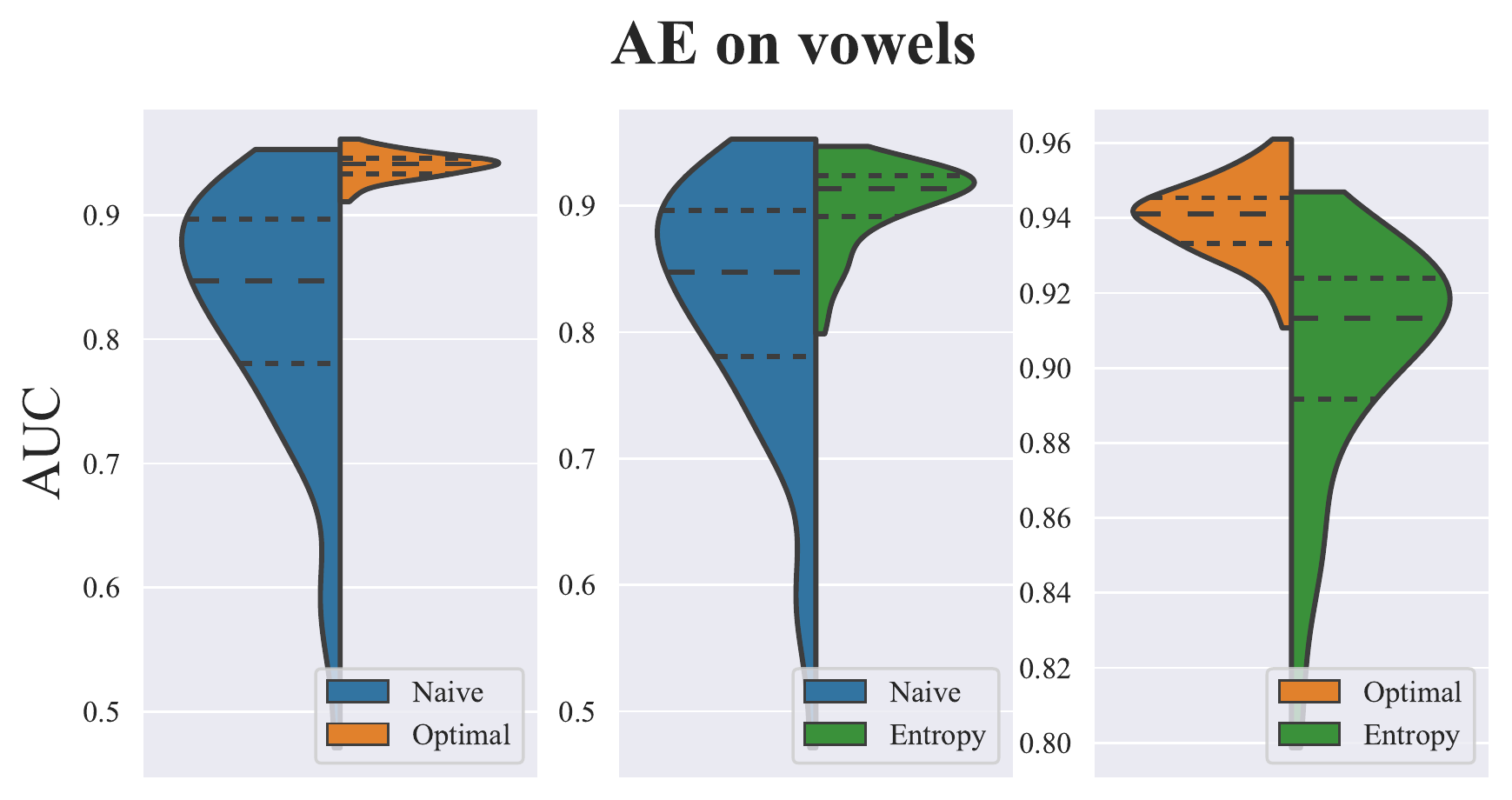}
    \includegraphics[width=130.00pt]{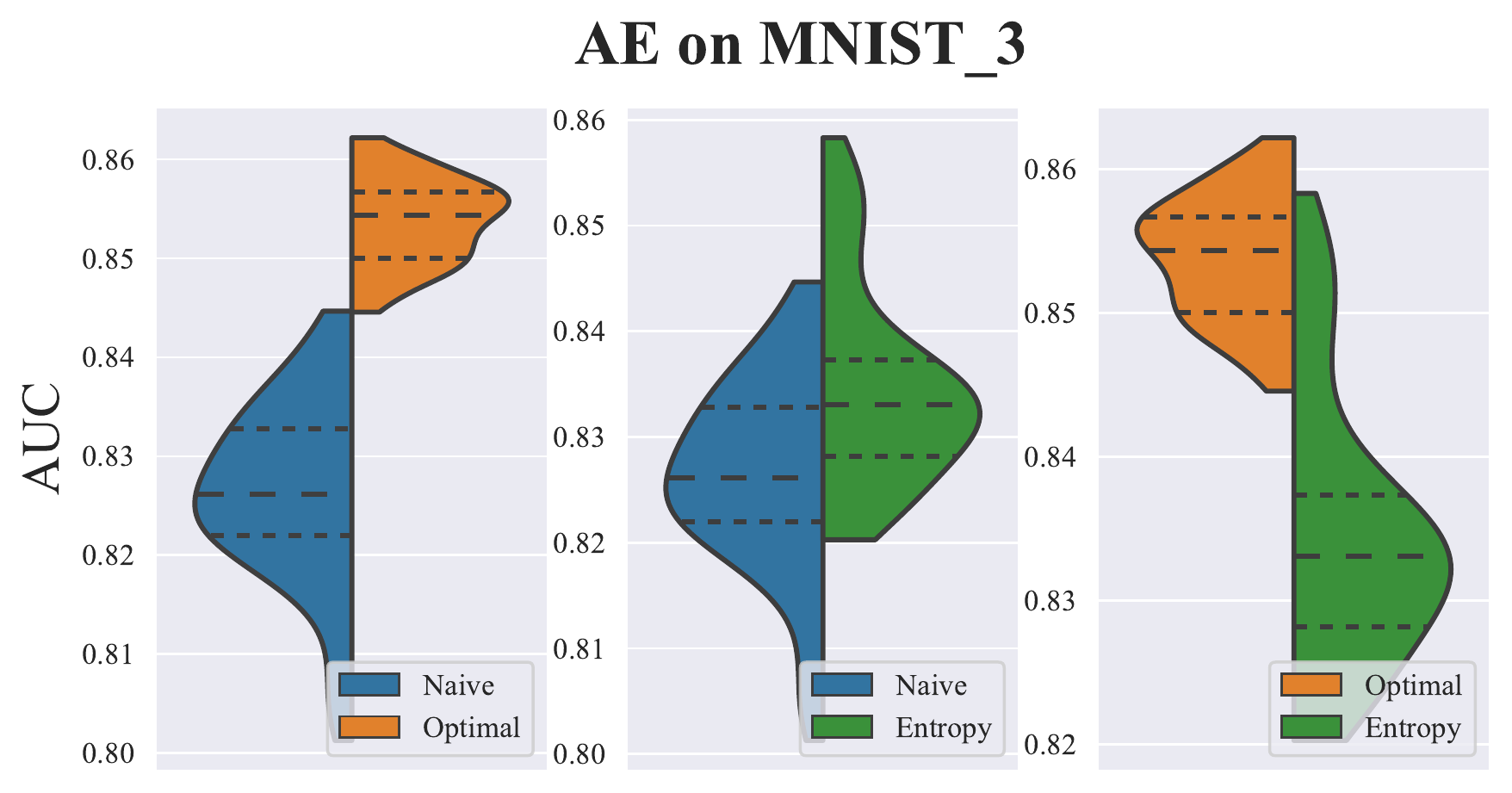}
     \includegraphics[width=130.00pt]{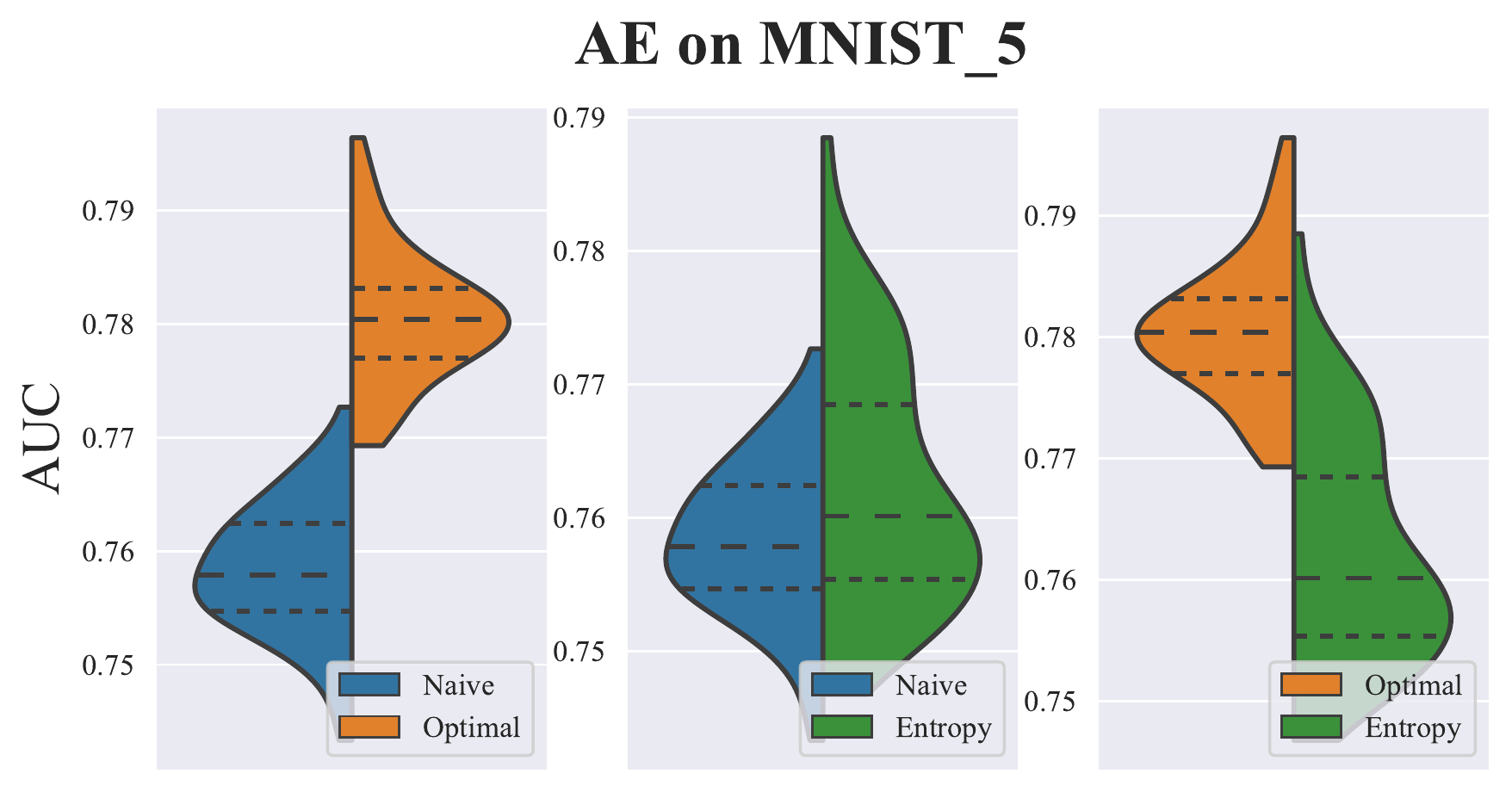}
      \includegraphics[width=130.00pt]{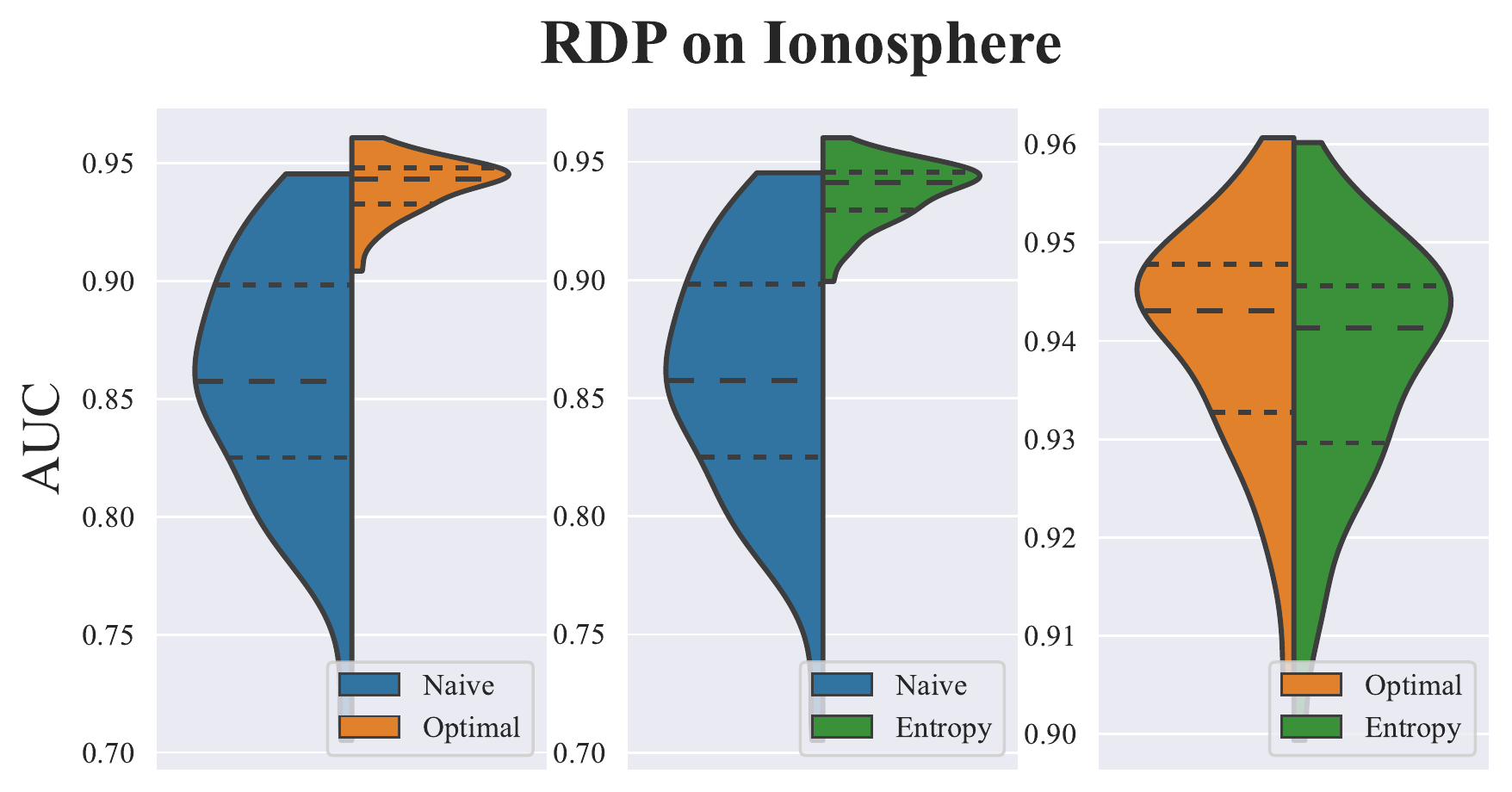}
       \includegraphics[width=130.00pt]{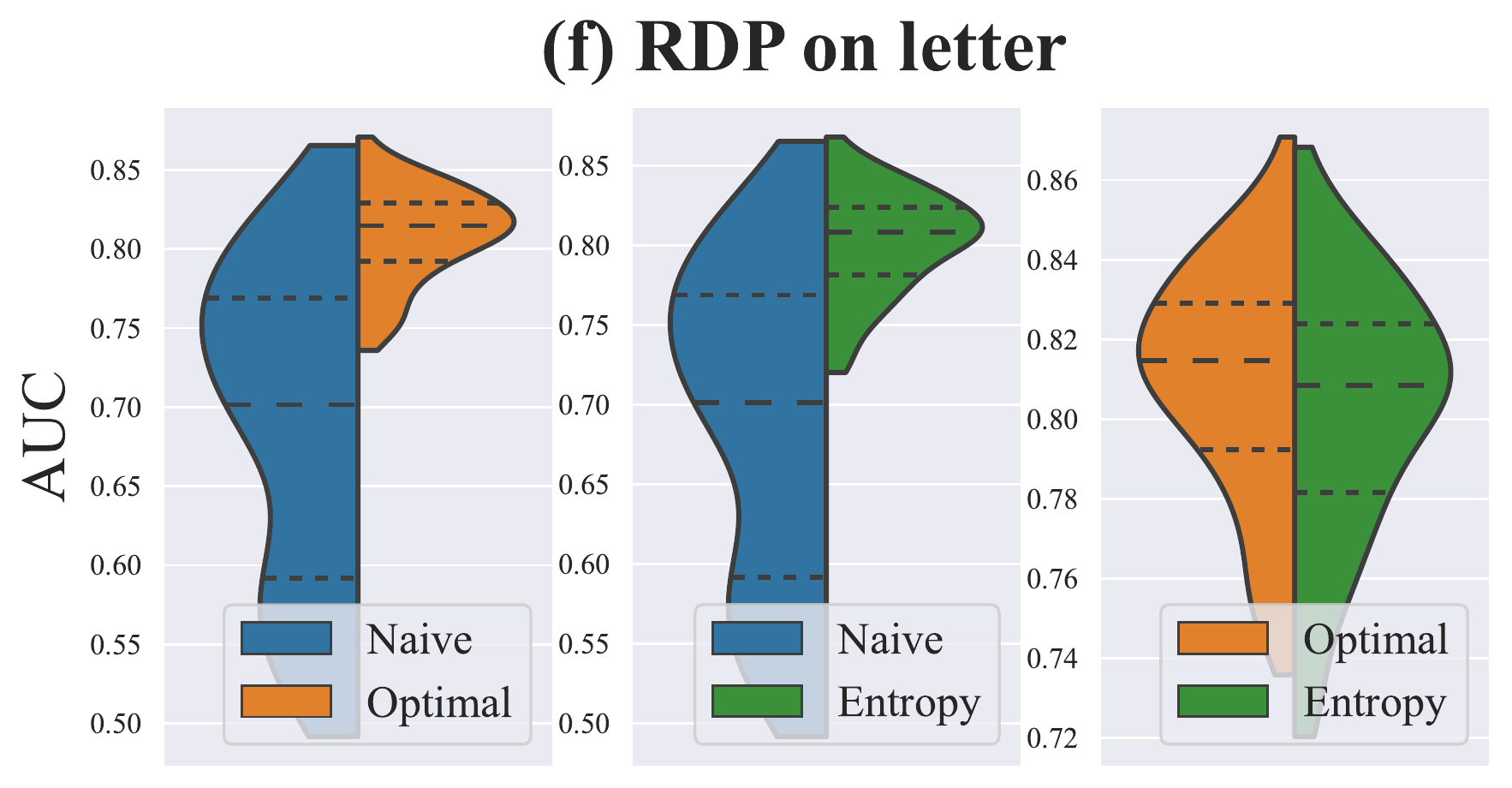}
        \includegraphics[width=130.00pt]{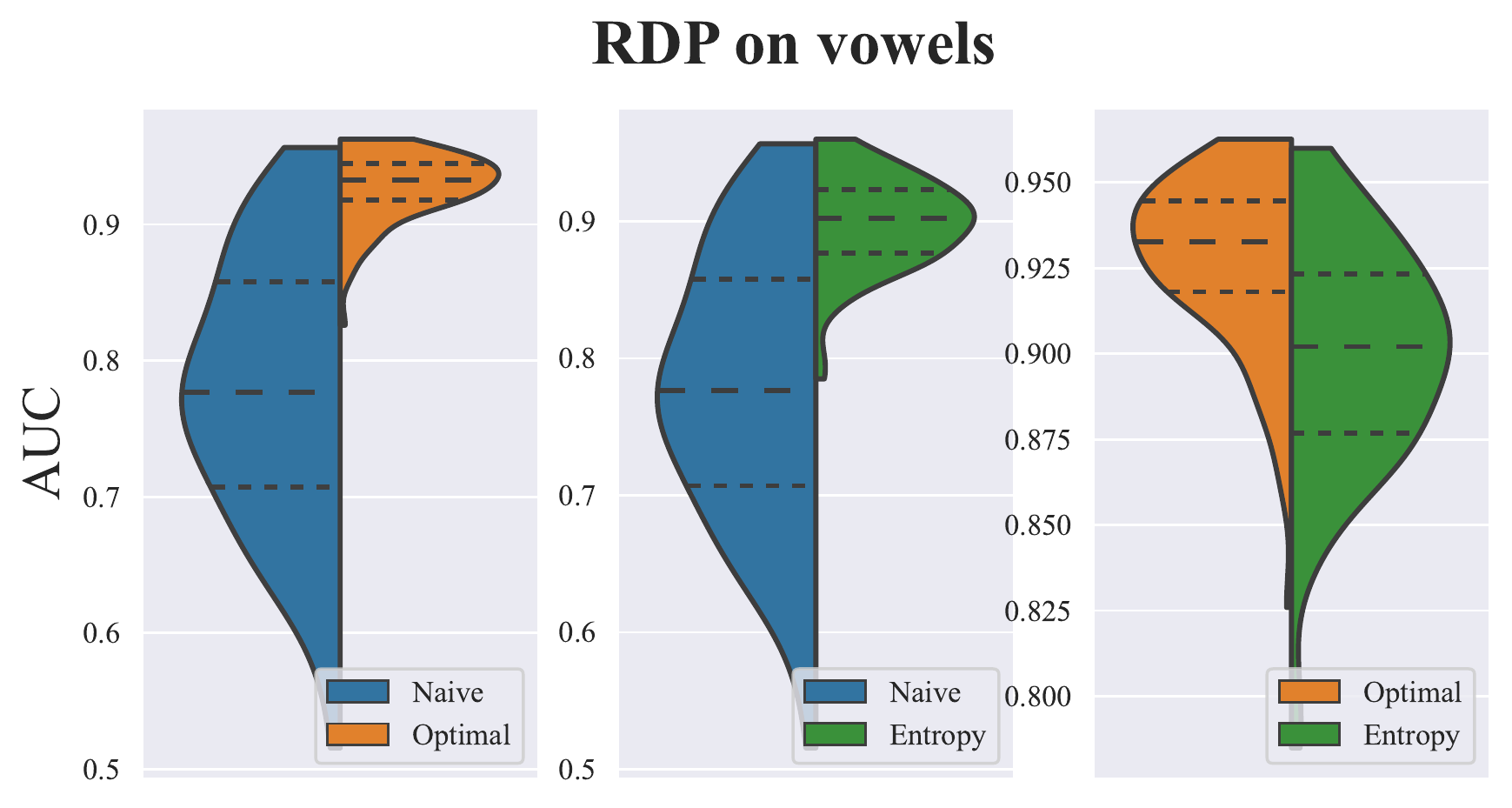}
         \includegraphics[width=130.00pt]{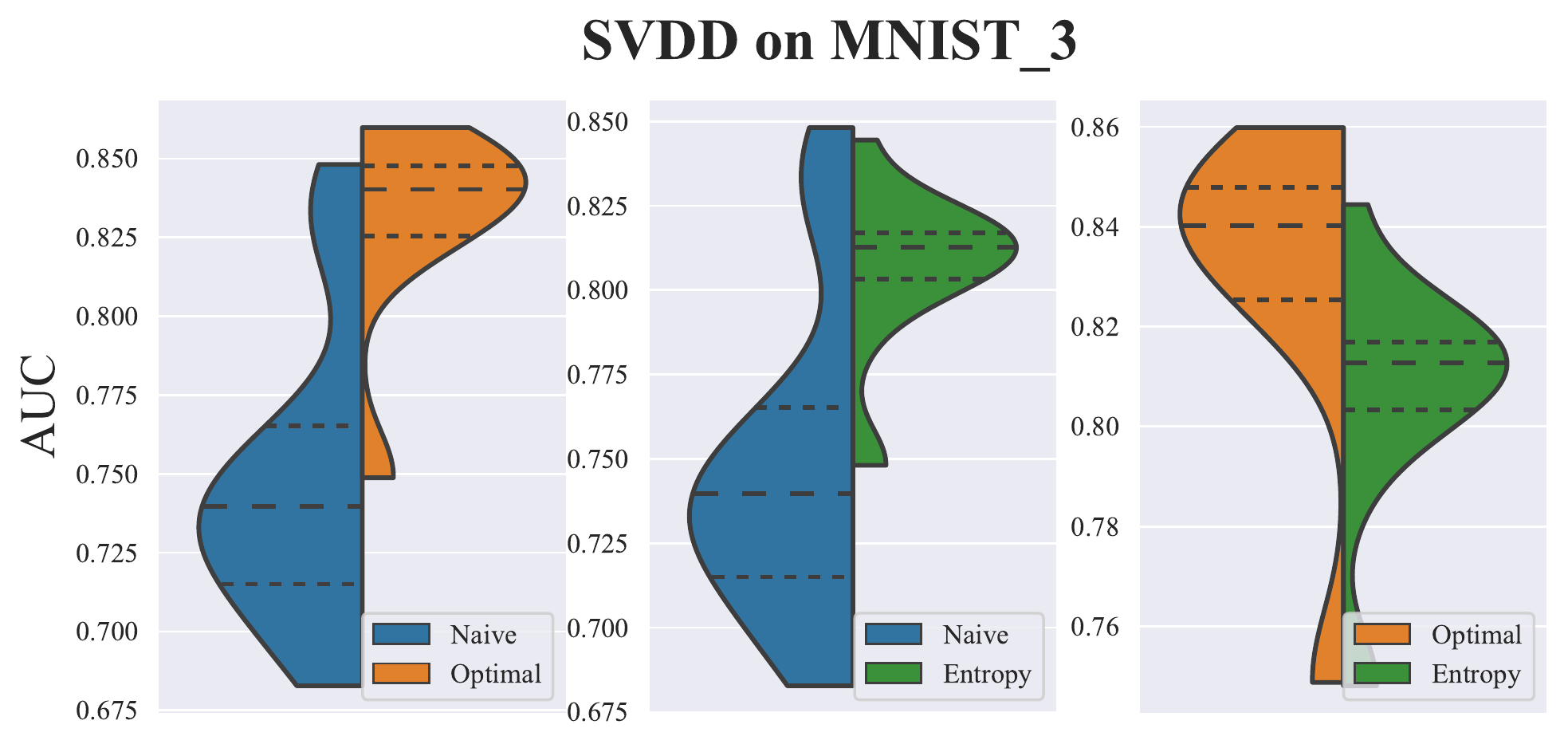}
          \includegraphics[width=130.00pt]{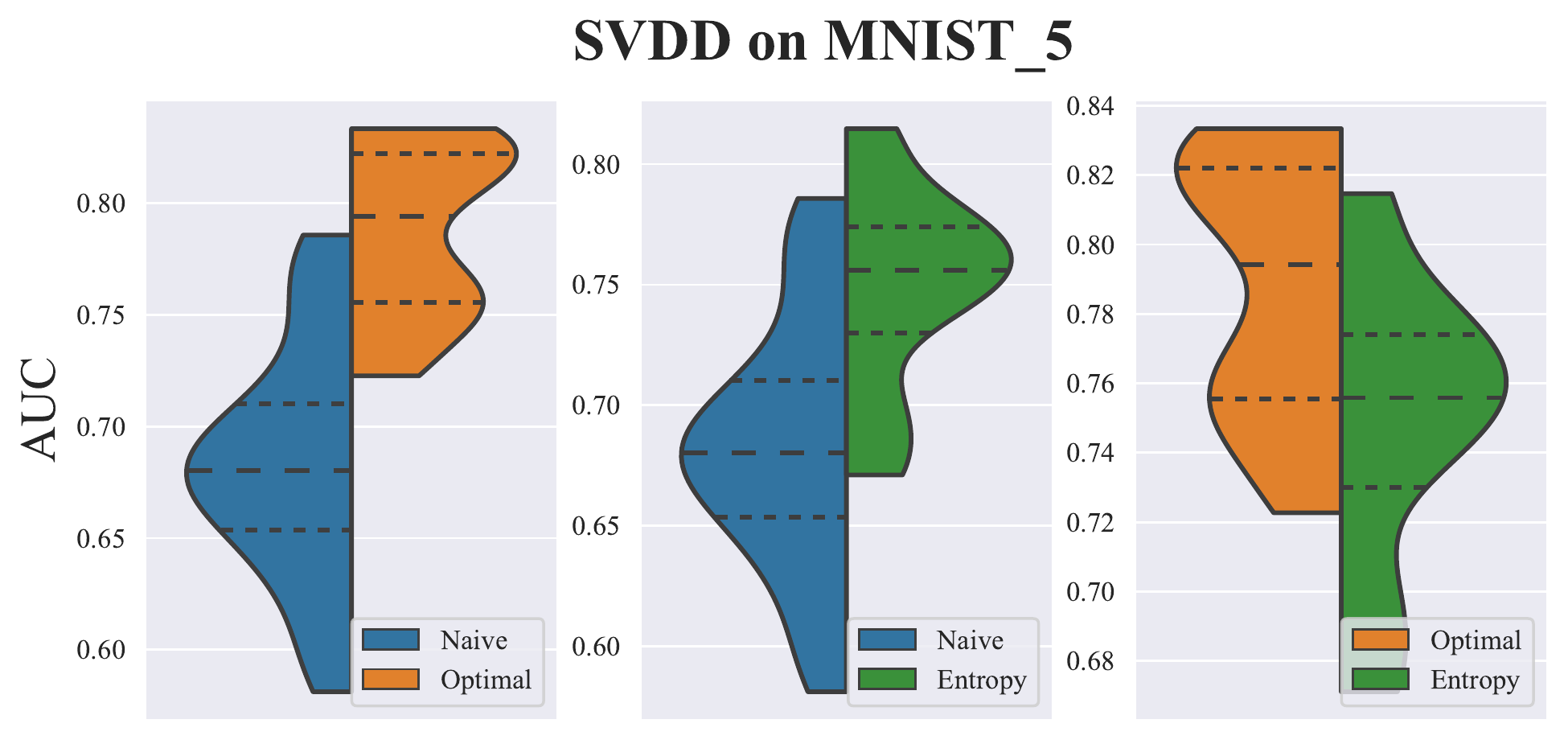}
  \caption{Comparison of AUC distribution  of \textit{Naive}, \textit{Optimal}, and \textit{Entropy} across varing HP configurations.}
  \label{Fig:appx-hp-auc-distribution}
\end{figure*}

The  comparison of the AUC distribution of the \textit{Naive}, \textit{Optimal}, and \textit{Entropy} across varing HP
configurations and models are shown in Fig \ref{Fig:appx-hp-auc-distribution}.

Figure \ref{Fig:appx-ae-n_eval} displays the loss entropy curve of the AE training process on the \textit{vowels} dataset with varying $N_{eval}$ values. The correlation between these entropy curves and the original entropy curve computed on the entire dataset is demonstrated in Figure \ref{fig:para-study} (b) of our primary paper. The results indicate that even with a small $N_{eval}$ (e.g., 8), the loss entropy curve remains quite similar to the entropy curve obtained by a larger $N_{eval}$.

\end{document}